\documentclass[10pt,twocolumn,letterpaper]{article}

\usepackage{iccv}
\usepackage{times}
\usepackage{epsfig}
\usepackage{graphicx}
\usepackage{amsmath}
\usepackage{amssymb}

\usepackage{amsfonts}
\usepackage{amsthm}
\usepackage{mathrsfs}
\usepackage{subcaption}
\usepackage{paralist}

\usepackage{color}

\usepackage[pagebackref=true,breaklinks=true,letterpaper=true,colorlinks,bookmarks=false]{hyperref}

\iccvfinalcopy 

\def\iccvPaperID{1520} 

\ificcvfinal\fi
\begin{document}

 \language0
\lefthyphenmin=2
\righthyphenmin=3

\title{ZM-Net: Real-time Zero-shot Image Manipulation Network}

\def\Blue{\color{blue}}
\def\Purple{\color{purple}}

\def\A{{\bf A}}
\def\a{{\bf a}}
\def\B{{\bf B}}
\def\b{{\bf b}}
\def\C{{\bf C}}
\def\c{{\bf c}}
\def\D{{\bf D}}
\def\d{{\bf d}}
\def\E{{\bf E}}
\def\e{{\bf e}}
\def\f{{\bf f}}
\def\F{{\bf F}}
\def\G{{\bf G}}
\def\K{{\bf K}}
\def\k{{\bf k}}
\def\L{{\bf L}}
\def\H{{\bf H}}
\def\h{{\bf h}}
\def\G{{\bf G}}
\def\g{{\bf g}}
\def\I{{\bf I}}
\def\R{{\bf R}}
\def\X{{\bf X}}
\def\Y{{\bf Y}}
\def\P{{\bf P}}
\def\Q{{\bf Q}}
\def\s{{\bf s}}
\def\S{{\bf S}}
\def\t{{\bf t}}
\def\T{{\bf T}}
\def\x{{\bf x}}
\def\y{{\bf y}}
\def\z{{\bf z}}
\def\Z{{\bf Z}}
\def\M{{\bf M}}
\def\m{{\bf m}}
\def\n{{\bf n}}
\def\U{{\bf U}}
\def\u{{\bf u}}
\def\V{{\bf V}}
\def\v{{\bf v}}
\def\W{{\bf W}}
\def\w{{\bf w}}
\def\0{{\bf 0}}
\def\1{{\bf 1}}

\def\AM{{\mathcal A}}
\def\EM{{\mathcal E}}
\def\FM{{\mathcal F}}
\def\TM{{\mathcal T}}
\def\UM{{\mathcal U}}
\def\XM{{\mathcal X}}
\def\YM{{\mathcal Y}}
\def\NM{{\mathcal N}}
\def\OM{{\mathcal O}}
\def\IM{{\mathcal I}}
\def\GM{{\mathcal G}}
\def\PM{{\mathcal P}}
\def\LM{{\mathcal L}}
\def\MM{{\mathcal M}}
\def\DM{{\mathcal D}}
\def\SM{{\mathcal S}}
\def\RB{{\mathbb R}}
\def\EB{{\mathbb E}}

\def\tx{\tilde{\bf x}}
\def\ty{\tilde{\bf y}}
\def\tz{\tilde{\bf z}}
\def\hd{\hat{d}}
\def\HD{\hat{\bf D}}
\def\hx{\hat{\bf x}}
\def\hR{\hat{R}}

\def\Ome{\mbox{\boldmath$\omega$\unboldmath}}
\def\bet{\mbox{\boldmath$\beta$\unboldmath}}
\def\ep{\mbox{\boldmath$\epsilon$\unboldmath}}
\def\et{\mbox{\boldmath$\eta$\unboldmath}}
\def\ph{\mbox{\boldmath$\phi$\unboldmath}}
\def\Pii{\mbox{\boldmath$\Pi$\unboldmath}}
\def\pii{\mbox{\boldmath$\pi$\unboldmath}}
\def\Ph{\mbox{\boldmath$\Phi$\unboldmath}}
\def\ps{\mbox{\boldmath$\psi$\unboldmath}}
\def\Ps{\mbox{\boldmath$\Psi$\unboldmath}}
\def\tha{\mbox{\boldmath$\theta$\unboldmath}}
\def\muu{\mbox{\boldmath$\mu$\unboldmath}}
\def\Si{\mbox{\boldmath$\Sigma$\unboldmath}}
\def\Gam{\mbox{\boldmath$\Gamma$\unboldmath}}
\def\ga{\mbox{\boldmath$\gamma$\unboldmath}}
\def\Lam{\mbox{\boldmath$\Lambda$\unboldmath}}
\def\De{\mbox{\boldmath$\Delta$\unboldmath}}
\def\vps{\mbox{\boldmath$\varepsilon$\unboldmath}}
\def\Up{\mbox{\boldmath$\Upsilon$\unboldmath}}
\def\Lap{\mbox{\boldmath$\LM$\unboldmath}}
\newcommand{\ti}[1]{\tilde{#1}}

\def\tr{\mathrm{tr}}
\def\etr{\mathrm{etr}}
\def\etal{{\em et al.\/}\,}
\newcommand{\indep}{{\;\bot\!\!\!\!\!\!\bot\;}}
\def\argmax{\mathop{\rm argmax}}
\def\argmin{\mathop{\rm argmin}}
\def\vec{\text{vec}}
\def\cov{\text{cov}}
\def\dg{\text{diag}}


\author{
Hao Wang$^1$, Xiaodan Liang$^2$, Hao Zhang$^2$$^,$$^3$, Dit-Yan Yeung$^1$, and Eric P. Xing$^3$\\
\vspace{-8pt}
              \\
       $^1$Hong Kong University of Science and Technology \ $^2$Carnegie Mellon University \ $^3$Petuum Inc. \\
       \small \{hwangaz, dyyeung\}@cse.ust.hk, \{xiaodan1, hao\}@cs.cmu.edu, eric.xing@petuum.com\\
}

\twocolumn[{%
\renewcommand\twocolumn[1][]{#1}%
\vspace{-1em}
\maketitle
\vspace{-1em}
\begin{center}
    \centering
        \includegraphics[width=0.13\linewidth]{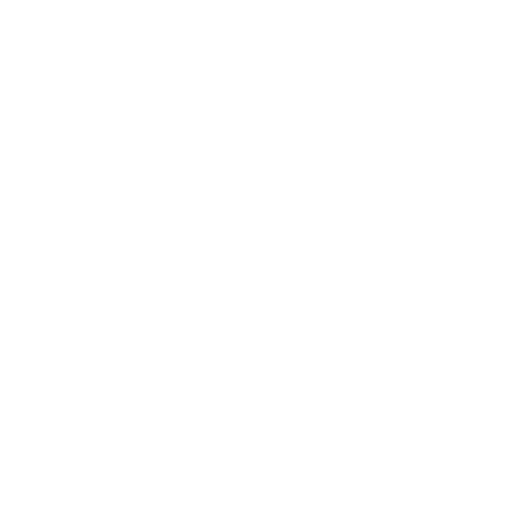}
        \includegraphics[width=0.13\linewidth]{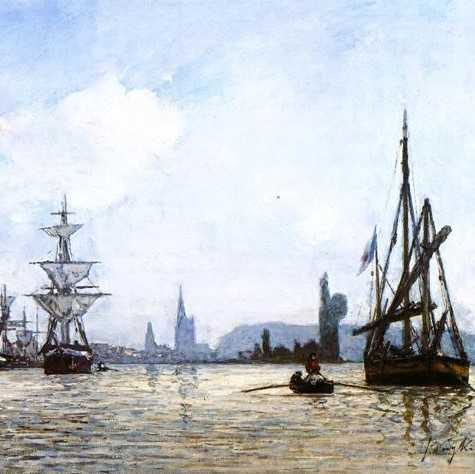}
        \includegraphics[width=0.13\linewidth]{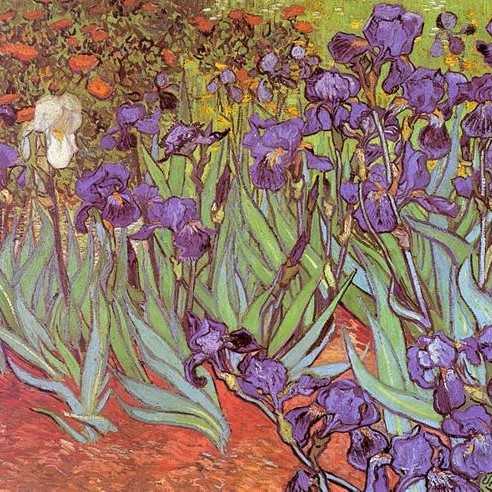}
        \includegraphics[width=0.13\linewidth]{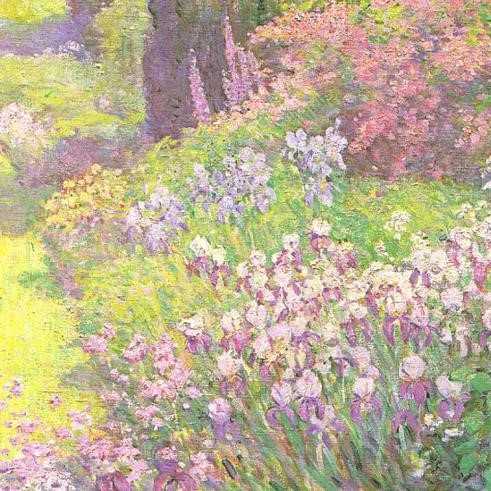}
        \includegraphics[width=0.13\linewidth]{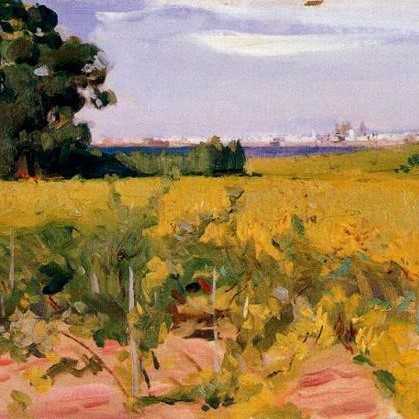}
        \includegraphics[width=0.13\linewidth]{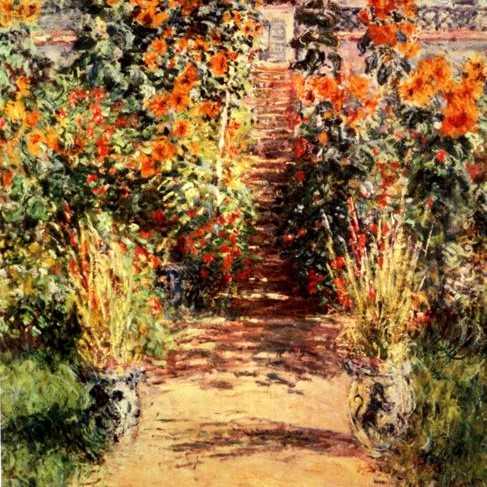}
        \includegraphics[width=0.13\linewidth]{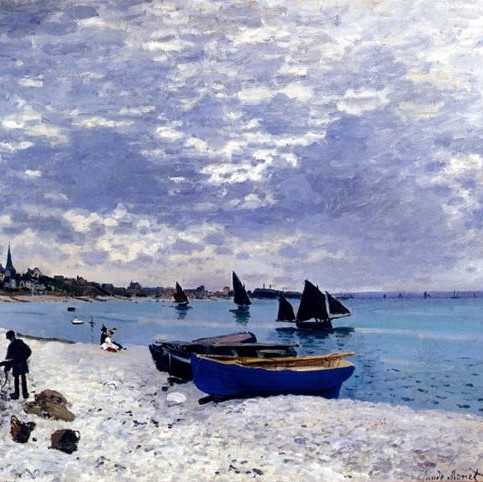} \\
        \vskip 0.06cm
        \includegraphics[width=0.13\linewidth]{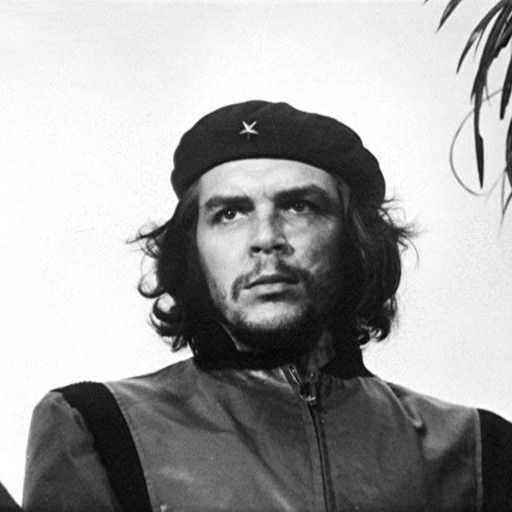}
        \includegraphics[width=0.13\linewidth]{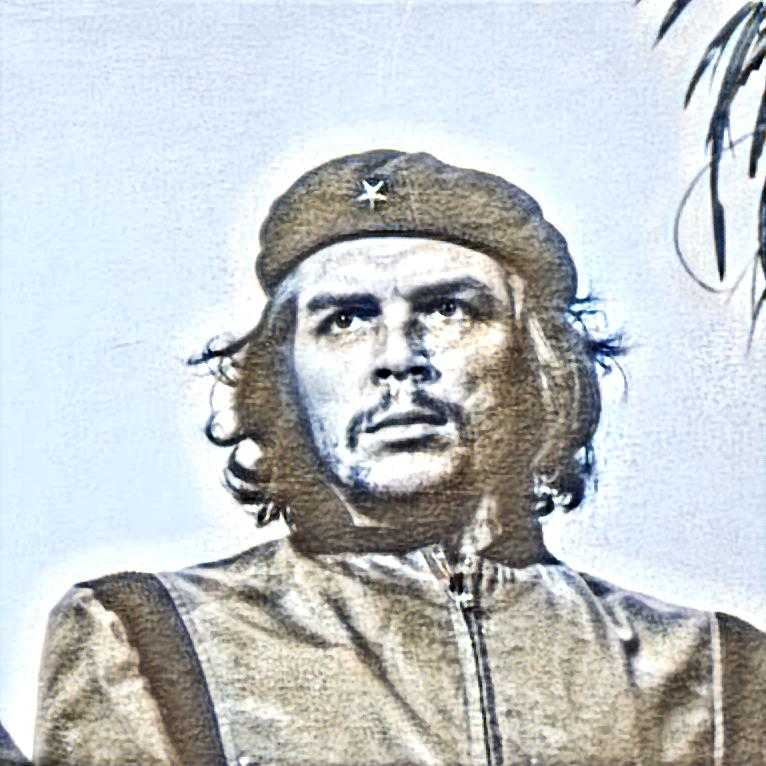}
        \includegraphics[width=0.13\linewidth]{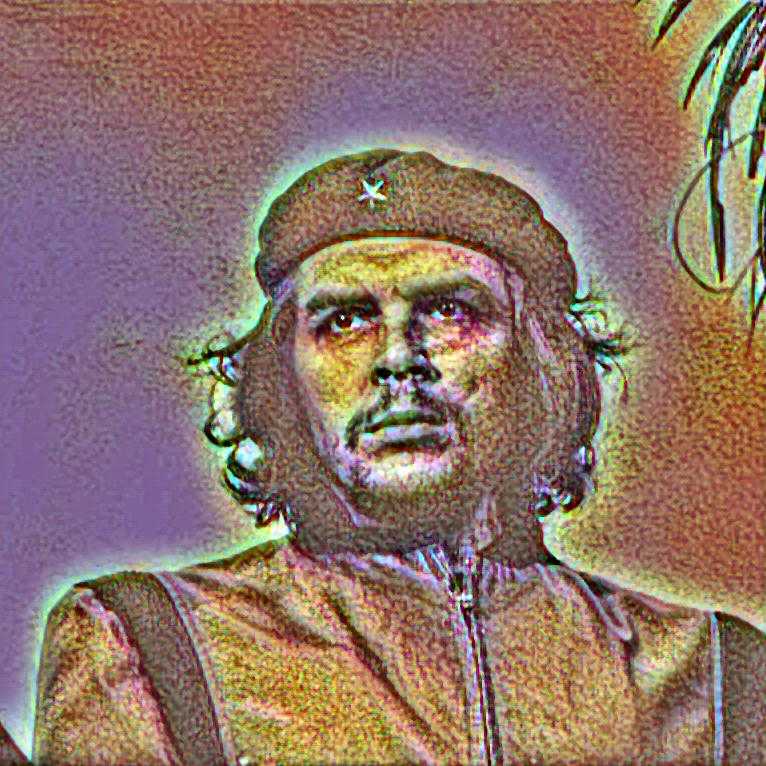}
        \includegraphics[width=0.13\linewidth]{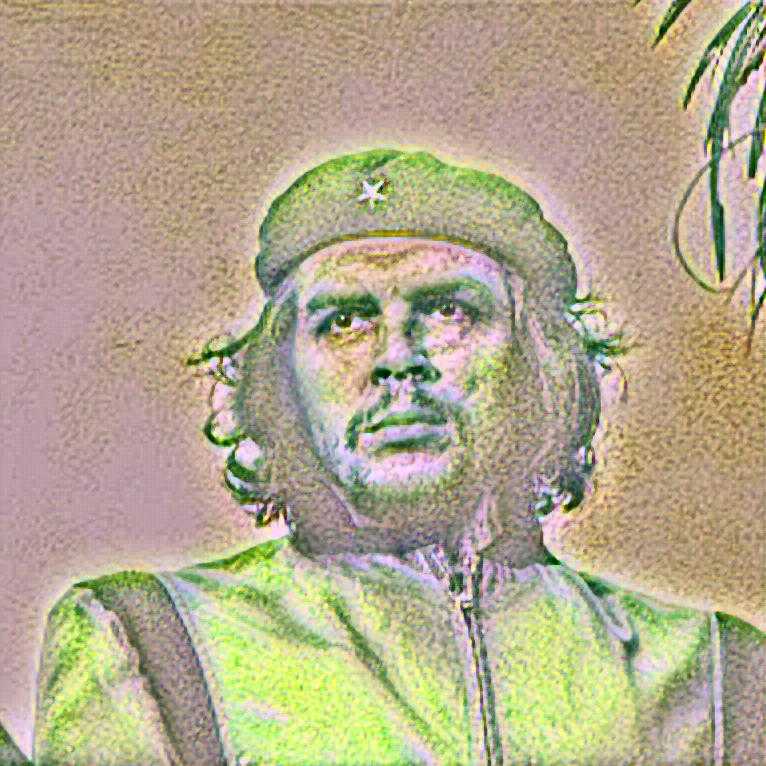}
        \includegraphics[width=0.13\linewidth]{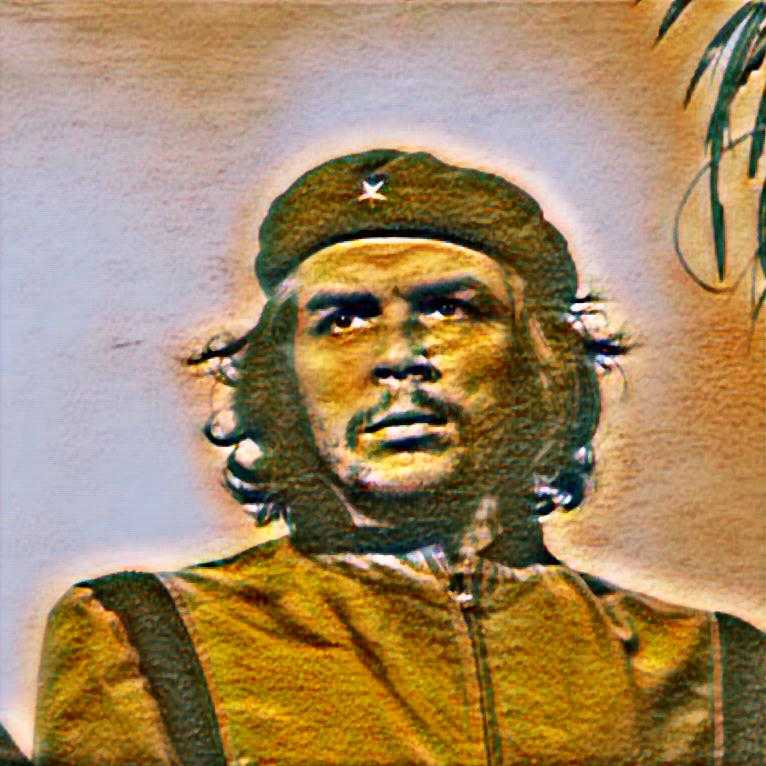}
        \includegraphics[width=0.13\linewidth]{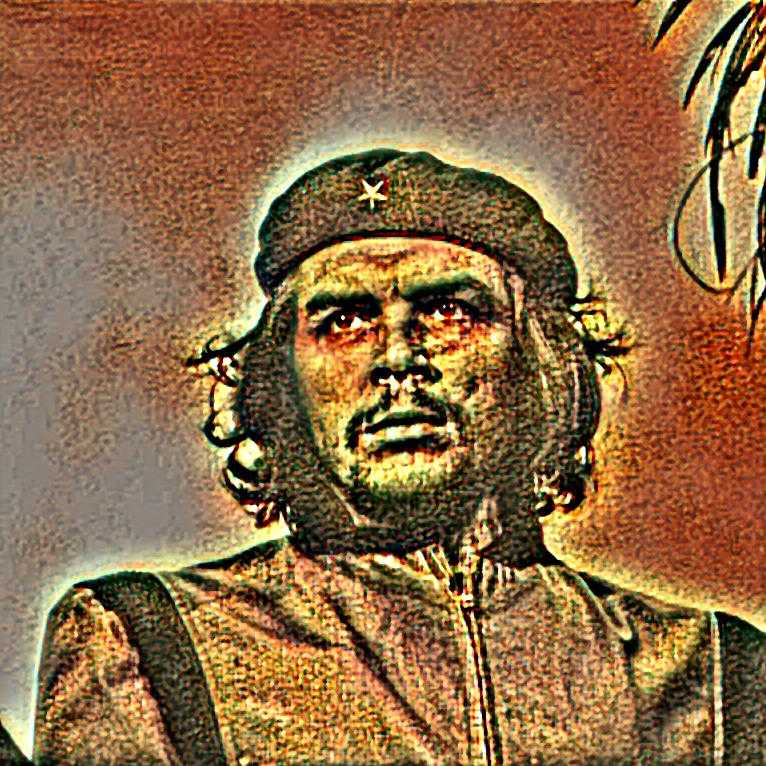}
        \includegraphics[width=0.13\linewidth]{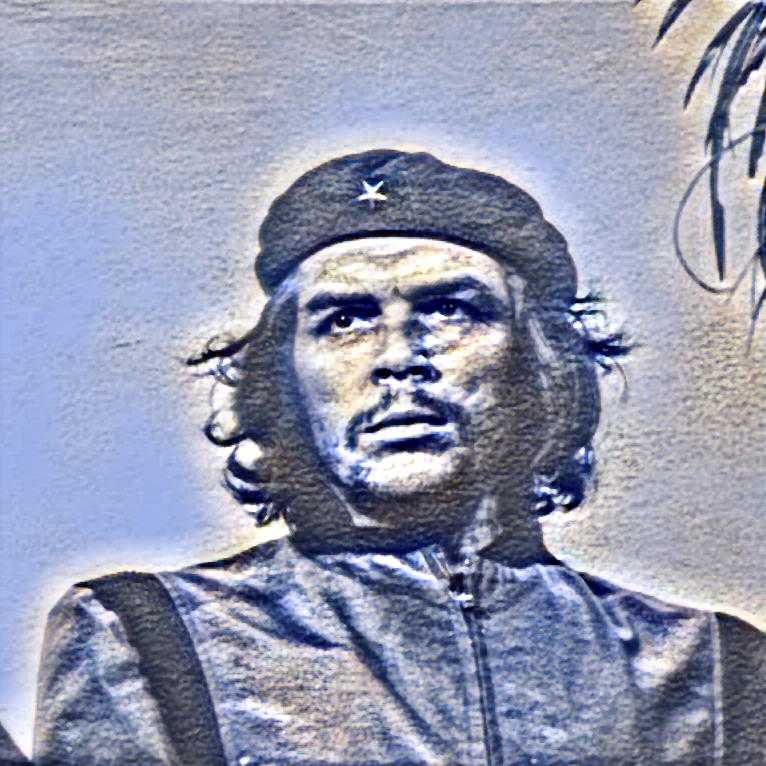} \\
        \vskip 0.06cm
        \includegraphics[width=0.13\linewidth]{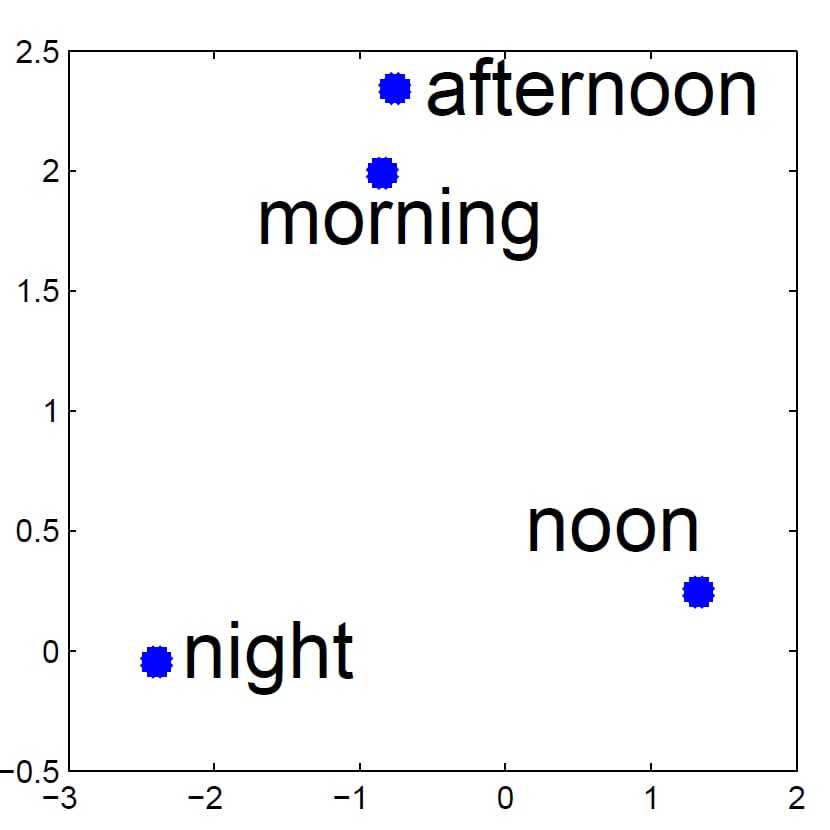}
        \includegraphics[width=0.13\linewidth]{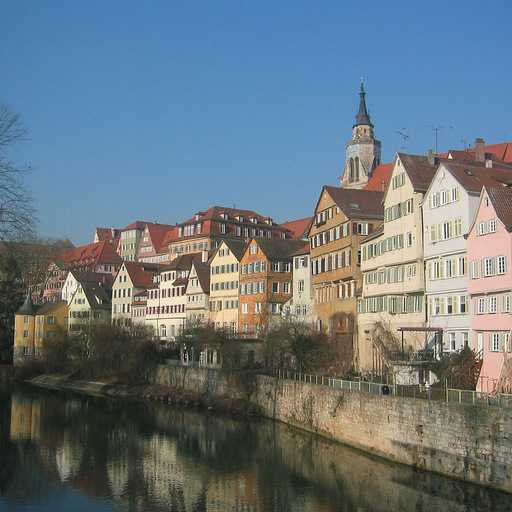}
        \includegraphics[width=0.13\linewidth]{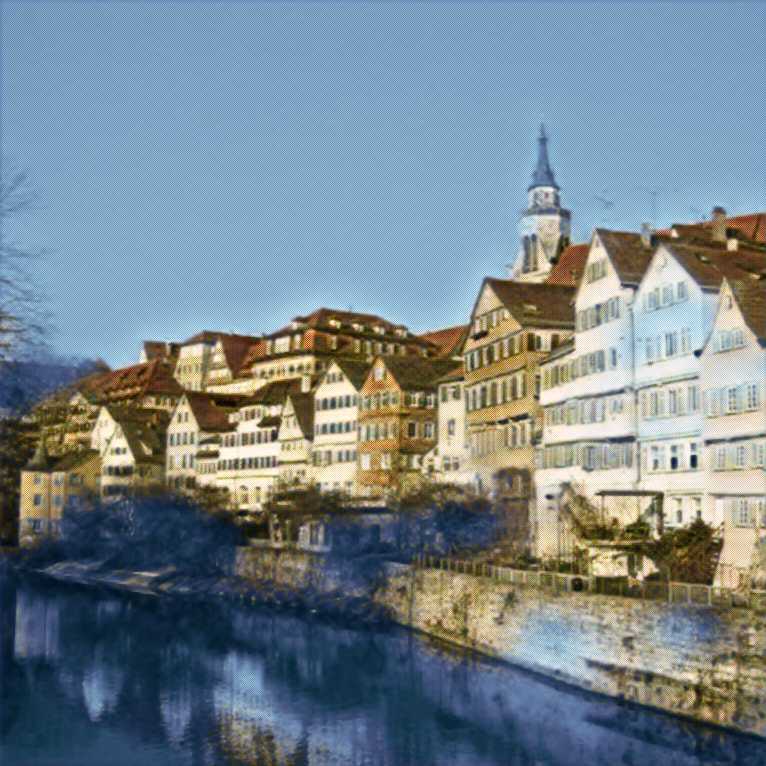}
        \includegraphics[width=0.13\linewidth]{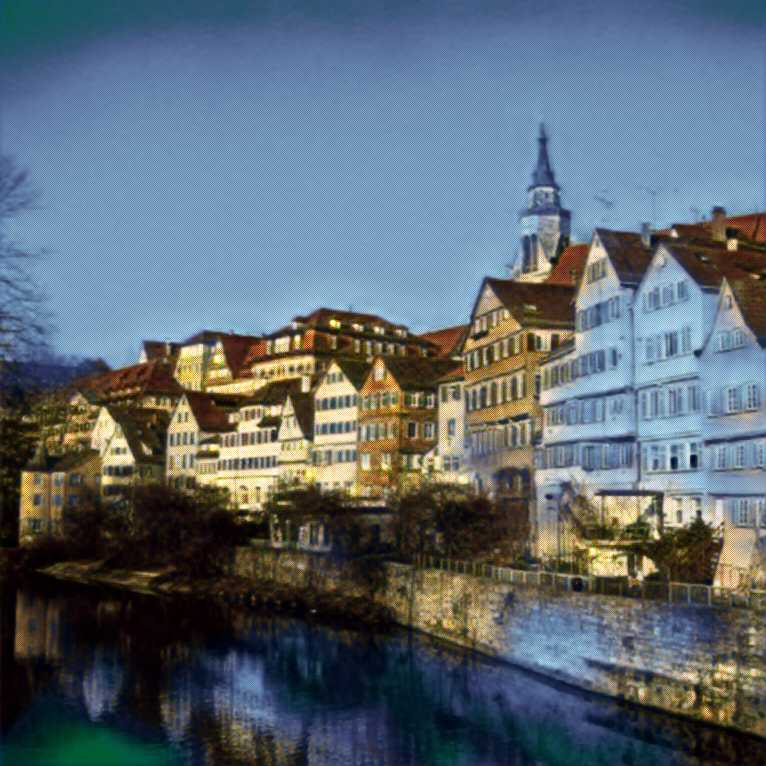}
        \includegraphics[width=0.13\linewidth]{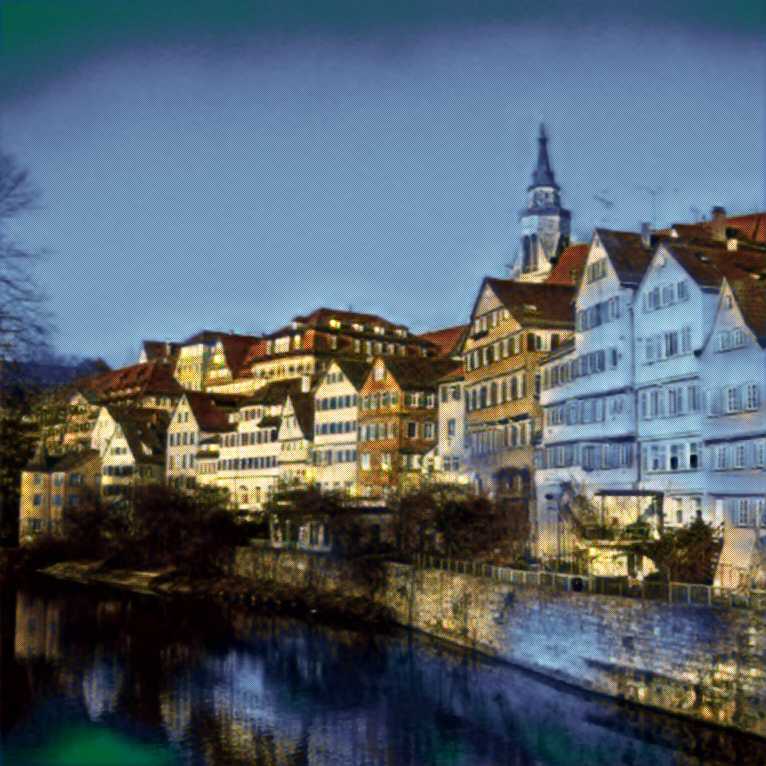}
        \includegraphics[width=0.13\linewidth]{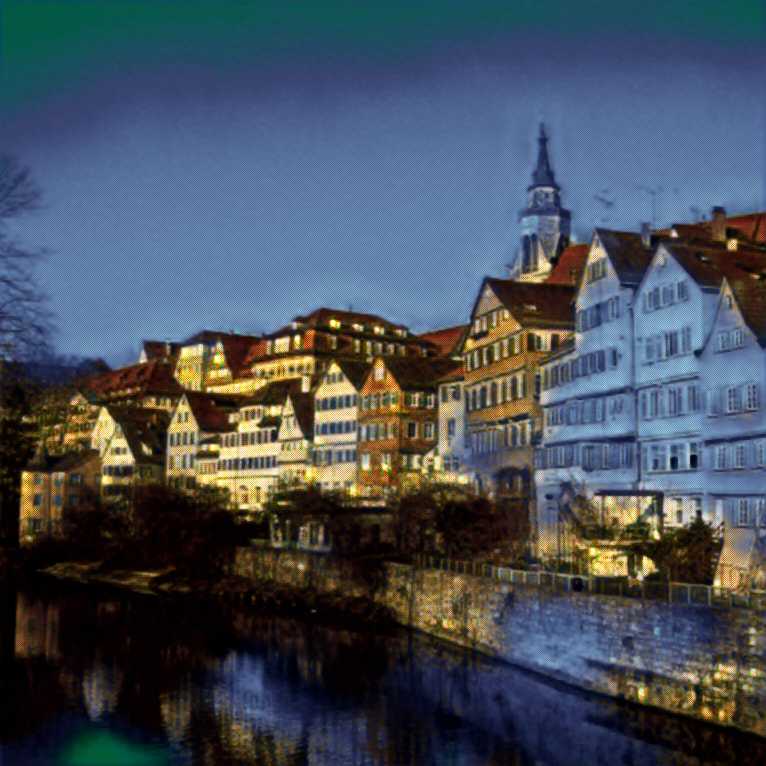}
        \includegraphics[width=0.13\linewidth]{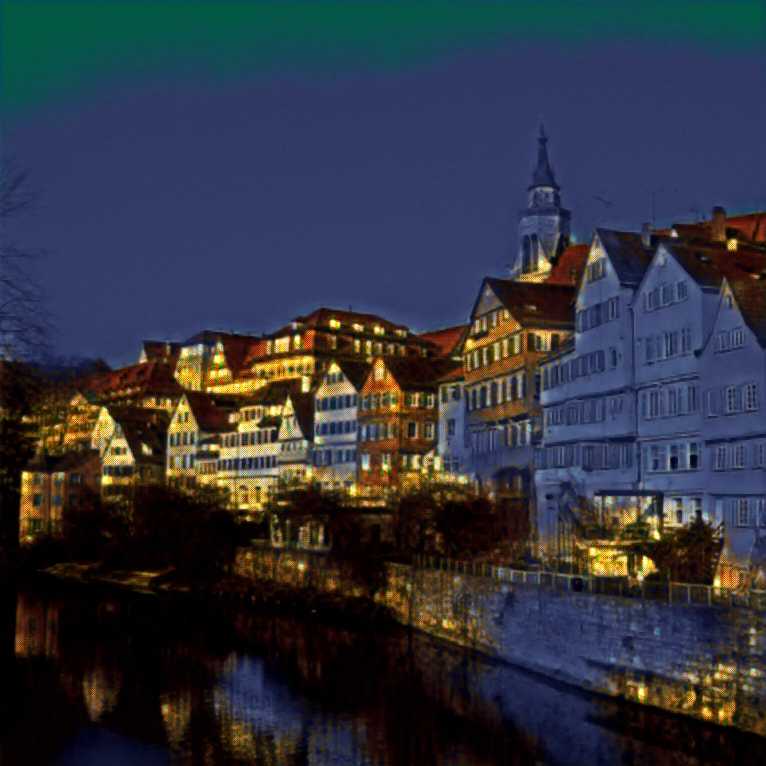}
        \vskip -0.20cm
    \captionof{figure}{\small Example results by our Zero-shot image Manipulation Network (ZM-Net) that can manipulate images guided by any personalized signals in real-time.
Row 2: zero-shot style transfer guided by different landscape paintings from Row 1.
Row 3: image manipulation conditioned on descriptive attributes; from left to right are
descriptive attributes, the input image, and the $5$ transformed images corresponding to the text `noon', `afternoon', `morning', $0.5\mbox{`morning'}+0.5\mbox{`night'}$, and `night', respectively.}
    \label{teaser}
    \vspace{-5pt}
\end{center}%
}]


\begin{abstract}
\vspace{-15pt}
Many problems in image processing and computer vision (e.g. colorization, style transfer) can be posed as ``manipulating'' an input image into a corresponding output image given a user-specified guiding signal. A holy-grail solution towards generic image manipulation should be able to efficiently alter an input image with any personalized signals (even signals unseen during training), such as diverse paintings and arbitrary descriptive attributes. However, existing methods are either inefficient to simultaneously process multiple signals (let alone generalize to unseen signals), or unable to handle signals from other modalities. In this paper, we make the first attempt to address the zero-shot image manipulation task. We cast this problem as manipulating an input image according to a parametric model whose key parameters can be conditionally generated from any guiding signal (even unseen ones). To this end, we propose the Zero-shot Manipulation Net (ZM-Net), a fully-differentiable architecture that jointly optimizes an image-transformation network (TNet) and a parameter network (PNet). The PNet learns to generate key transformation parameters for the TNet given any guiding signal while the TNet performs fast zero-shot image manipulation according to both signal-dependent parameters from the PNet and signal-invariant parameters from the TNet itself.  Extensive experiments show that our ZM-Net can perform high-quality image manipulation conditioned on different forms of guiding signals (e.g. style images and attributes) in real-time (tens of milliseconds per image) even for unseen signals. Moreover, a large-scale style dataset with over 20,000 style images is also constructed to promote further research.
\end{abstract}

\vspace{-18pt}
\section{Introduction}
\vspace{-5pt}
Image manipulation, which aims to manipulate an input image based on personalized guiding signals expressed in diverse modalities (e.g. art paintings or text attributes), has recently attracted ever-growing research interest and derived  
various real-world applications, such as attribute-driven image editing and artistic style transfer (e.g. Prisma).


An image manipulation model is usually deployed in various devices, ranging from a GPU desktop to a mobile phone. For such a solution to be applicable, we argue that it must meet three requirements: first, the model should be zero-shot -- it can immediately capture the intrinsic manipulation principles conveyed by the guiding signal and apply it on the target image, without retraining distinct models for every user input. Further, to support the downstream mobile applications, the inference process for a target image should be really efficient (regardless of where it happens, remote server or local mobile), so that the user can immediately obtain the desired output without waiting for seconds to minutes. Third, a personalized guiding signal usually comes in different forms -- it could either be an artistic style conveyed by an art painting (Fig~\ref{teaser} second row), or some descriptive phrases typed in by the user (Fig~\ref{teaser} third row), or even a speech instruction -- therefore, it is preferable that the model possesses the capability of receiving arbitrary guiding signal in multiple modalities.

A variety of relevant approaches have been developed towards the goal of real-time zero-shot image manipulation. Existing approaches, such as ~\cite{pami/Farabet13,icml/Pinheiro14,FCN,iccv/Eigen15,iccv/Noh15,CRFRNN}, mainly focus on training transformation neural networks that corresponds to a small set of guiding signals, such as a few art paintings. Among them, some CNN-based methods can process images (nearly) in real-time~\cite{PLoss}; however, each of their networks is only tied to one specific guiding signal (e.g. a single style image) and cannot generalize to unseen types specified by users, unless retraining as many networks as the number of guiding signals, which is both computationally and time prohibitive. Although some recent approaches~\cite{DNS} try to encode multiple styles within a single network, they fail to perform zero-shot style transfer  and cannot process guiding signals either in real-time or from distinct modalities (e.g. text attributes).

In this paper, we make the first attempt to explore the real-time zero-shot image manipulation task, to our best knowledge. This task is challenging since the model should be able to exploit and transform diverse and complex patterns from arbitrary guiding signals into transformation parameters, and perform the image manipulation in real-time. To this end, we propose a novel Zero-shot Manipulation Net (ZM-Net) that combines a parameter network (PNet) and an image-transformation network (TNet) into an end-to-end framework. The PNet is a generic model to produce a hierarchy of key transformation parameters, while TNet takes these generated parameters, combines with its own signal-invariant parameters, to generate a new image. In the sense of image style transfer, the PNet can embed any style image into the hierarchical parameters, which are used by TNet to transform the content image to a stylized image. We show that ZM-Net can digest over \emph{$20{,}000$ style images in a single network} rather than training one network per style as most previous methods did. It can also be trained to process guiding signals in other forms, such as descriptive attributes. Moreover, with the ability of fast zero-shot manipulation, the proposed ZM-Net can generate \emph{animation} of a single image in real-time (tens of milliseconds per image) even though the model is trained on \emph{images rather than videos}.

In summary, our main contributions are as follows: (1) To our best knowledge, this is the first scalable solution for the real-time zero-shot image manipulation task -- the proposed ZM-Net is able to digest over $20{,}000$ style images with a single model and perform zero-shot style transfer in real-time. Interestingly, even in the zero-shot setting (no retraining/finetuning), ZM-Net can still generate images with quality comparable to previous methods that need to retrain models for new style images.
(2) Our ZM-Net can handle more general image manipulation tasks (beyond style transfer) with different forms of guiding signals (e.g. text attributes).
(3) Using a small set of $984$ seed style images, we construct a much larger dataset of $23{,}307$ style images with much more content diversity. Experiments show that training on this dataset can dramatically decrease the testing loss nearly by half.
\vspace{-5pt}
\section{Related Work}
\vspace{-5pt}
A lot of research efforts have been devoted to the image manipulation task, among which the most common and efficient approach is to train a convolutional neural network (CNN) which directly outputs a transformed image for the input content image~\cite{pami/Farabet13,icml/Pinheiro14,FCN,iccv/Eigen15,iccv/Noh15,CRFRNN,cvpr/Liu15}.
For example, 
in~\cite{DeepColor,eccv/Zhang16}, a CNN is trained to perform colorization on input images, and in~\cite{PLoss,TextureNet,RNS,DNS} to transform content images according to specific styles. Although the most recent method by~\cite{PLoss} can process images (nearly) in real-time, it has to train a single network for each specific type of manipulation (e.g. a specific style image in style transfer) and cannot generalize to other types of manipulation (new style images or other forms of guiding signals) unless retraining the model for every type, which usually takes several hours and 
prevents them from being scaled to real-world applications. One of the most relevant works with ours in~\cite{DNS} tries to encode multiple styles within a single network; however, their model focuses on increasing the diversity of output images and are still unable to handle diverse and unseen guiding signals from distinct modalities (e.g. text attributes).

On the other hand, some iterative approaches~\cite{ImageAn,NS,SplitMatch,DeNS} have been  proposed to manipulate image, either patch by patch~\cite{ImageAn,SplitMatch}, or by iteratively updating the input image with hundreds of refinement \cite{NS} to obtain the transformed image. Although these methods require no additional training for each new guiding signal, the iterative evaluation process usually takes tens of seconds even with GPU acceleration \cite{PLoss}, which might be impractical especially for online users.

\begin{figure}[!tb]
\centering
\vskip -0.00in
\includegraphics[scale = 0.13]{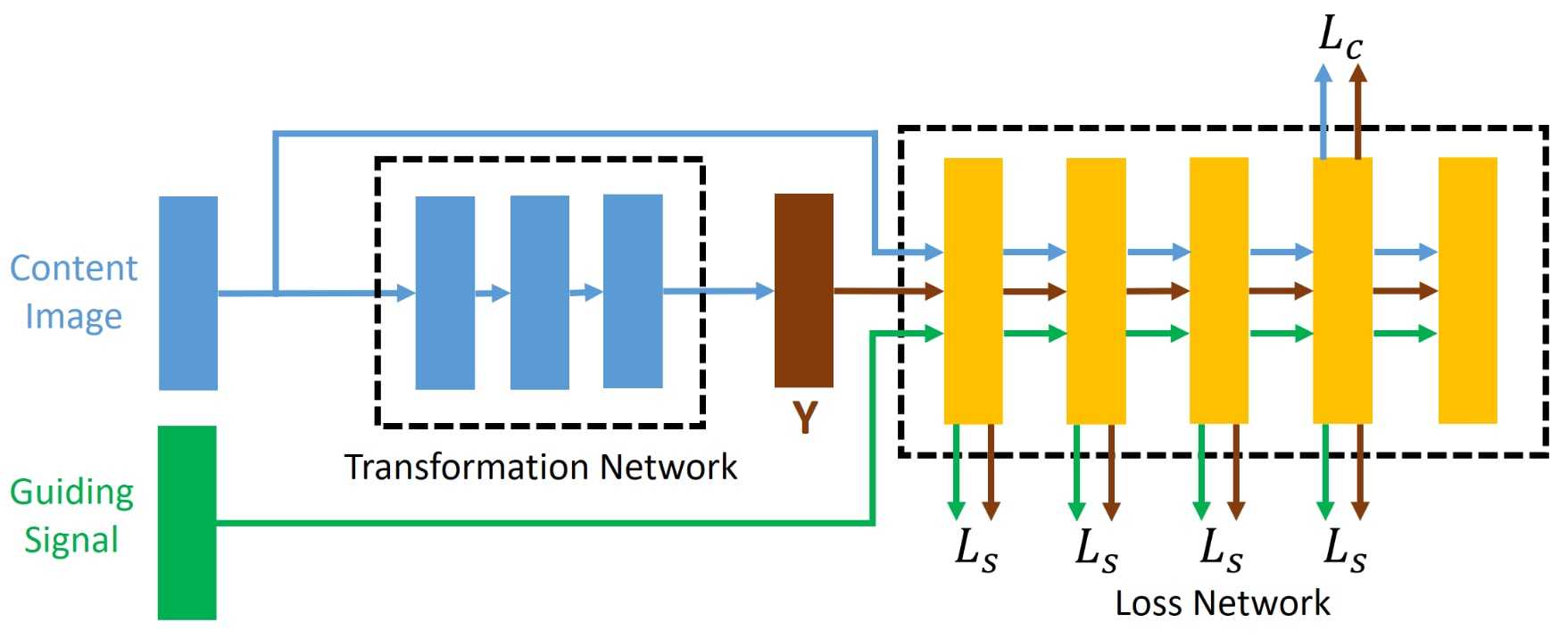}
\vskip -0.2cm
\caption{\small \label{fig:PLoss}An image transformation network with a fixed loss network as described in \cite{PLoss}. For the style transfer task, the guiding signal is a style image. Note that 
one transformation network works for only one style.}
\vskip -0.6cm
\end{figure}

\vspace{-2pt}
\section{Real-time Zero-shot Image Manipulation}
\vspace{-3pt}
In this section we first review the pipelines of current state-of-the-art CNN-based methods, and discuss their limitations in the zero-shot setting. Then, we present the Zero-shot Manipulation Network (ZM-Net), a unified network structure that jointly optimizes a parameter network (PNet) and a image-transformation network (TNet).

\subsection{Image Manipulation with CNNs}
An image manipulation task~\cite{PLoss,NS} can be formally defined as: given a content image $\X_c\in \mathbb{R}^{H \times W \times 3}$ and a guiding signal (e.g. a style image) $\X_s\in \mathbb{R}^{H \times W \times 3}$, output a transformed image $\Y\in \mathbb{R}^{H \times W \times 3}$ such that $\Y$ is similar to $\X_c$ in content and simultaneously similar to $\X_s$ in style. Learning effective representations of content and styles are hence equally essential to perform plausible image manipulation. As in~\cite{NS}, using a fixed deep CNN $\ph(\cdot)$, the feature maps $\ph_l(\X) \in \mathbb{R}^{C_l\times H_l \times W_l}$ in the layer $l$ can represent the content of the image $\X$, and the Gram matrix of $\ph_l(\X)$, denoted as $\G(\ph_l(\X)) \in \mathbb{R}^{C_l \times C_l }$ 
which is computed as
\begin{equation}
\footnotesize
\begin{aligned}
\G(\ph_l(\X))_{c,c'}=\sum\limits_{h=1}^{H_l}\sum\limits_{w=1}^{W_l}\ph_l(\X)_{c,h,w}\ph_l(\X)_{c',h,w}
\end{aligned}
\end{equation}
can express the desired style patterns of the image $\X$.
Two images are assessed to be similar in content or style only if the difference between each corresponding representation (i.e. $\ph_l(\X)$ or $\G(\ph_l(\X))$) has a small Frobenius norm. Therefore, we can train a feedforward image transformation network $\Y=\T(\X_c)$, which is typically a deep CNN, with the loss function: 
\begin{equation}
\footnotesize
\begin{aligned}
\mathcal{L} &= \lambda_s\mathcal{L}_s(\Y)+\lambda_c\mathcal{L}_c(\Y), \\
\mathcal{L}_s(\Y) &= \sum\limits_{l\in \mathcal{S}} \frac{1}{Z_l^2}\|\G(\ph_l(\Y))-\G(\ph_l(\X_s))\|_F^2, \\
\mathcal{L}_c(\Y) &= \sum\limits_{l\in \mathcal{C}} \frac{1}{Z_l}\|\ph_l(\Y)-\ph_l(\X_c)\|_F^2,
\end{aligned}
\end{equation}
where $\mathcal{L}_s(\Y)$ is the style loss for the generated image $\Y$, $\mathcal{L}_c(\Y)$ is the content loss, and $\lambda_s$, $\lambda_c$ are hyperparameters. $\mathcal{S}$ is the set of ``style layers'', $\mathcal{C}$ is the set of ``content layers'', and $Z_l$ is the total number of neurons in layer $l$ \cite{PLoss}. After the transformation network $\T(\cdot)$ is trained, given a new content image $\X'_{c}$, we can generate the stylized image $\Y=\T(\X'_{c})$ without using the loss network. Figure \ref{fig:PLoss} shows an overview of this model. Note that the computation of $\ph_l(\cdot)$ is defined by a \emph{fixed} loss network (e.g. a $16$-layer VGG network~\cite{VGG} pretrained on ImageNet~\cite{ImageNet}) while the transformation network $\T(\cdot)$ is learned given a set of training content images and a style image.
Although performing image manipulation with a single feedforward pass of CNN is usually three orders of magnitude faster than the optimization-based methods in \cite{NS}
, this approach~\cite{PLoss} is largely restricted by that one single transformation network is tied to one specific style image, meaning that $N$ separate networks have to be trained to enable transfer from $N$ style images. The disadvantages are obvious: (1) it is time-consuming to train $N$ separate networks; (2) it needs much more memory to store $N$ networks, which is impractical for mobile devices; (3) it is not scalable and cannot generalize to new styles (a new model needs to be trained for very new incoming styles).

\begin{figure*}[!tb]
\centering
\vskip -0.00in
\includegraphics[height=3.4cm]{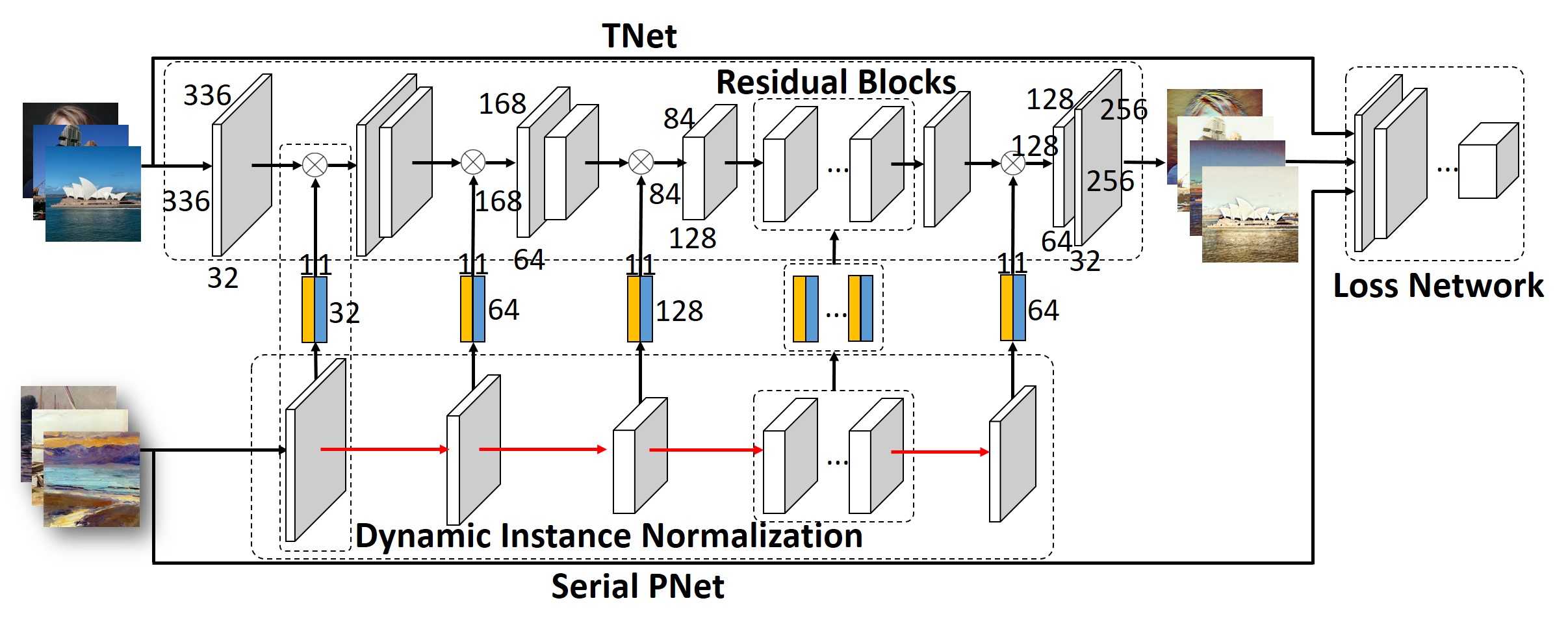}
\includegraphics[height=3.4cm]{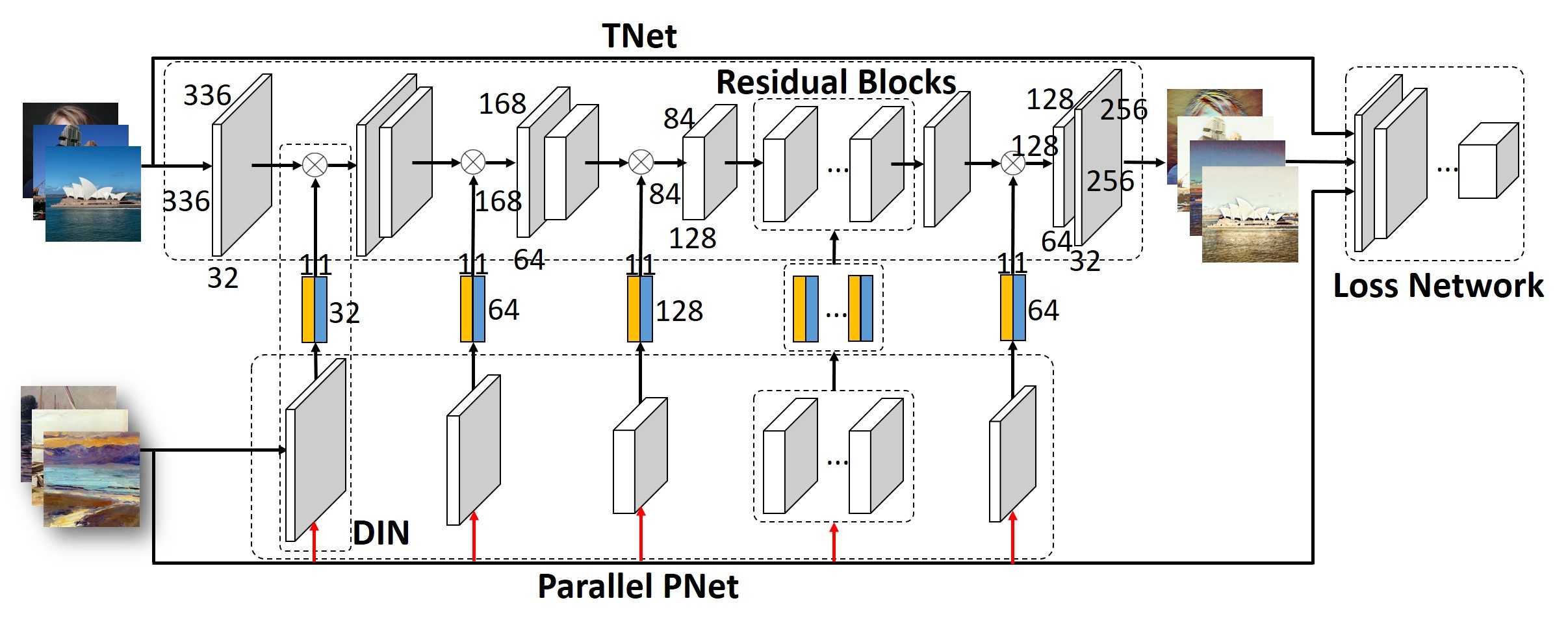}
\vskip -0.5cm
\caption{\small \label{fig:pnet} An overview of the serial architecture (left) and the parallel architecture (right) of our ZM-Net. Details of the loss network are the same as Figure \ref{fig:PLoss} and omitted here.}
\vskip -0.5cm
\end{figure*}


\vspace{-8pt}
\subsection{ZM-Net }\label{sec:fast}
To address the aforementioned problems and enable real-time zero-shot image manipulation, we propose a general architecture, ZM-Net, that combines an image-transformation network (TNet) and a parameter network (PNet). Different from prior works that only adopt a TNet to transform images, we train an extra parameter network (PNet) to produce key parameters of the TNet conditioned on the guiding signals (e.g. style images). As parameters are generated on the fly given arbitrary guiding signals, our ZM-Net avoids training and storing many different network parameters for distinct signals like prior works. 
Moreover, as the PNet learns to embed the guiding signal into a shared space, our ZM-Net is able to perform zero-shot image manipulation given unseen guiding signals.

Here we generalize the notion of style images (in style transfer) to guiding signals (in general image manipulation tasks), i.e. the input $\X_s$ can be any guiding signals beyond style images, for example, word embeddings that express the descriptive attributes in order to impose specific semantics on the input image $\X_c$ (Section \ref{sec:experiments}) or color histograms (a vector representing the pixel color distribution) to guide the colorization of $\X_c$.
In the following, we first present the design of a TNet with our proposed dynamic instance normalization based on \cite{InsNorm}, then introduce a PNet and its variants including the serial PNet and the parallel PNet.

\vspace{-15pt}
\subsubsection{TNet with Dynamic Instance Normalization}
\vspace{-5pt}
To enable zero-shot image manipulation, we must design a principled way to dynamically specify the network parameters of TNet during testing, so that it can handle unseen signals. A naive way would be to directly generate the filters of the TNet, based on feature maps from the PNet conditioning on the guiding signal $\X_s$. 
However, in practice, each layer of TNet typically has over $100{,}000$ parameters (e.g. $128\times 128 \times 3 \times 3$) while feature maps in each layer of PNet usually have about $1{,}000{,}000$ entries (e.g. $128\times 80 \times 80$). 
It is thus difficult to efficiently transform a high dimensional vector to another one. Inspired by \cite{RNS}, we resort to dynamically augmenting the instance normalization (performed after each convolutional layer in TNet) \cite{InsNorm} with the produced scaling and shifting parameters $\ga(\X_s)$ and $\bet(\X_s)$ by PNet. Here the scaling and shifting factors $\ga(\X_s)$ and $\bet(\X_s)$ are treated as key parameters in each layer of TNet. Formally, let $x\in \mathbb{R}^{C_l\times H_l \times W_l}$ be a tensor before instance normalization. $x_{ijk}$ denotes the $ijk$-th element, where $i$ indexes the feature maps and $j,k$ span spatial dimensions. The output $y\in \mathbb{R}^{C_l\times H_l \times W_l}$ of our \emph{dynamic instance normalization} (DIN) is thus computed as (the layer index $l$ is omitted):

\begin{align}
y_{ijk}&=\frac{x_{ijk}-\mu_i}{\sqrt{\sigma_i^2+\epsilon}}\ga_i(\X_s)+\bet_i(\X_s), \label{eq:din} \\
\mu_i&=\frac{1}{HW}\sum\limits_{j=1}^H\sum\limits_{k=1}^W x_{ijk}, \nonumber \\
\sigma_i^2&=\frac{1}{HW}\sum\limits_{j=1}^H \sum\limits_{k=1}^W (x_{ijk}-\mu_i)^2, \nonumber
\end{align}

where $\mu_i$ is the average value in feature map $i$ and $\sigma_i^2$ is the corresponding variance. $\ga_i(\X_s)$ is the $i$-th element of an $C_i$-dimensional vector $\ga(\X_s)$ generated by the PNet and similarly for $\bet_i(\X_s)$. Here if $\ga_i(\X_s)=1$ and $\bet_i(\X_s)=0$, DIN degenerates to the vanilla instance normalization~\cite{InsNorm}. If $\ga_i(\X_s)=\ga_i$ and $\bet_i(\X_s)=\bet_i$, they become directly learnable parameters irrelevant to the PNet. DIN then degenerates to the conditional instance normalization (CIN) in \cite{RNS}. \emph{In both cases, the model loses its ability of zero-shot learning and therefore cannot generalize to unseen signals}.

The PNet that aims to generate $\ga(\X_s)$ and $\bet(\X_s)$ can be a CNN, a multilayer perceptron (MLP), or even a recurrent neural network (RNN). We use a CNN and an MLP as the PNet in Section~\ref{sec:experiments} to demonstrate the generality of our proposed ZM-Net. Since content images and guiding signals are inherently different, the input pair for image manipulation is \emph{non-exchangeable}, making this problem much more difficult than typical problems such as image matching with an exchangeable input image pair. Due to the non-exchangeability, the connection between the TNet and the PNet should be asymmetric. 

%

\subsubsection{Parameter Network (PNet)}\label{sec:serial}
To drive the TNet with dynamic instance normalization, a PNet can have either a serial or a parallel architecture. 

\textbf{Serial PNet.} In a serial PNet, one can use a deep CNN, with a structure similar to the TNet, to generate $\ga^{(l)}(\X_s)$ and $\bet^{(l)}(\X_s)$ in layer $l$. Figure \ref{fig:pnet} (left) shows an overview of this serial architecture. In the serial PNet, $\ga^{(l)}(\X_s)$ and $\bet^{(l)}(\X_s)$ of Equation (\ref{eq:din}) (yellow and blue boxes in Figure \ref{fig:pnet}) are conditioned on the feature maps, denoted as $\ps_l(\X_s)$, in layer $l$ of the PNet.   Specifically,
\begin{align}
\ga^{(l)}(\X_s) &= \ps_l(\X_s)\W_{\gamma}^{(l)}+\b_{\gamma}^{(l)}, \label{eq:gamma} \\
\bet^{(l)}(\X_s) &= \ps_l(\X_s)\W_{\beta}^{(l)}+\b_{\beta}^{(l)} \label{eq:beta}.
\end{align}

Here if the input $\X_s$ is an image, $\ps_l(\X_s)$ can be the output of convolutional layers in the TNet. If the input $\X_s$ is a word embedding (a vector), $\ps_l(\X_s)$ can be the output of fully connected layers. $\W_{\gamma}^{(l)}$, $\b_{\gamma}^{(l)}$, $\W_{\beta}^{(l)}$, and $\b_{\beta}^{(l)}$ are parameters to learn.

\begin{figure*}[t]
    \begin{center}
        \includegraphics[width=0.13\linewidth]{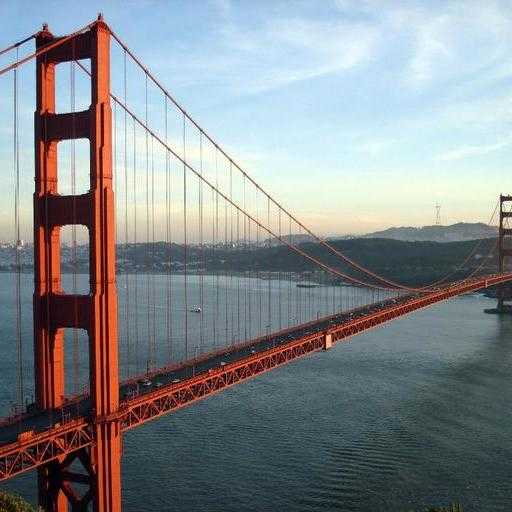}
        \includegraphics[width=0.13\linewidth]{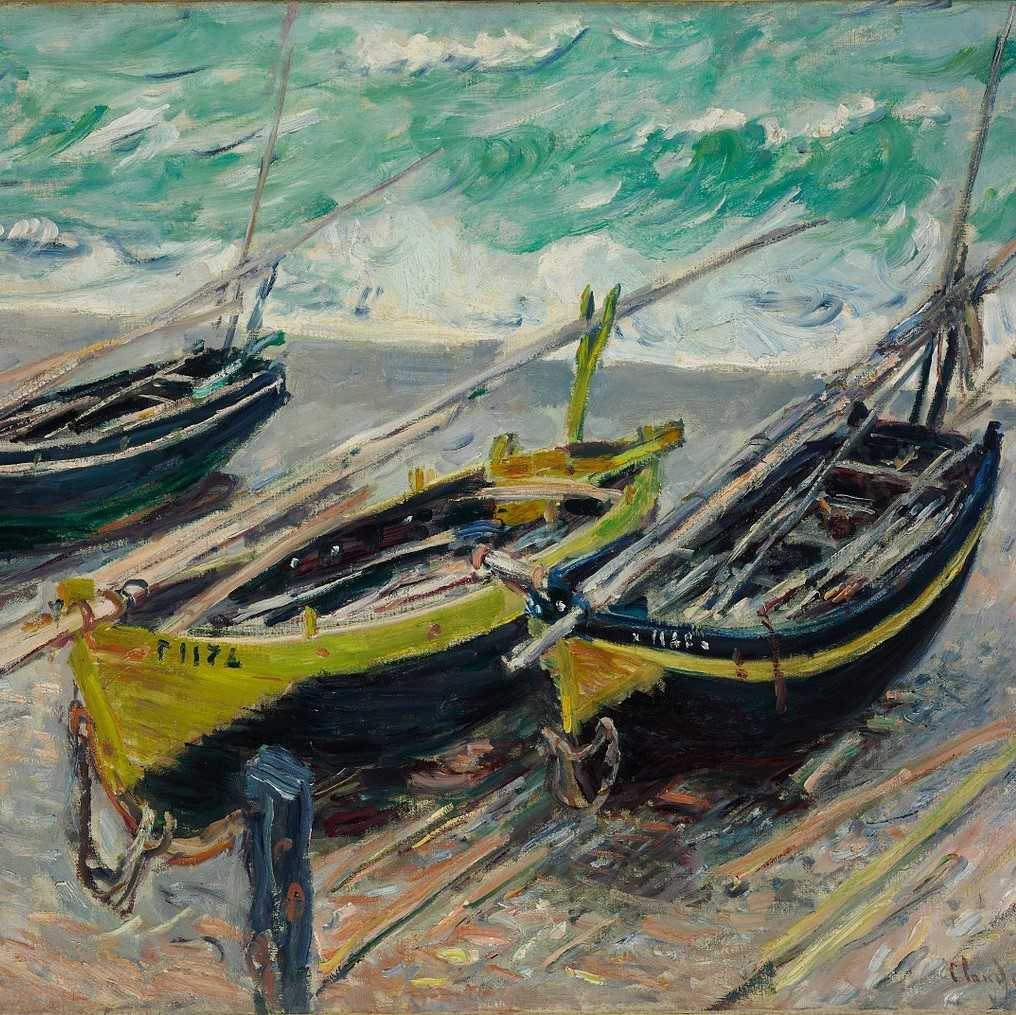}
        \includegraphics[width=0.13\linewidth]{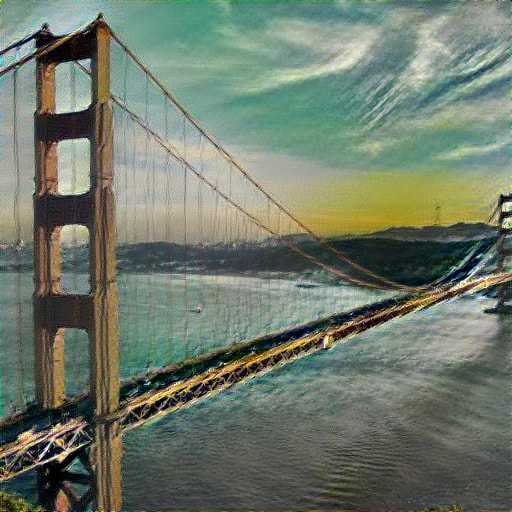}
        \includegraphics[width=0.13\linewidth]{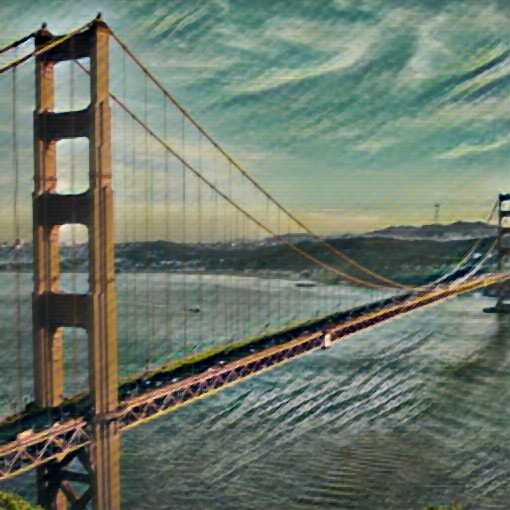}
        \includegraphics[width=0.13\linewidth]{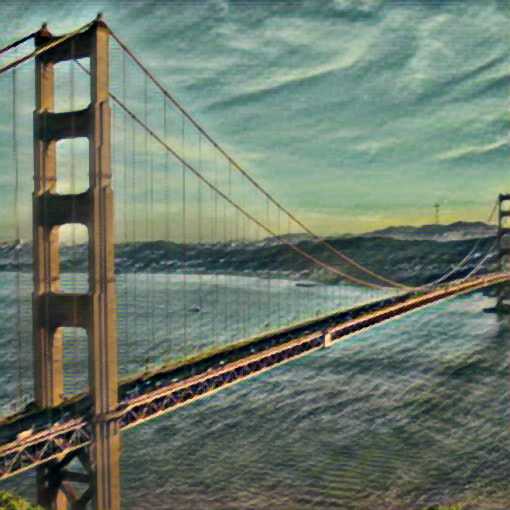}
        \includegraphics[width=0.13\linewidth]{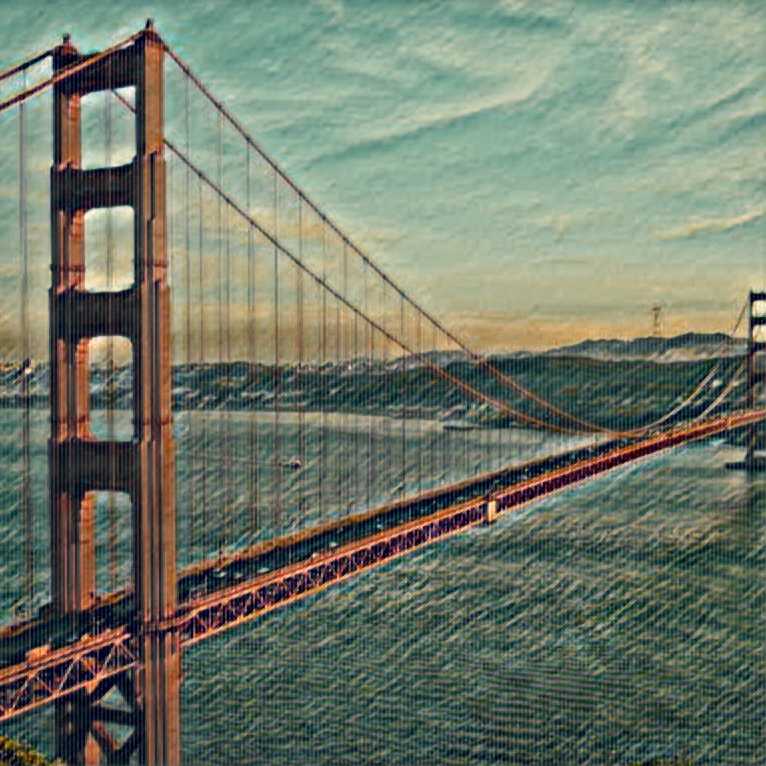} \\
        \includegraphics[width=0.13\linewidth]{fig/blank.jpg}
        \includegraphics[width=0.13\linewidth]{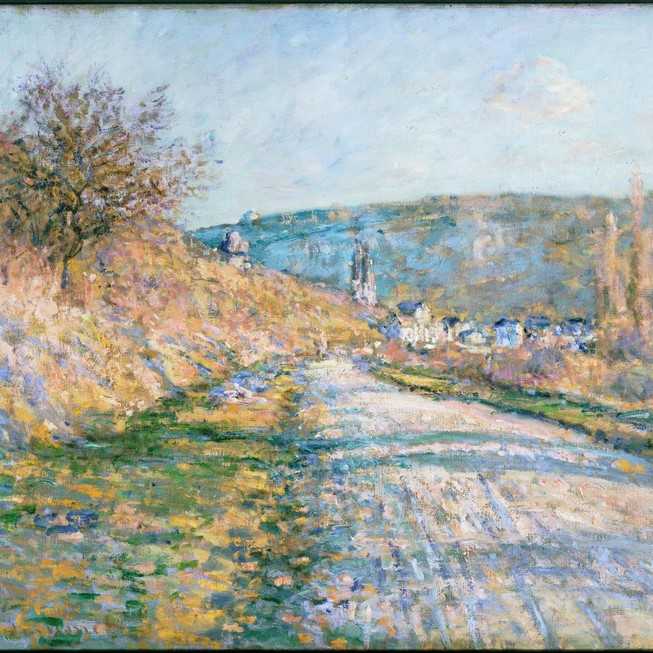}
        \includegraphics[width=0.13\linewidth]{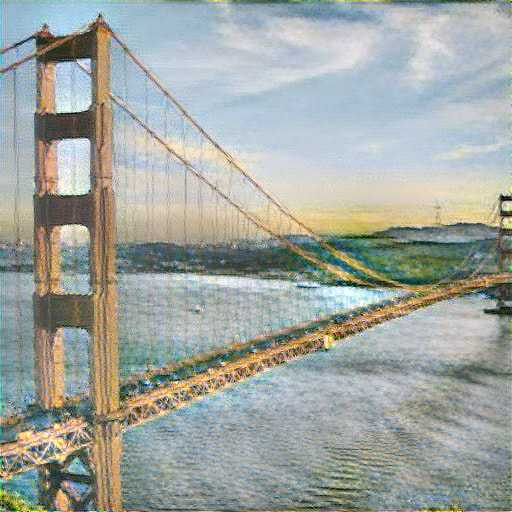}
        \includegraphics[width=0.13\linewidth]{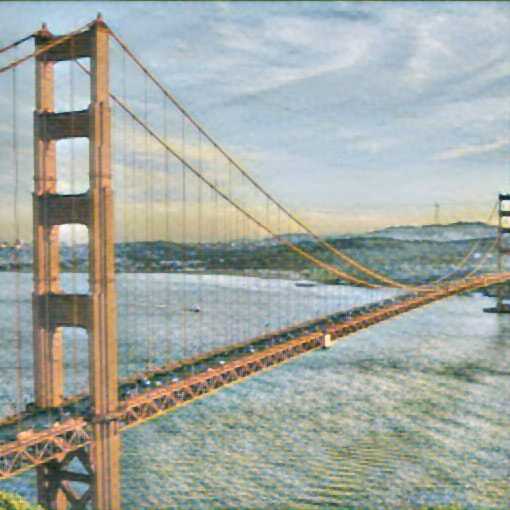}
        \includegraphics[width=0.13\linewidth]{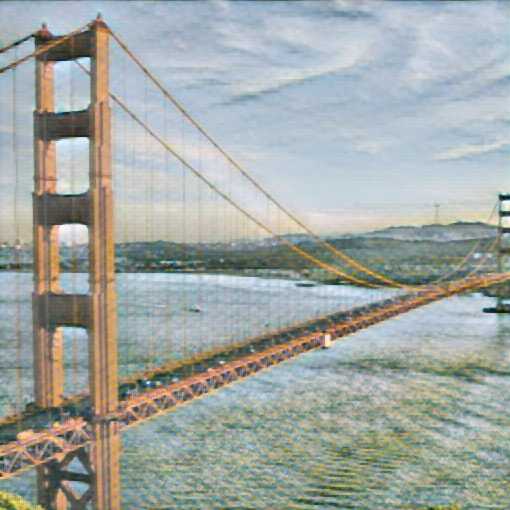}
        \includegraphics[width=0.13\linewidth]{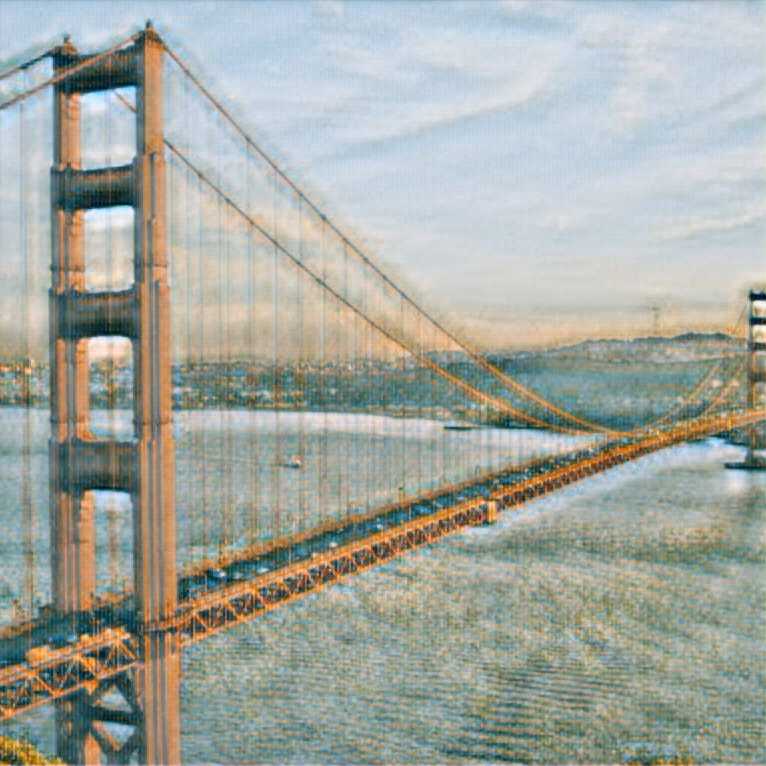}
        \vskip -0.1in
        \caption{\label{fig:capacity_small} \small Results (from Column 3 to 6) of OST \cite{NS}, FST \cite{PLoss}, CIN \cite{RNS}, and a $10$-style ZM-Net. Column 1 is the content image and Column 2 contains $2$ of the $10$ style images used during training. Golden Gate Bridge
            photograph by Rich Niewiroski Jr.}
    \end{center}
    \vskip -0.3in
\end{figure*}

\begin{figure*}[t]
    \begin{center}
        \includegraphics[width=0.13\linewidth]{fig/blank.jpg}
        \includegraphics[width=0.13\linewidth]{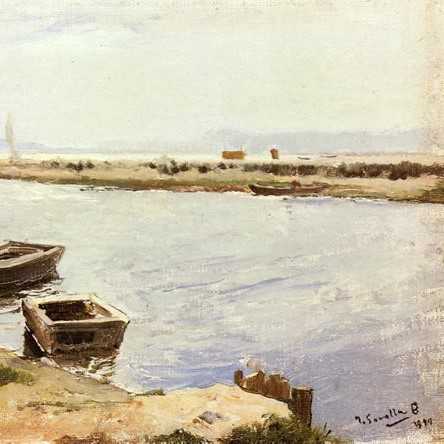}
        \includegraphics[width=0.13\linewidth]{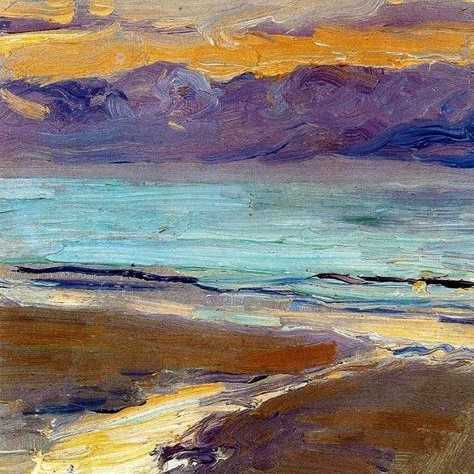}
        \includegraphics[width=0.13\linewidth]{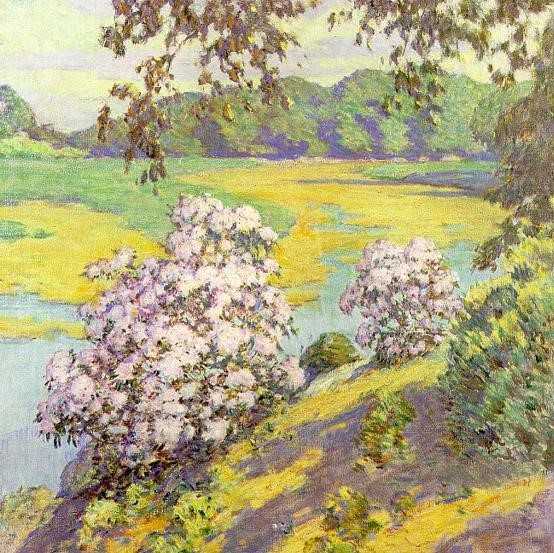}
        \includegraphics[width=0.13\linewidth]{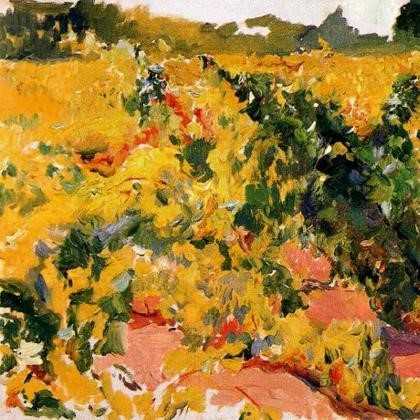}
        \includegraphics[width=0.13\linewidth]{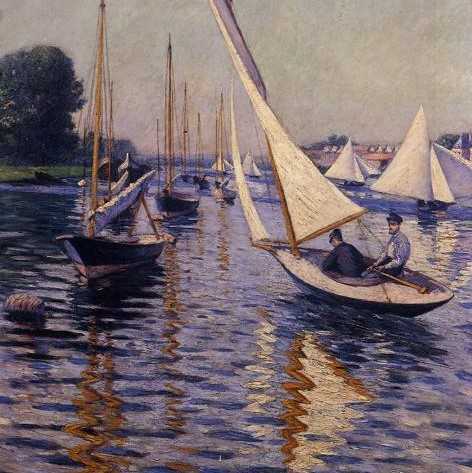} \\
        \includegraphics[width=0.13\linewidth]{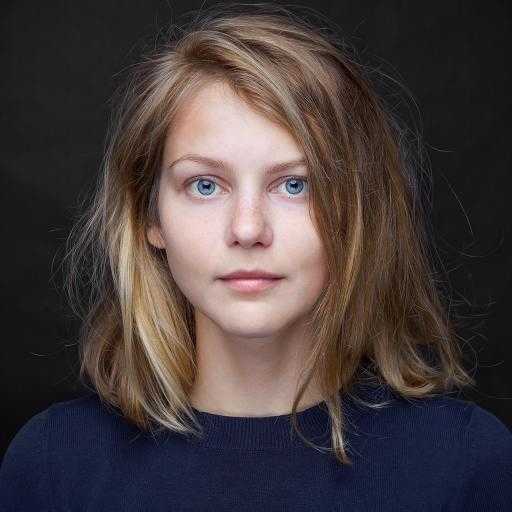}
        \includegraphics[width=0.13\linewidth]{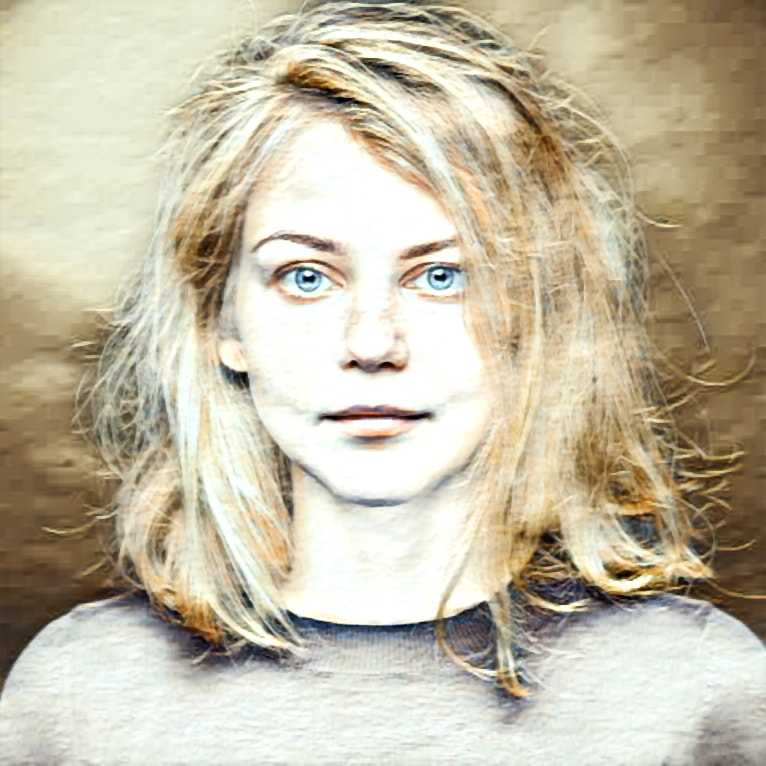}
        \includegraphics[width=0.13\linewidth]{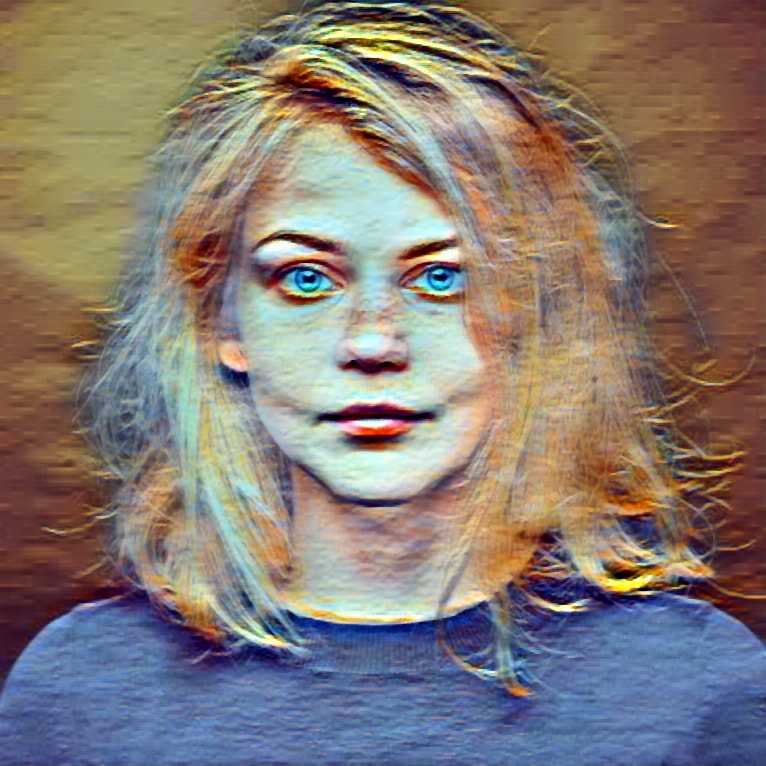}
        \includegraphics[width=0.13\linewidth]{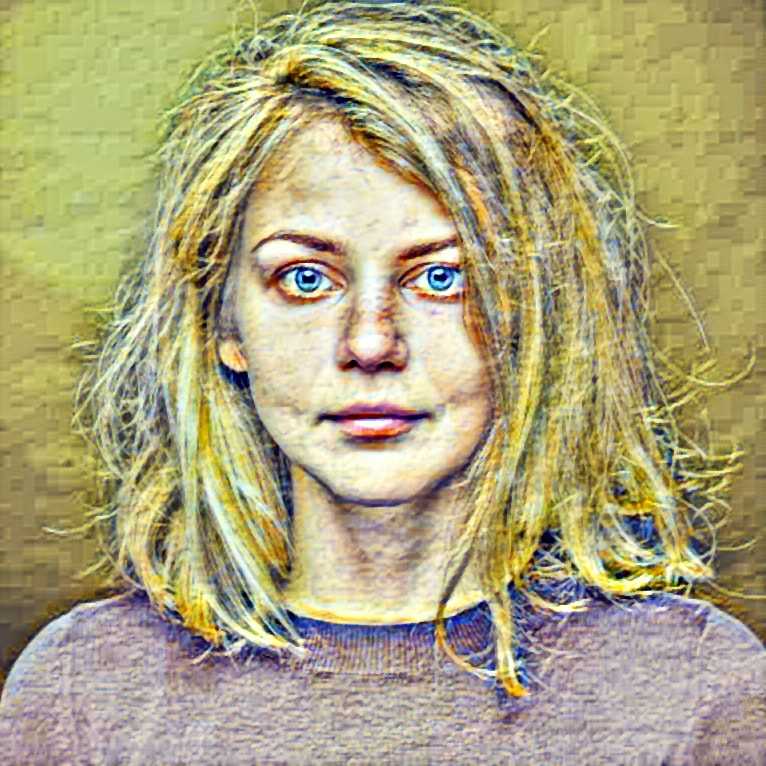}
        \includegraphics[width=0.13\linewidth]{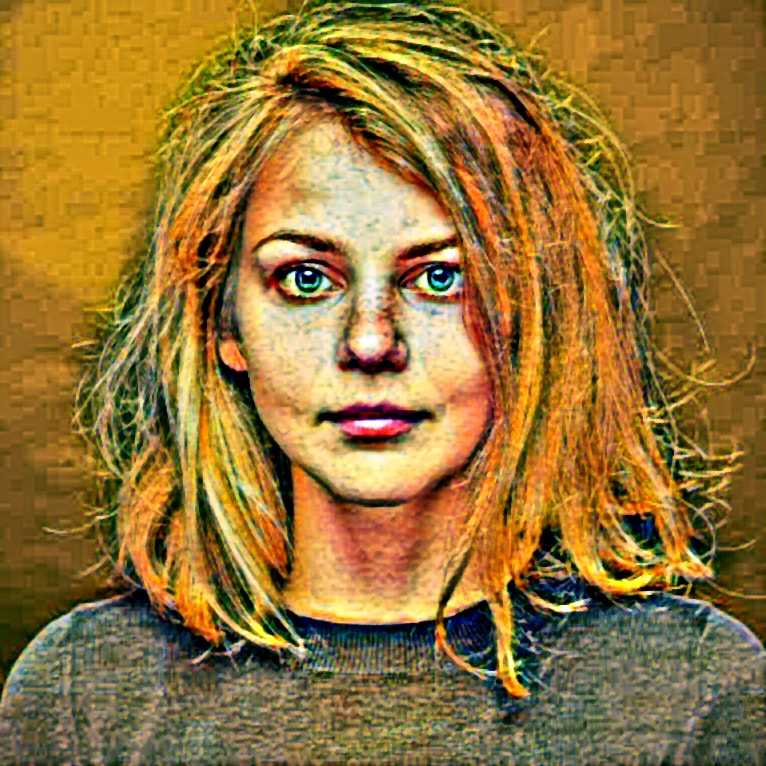}
        \includegraphics[width=0.13\linewidth]{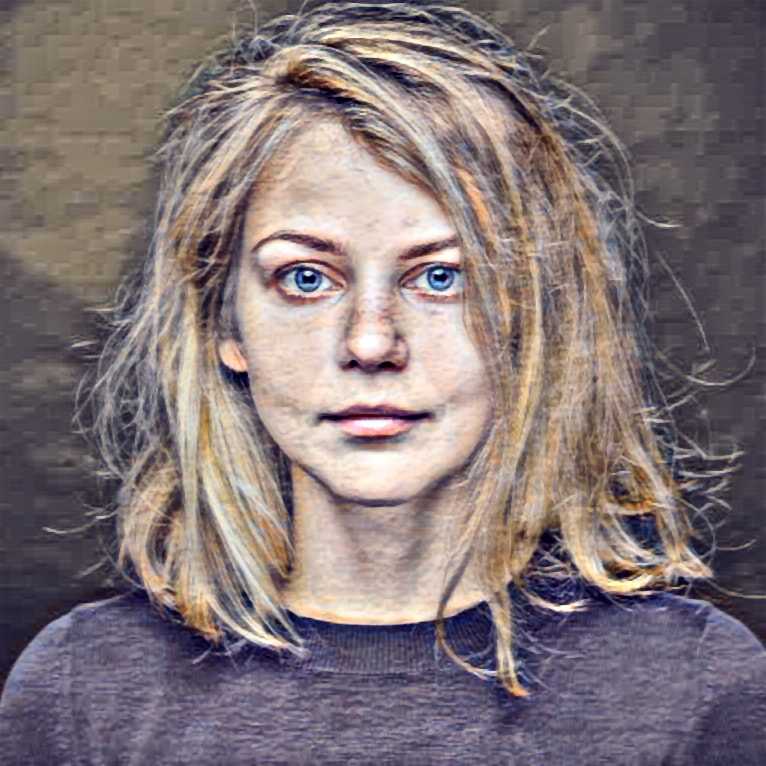}
        \vskip -0.1in
        \caption{\label{fig:capacity_large} \small Results of a $20{,}938$-style ZM-Net. Column 1 is the content image and Column 2 to 6 are randomly selected training style images and corresponding generated images.}
    \end{center}
    \vskip -0.39in
\end{figure*}

Note that in Equation (\ref{eq:din}), $y_{ijk}$ with different $j$ and $k$ share the same $\bet_i(\X_s)$, this design significantly reduces the number of parameters and increases the generalization of the model. Interestingly, if we let $\ga_i(\X_s)=1$ and replace $\bet_i(\X_s)$ with $\bet_{ijk}(\X_s)$, which is computed as the output of a convolutional layer with input $\ps_{l-1}(\X_s)$, followed by the vanilla instance normalization, Equation (\ref{eq:din}) is equivalent to concatenating $\ph_{l-1}(\X_c)$ and $\ps_{l-1}(\X_s)$ followed by a convolutional layer and the vanilla instance normalization, as used in \cite{DNS}. Our preliminary experiments show that although structures similar to \cite{DNS} has sufficient model capacity to perform image manipulation given guiding signals (e.g. style images) in the training set, \emph{it generalizes poorly to unseen guiding signals} and cannot be used for zero-shot image manipulation.

\textbf{Parallel PNet.}
Alternatively, one can use separate shallow networks (either fully connected or convolutional ones) to generate $\ps_l(\X_s)$ in layer $l$, which is then used to compute $\ga^{(l)}(\X_s)$ and $\bet^{(l)}(\X_s)$ according to Equation (\ref{eq:gamma}) and (\ref{eq:beta}). Figure \ref{fig:pnet} (right) shows the architecture of this parallel PNet. Different from the serial PNet where higher levels of $\ga^{(l)}(\X_s)$ and $\bet^{(l)}(\X_s)$ are generated from higher levels of $\ps_l(\X_s)$, here the transformation from $\X_s$ to $\ga^{(l)}(\X_s)$ and $\bet^{(l)}(\X_s)$ follows a shallow and parallel structure. Our experiments (in Section \ref{sec:word}) show that this design would limit the effectiveness of the PNet and slightly decrease the quality of the generated TNet and consequently the generated images $\Y$. Therefore, in Section~\ref{sec:experiments} we use the serial PNet unless otherwise specified. 

\textbf{Training and Test.}
ZM-Net can be trained in an end-to-end manner with the supervision from the loss network, as shown in Figure~\ref{fig:pnet}. During the testing phase, the content image $\X_c$ and the guiding signal $\X_s$ are fed into the TNet and the PNet, respectively, generating the transformed image $\Y$. Note that the loss network is irrelevant during testing.

\section{Experiments}\label{sec:experiments}
In this section, we first demonstrate our ZM-Net's capacity of digesting over $20{,}000$ style images in \emph{one single network} (with a TNet and a PNet), followed by experiments showing the model's ability of zero-shot learning on image manipulation tasks (being able to generalize to \emph{unseen guiding signals}).
As another set of experiments, we also try using simplified word embeddings expressing the descriptive attributes rather than style images as guiding signals to embed specific semantics in content images. We show that with the ability of zero-shot learning and fast image manipulation, our model can generate \emph{animation} of a single image in real-time even though the model is \emph{image-based}. 

\subsection{Fast Zero-shot Style Transfer}
As shown in Table \ref{table:zero_shot}, current methods for fast style transfer \cite{PLoss,TextureNet,MGan} need to train different networks for different styles, costing too much time (several hours for each networks) and memory. Besides, it is also impossible for these methods to generalize to unseen styles (zero-shot style transfer). On the other hand, although the original optimization-based style transfer (OST) method \cite{NS} is capable of zero-shot transfer, it is several orders of magnitude slower than \cite{PLoss,TextureNet,MGan} when generating stylized images. Our ZM-Net is able to get the best of both worlds, performing both fast and zero-shot style transfer.

\begin{table}[tb]
\caption{\small Comparison of optimization-based style transfer \cite{NS}, fast style transfer \cite{PLoss,TextureNet,MGan,RNS}, and our ZM-Net. Note that ZM-Net's time cost per image is up to 0.038s for the first time it processes a new style, and drops to 0.015s after that.}
\label{table:zero_shot}
\begin{center}
\begin{footnotesize}
\vspace{-17pt}
\begin{tabular}{lcccr}
\hline
  & \cite{NS}  & \cite{PLoss,TextureNet,MGan,RNS} & ZM-Net \\
\hline
Speed         &15.86s  &0.015s &0.015s$\sim$0.038s \\
Zero-shot               &\checkmark  &\text{\sffamily X} &\checkmark \\
\hline
\end{tabular}
\vspace{-12pt}
\end{footnotesize}
\end{center}
\vskip -0.2in
\end{table}

\textbf{Datasets.}
We use the MS-COCO dataset \cite{COCO} as our content images. In order for ZM-Net to generalize well to unseen styles, the style images in the training set need sufficient diversity to prevent the model from overfitting to just a few styles. Unfortunately, unlike photos that can be massively produced, art work such as paintings (especially famous ones) is rare and difficult to collect. To address this problem, we use the $984$ impressionism paintings in the dataset \emph{Pandora} \cite{Pandora} as seed style images to produce a larger dataset of $23{,}307$ style images. Specifically, we first split the $984$ images into $784$ training images, $100$ validation images (for choosing hyperparameters), and $100$ testing images. We then randomly select a content image and an impressionism painting from one of the three sets as input to OST \cite{NS}, producing a new style image with a similar style but different content. \emph{Note that different from traditional dataset expansion, our expansion process can introduce much more content diversity to the dataset and hence prevent the training process from overfitting the content of the style images}. Our experiments show that using the expanded dataset rather than the original one can cut the testing loss $\mathcal{L}$ nearly by half (from $58342.2$ to $31860.4$).

\begin{figure*}[t]
    \begin{center}
        \includegraphics[width=0.13\linewidth]{fig/golden.jpg}
        \includegraphics[width=0.13\linewidth]{fig/m1s.jpg}
        \includegraphics[width=0.13\linewidth]{fig/m3s.jpg}
        \includegraphics[width=0.13\linewidth]{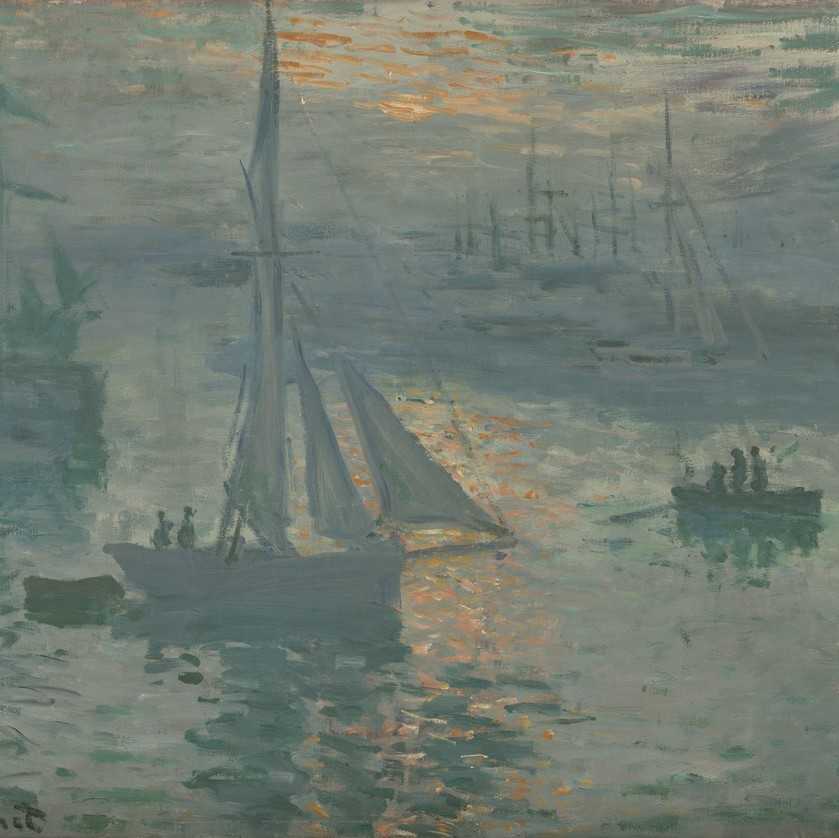}
        \includegraphics[width=0.13\linewidth]{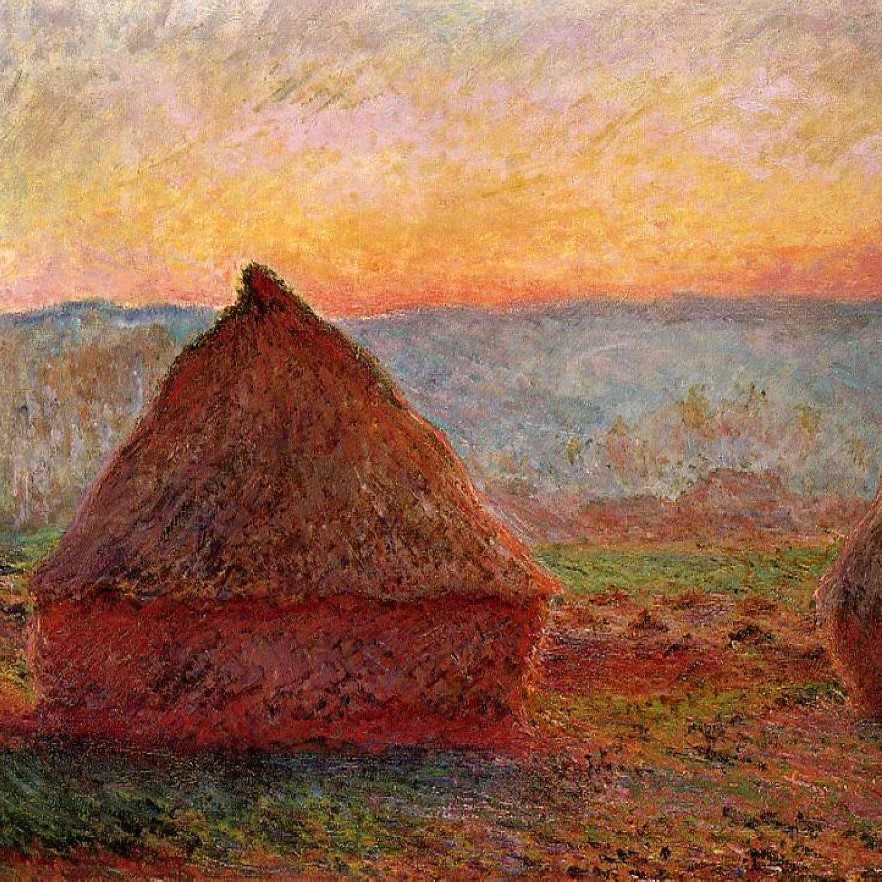}
        \includegraphics[width=0.13\linewidth]{fig/m11s.jpg}
        \includegraphics[width=0.13\linewidth]{fig/m12s.jpg} \\
        \includegraphics[width=0.13\linewidth]{fig/blank.jpg}
        \includegraphics[width=0.13\linewidth]{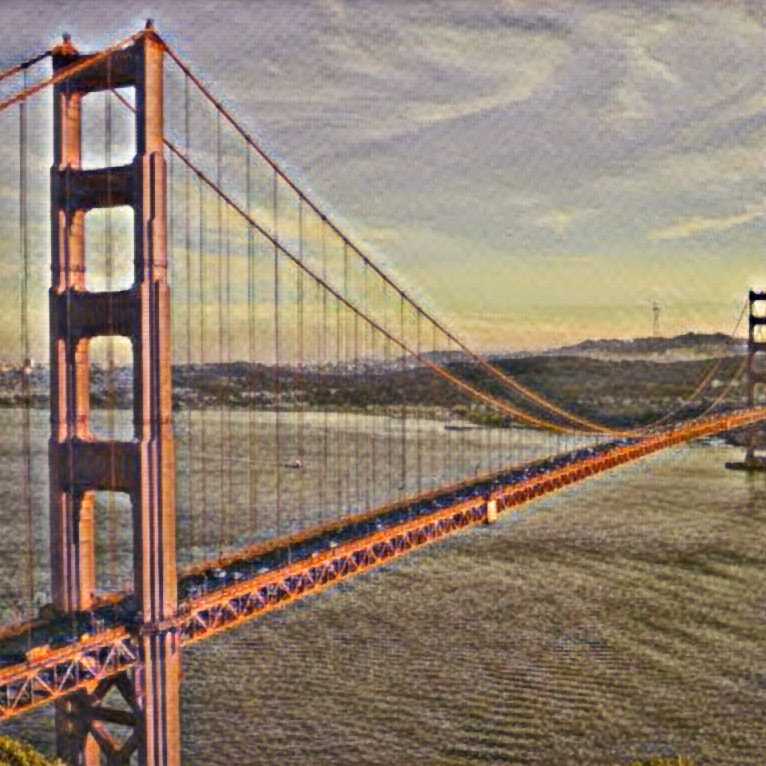}
        \includegraphics[width=0.13\linewidth]{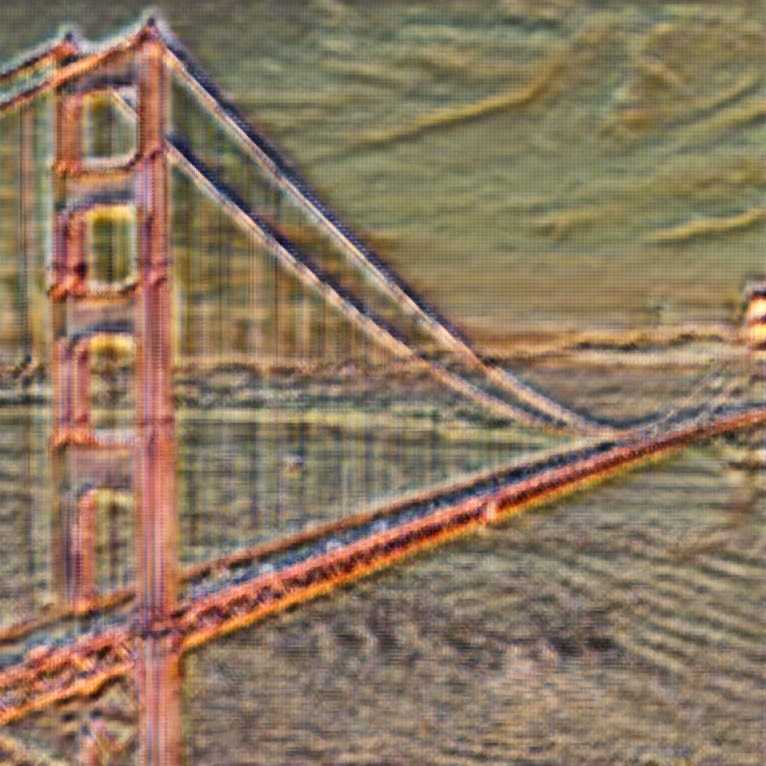}
        \includegraphics[width=0.13\linewidth]{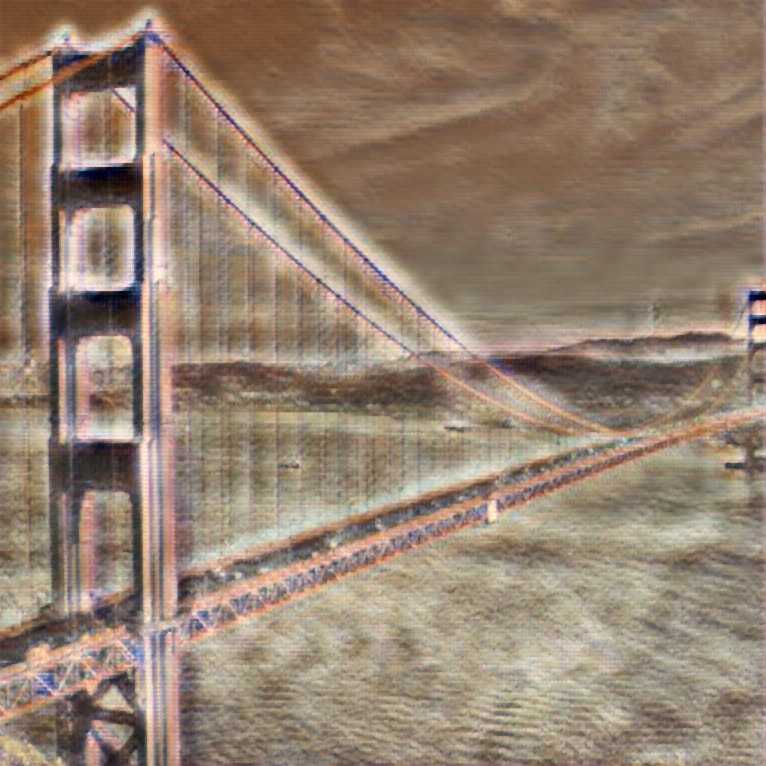}
        \includegraphics[width=0.13\linewidth]{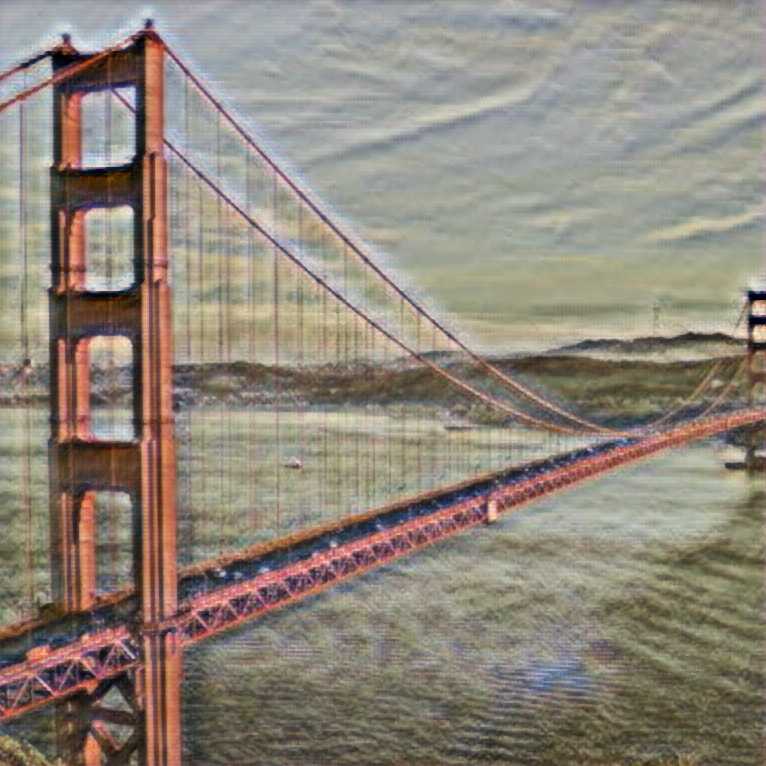}
        \includegraphics[width=0.13\linewidth]{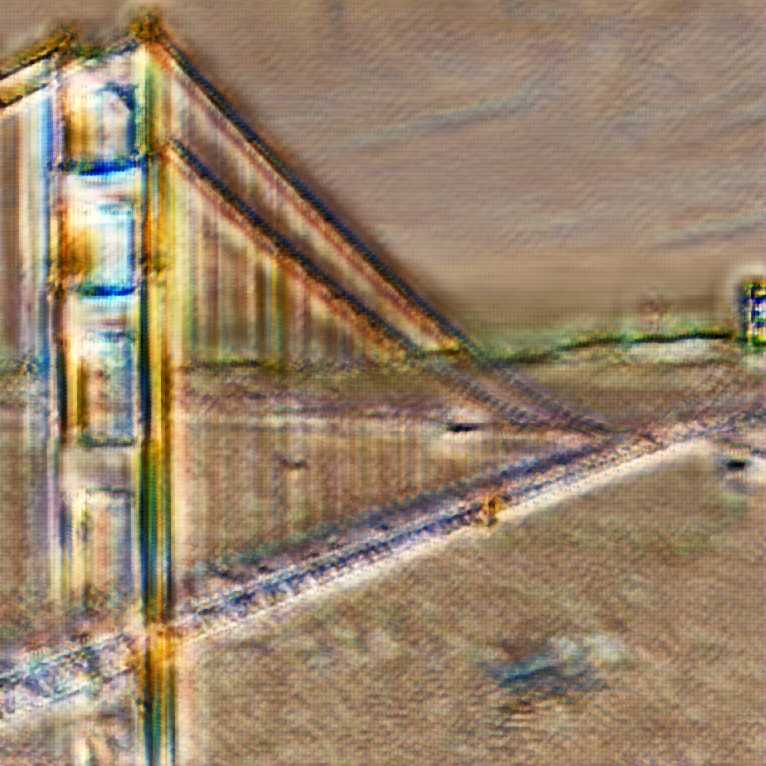}
        \includegraphics[width=0.13\linewidth]{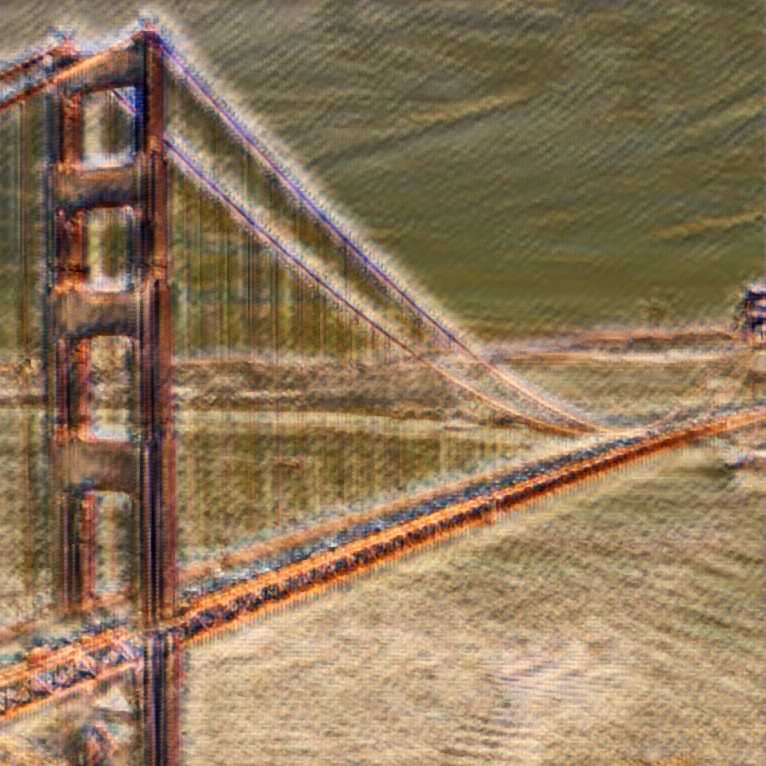} \\
        \includegraphics[width=0.13\linewidth]{fig/blank.jpg}
        \includegraphics[width=0.13\linewidth]{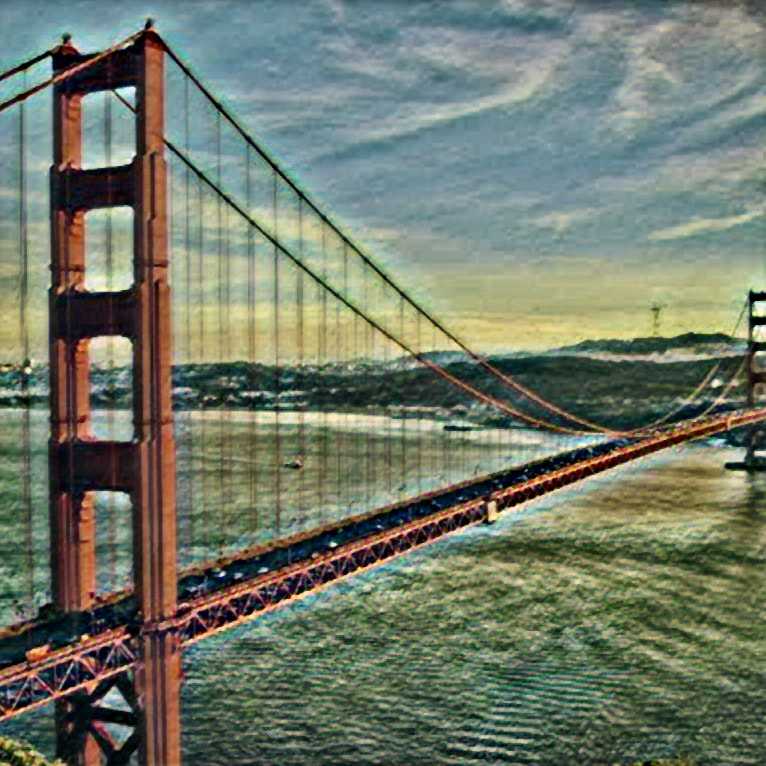}
        \includegraphics[width=0.13\linewidth]{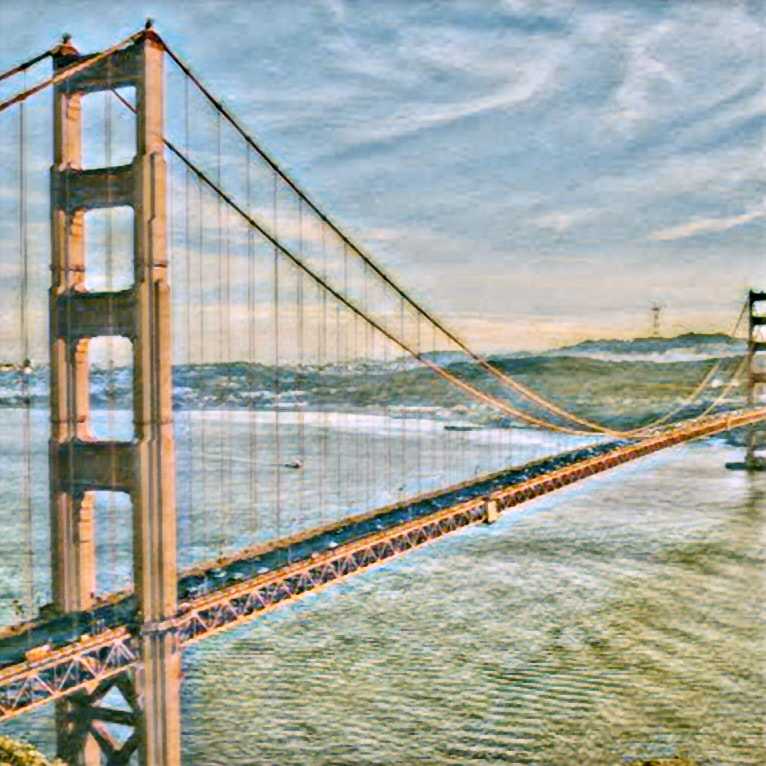}
        \includegraphics[width=0.13\linewidth]{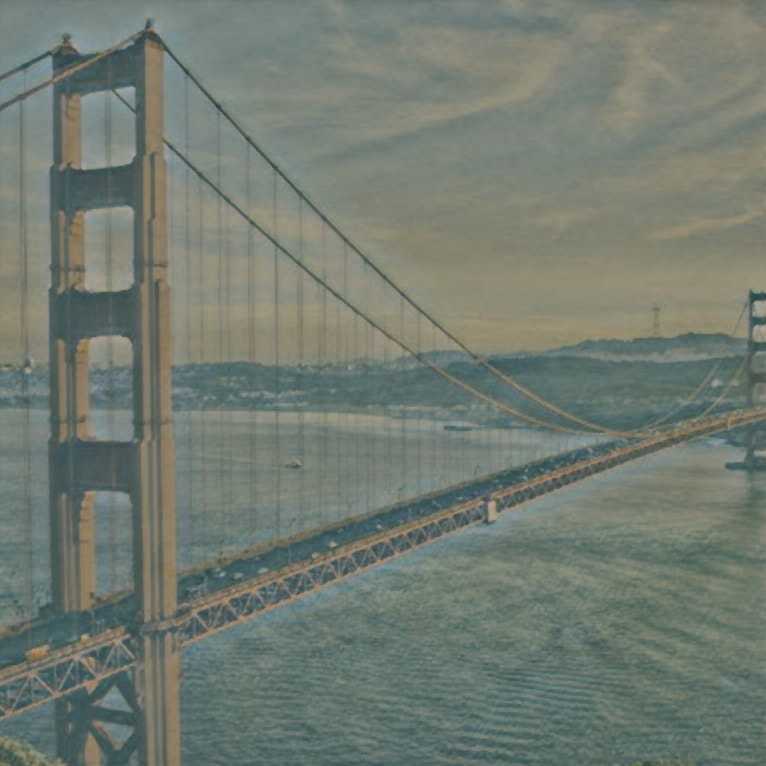}
        \includegraphics[width=0.13\linewidth]{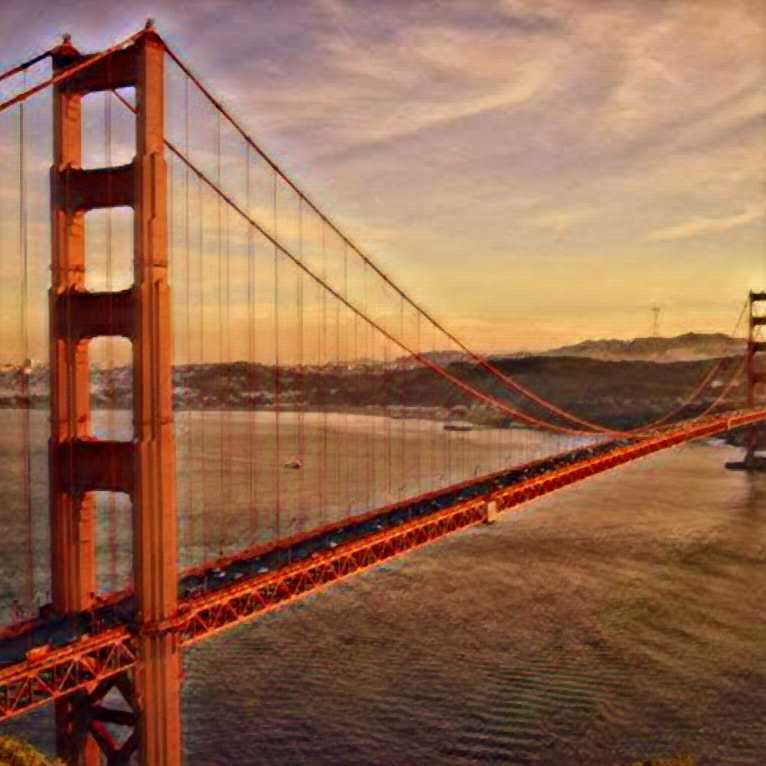}
        \includegraphics[width=0.13\linewidth]{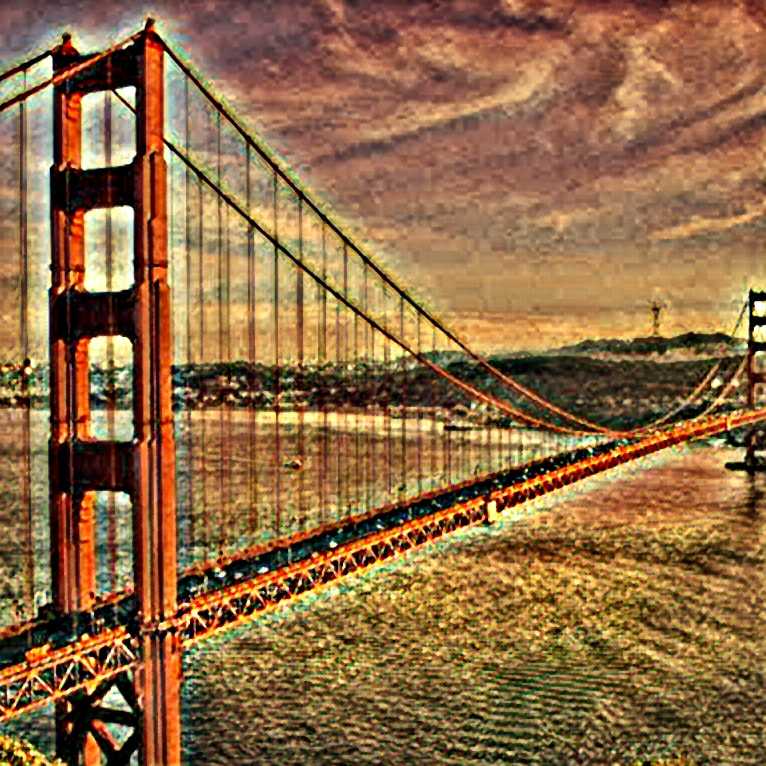}
        \includegraphics[width=0.13\linewidth]{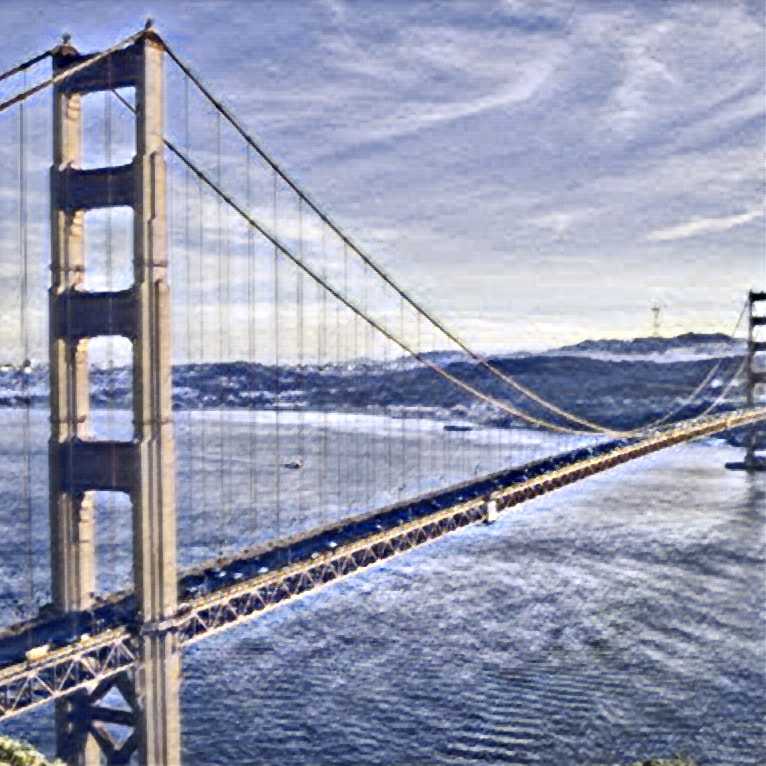} \\
        \includegraphics[width=0.13\linewidth]{fig/blank.jpg}
        \includegraphics[width=0.13\linewidth]{fig/golden-m1-justin.jpg}
        \includegraphics[width=0.13\linewidth]{fig/golden-m1-justin.jpg}
        \includegraphics[width=0.13\linewidth]{fig/golden-m1-justin.jpg}
        \includegraphics[width=0.13\linewidth]{fig/golden-m1-justin.jpg}
        \includegraphics[width=0.13\linewidth]{fig/golden-m1-justin.jpg}
        \includegraphics[width=0.13\linewidth]{fig/golden-m1-justin.jpg}
        \vskip -0.10in
        \caption{\label{fig:zero} \small Fast zero-shot style transfer results (from Row 2 to 4) using our $10$-style ZM-Net, $20{,}938$-style ZM-Net, and FST \cite{PLoss}. Row 1 shows the content image and the style images.}
    \end{center}
    \vskip -0.32in
\end{figure*}

\textbf{Experimental Settings.}
For the baselines, OST \cite{NS}, fast style transfer (FST) \cite{PLoss}, and CIN \cite{RNS}, we use the same network structures and hyperparameters mentioned in the papers. For our ZM-Net, we follow the network structure from \cite{PLoss} (with residual connections) for both the TNet and the PNet, except for the part connecting to DIN. We use a serial PNet for the style transfer task. As in \cite{PLoss,RNS}, we use the VGG-16 loss network  with the same content and style layers. All models are trained with a minibatch size of $4$ for $40{,}000$ iterations using Adam \cite{Adam} (for efficiency, content images in the same minibatch share the same style image). As an exception, we train the $20{,}938$-style ZM-Net for $160{,}000$ iterations with an initial learning rate of $1\times 10^{-3}$ and decay it by $0.1$ every $40{,}000$ iterations.

\textbf{Model Capacity.}
To show that ZM-Net has enough model capacity to digest multiple styles with one single network, we train ZM-Net with up to $20{,}938$ style images and evaluate its ability to stylize \emph{new content images} with \emph{style images in the training set}. Figure \ref{fig:capacity_small} shows the results of a $10$-style ZM-Net (the last column), OST \cite{NS}, FST \cite{PLoss}, and CIN \cite{RNS} (see the supplementary material for more results). Note that both FST and CIN need to train different networks for different style images\footnote{Although CIN can share parameters of convolutional layers across different styles, the other parts of the parameters still need to be trained separately for different styles.} while ZM-Net can be simultaneously trained on multiple styles with a single network. As we can see, ZM-Net can achieve comparable performance with one single network. Similarly, Figure \ref{fig:capacity_large} shows the results of a $20{,}938$-style ZM-Net. Surprisingly, ZM-Net has no problem digesting as many as $20{,}938$ with only one network either. Quantitatively, the final training loss (average over the last 100 iterations) of the $20{,}938$-style ZM-Net is very close to that of CIN \cite{RNS} ($157382.7$ versus $148374.3$), which again demonstrates ZM-Net's sufficient model capacity.

\begin{figure*}[t]
    \begin{center}
        \includegraphics[width=0.13\linewidth]{fig/m1s.jpg}
        \includegraphics[width=0.13\linewidth]{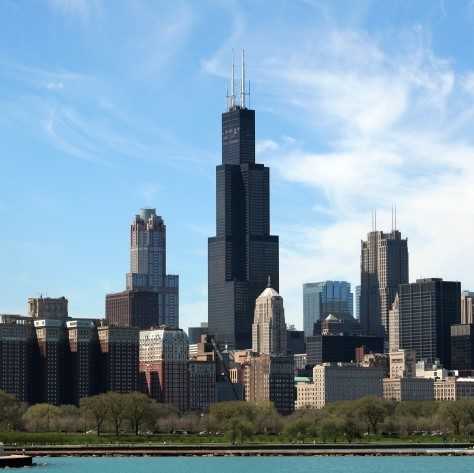}
        \includegraphics[width=0.13\linewidth]{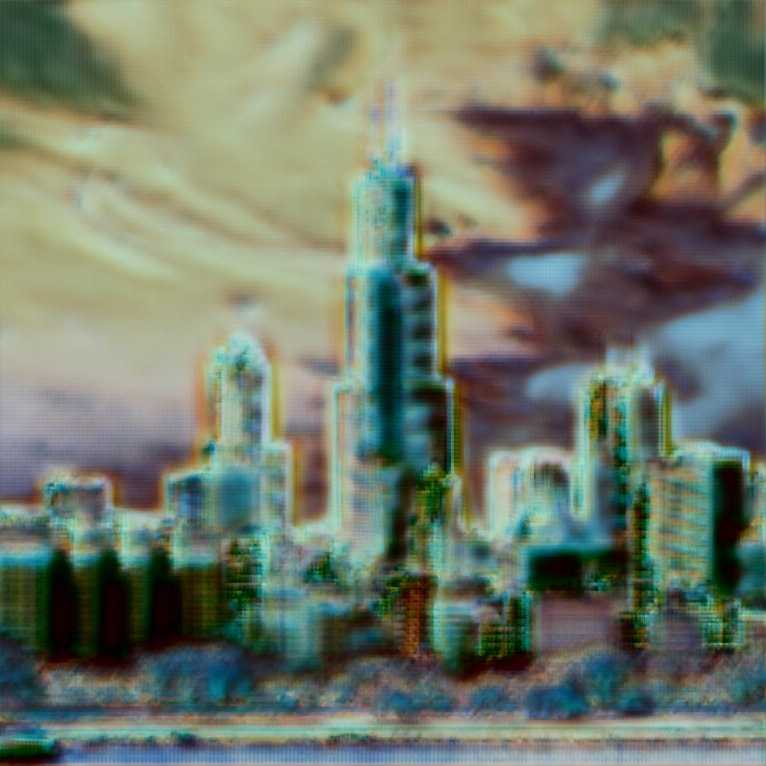}
        \includegraphics[width=0.13\linewidth]{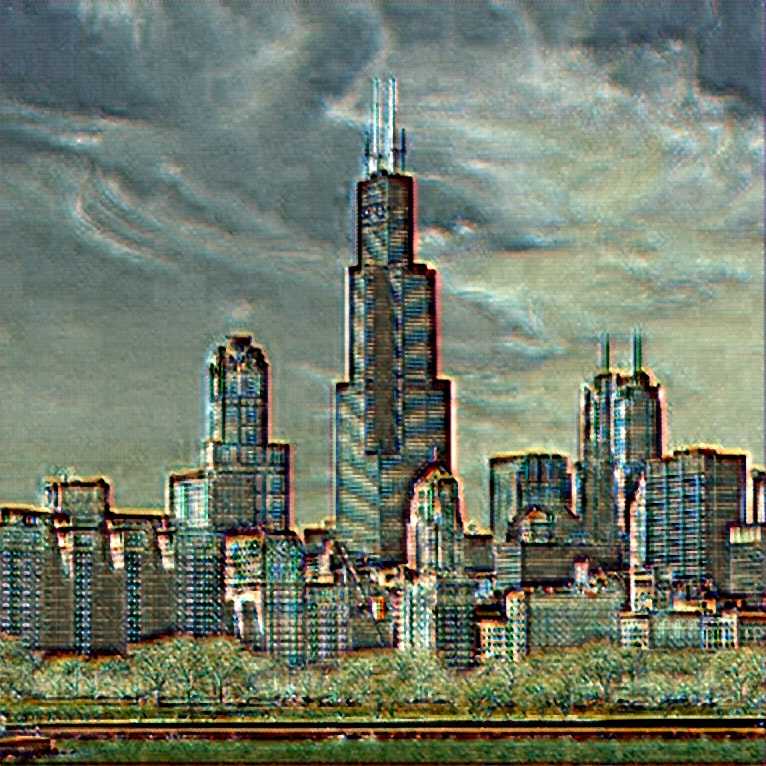}
        \includegraphics[width=0.13\linewidth]{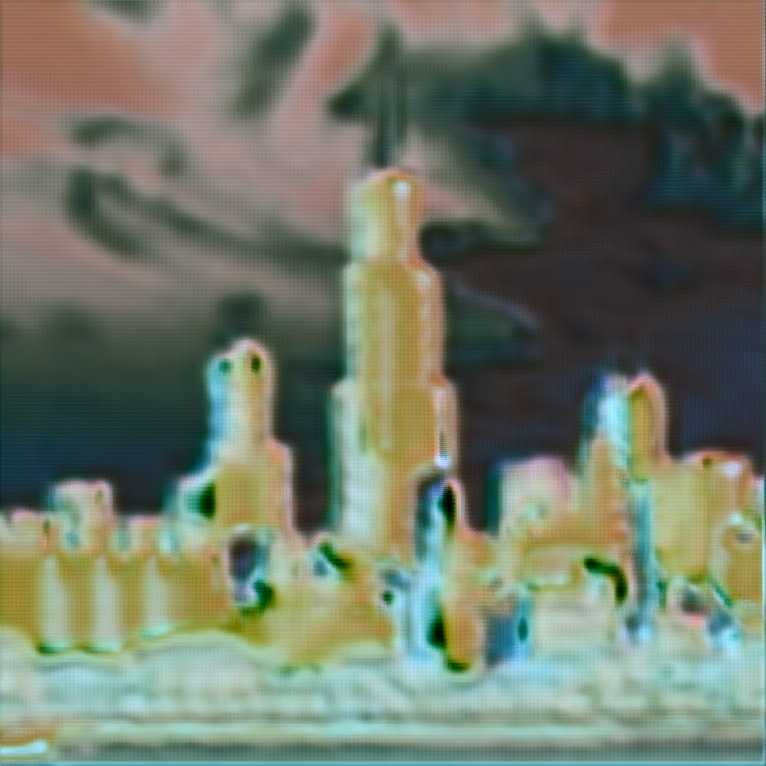}
        \includegraphics[width=0.13\linewidth]{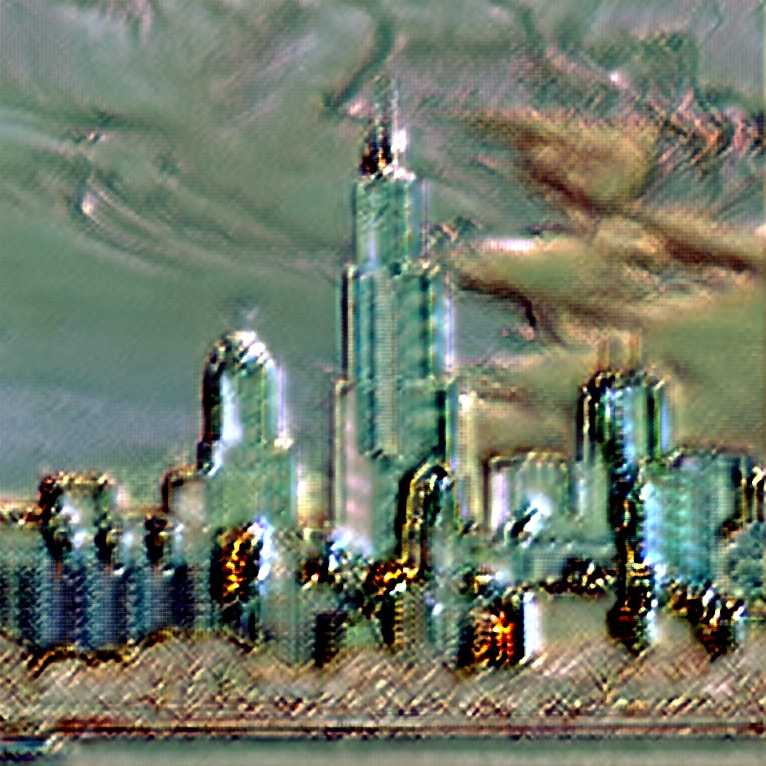}
        \includegraphics[width=0.13\linewidth]{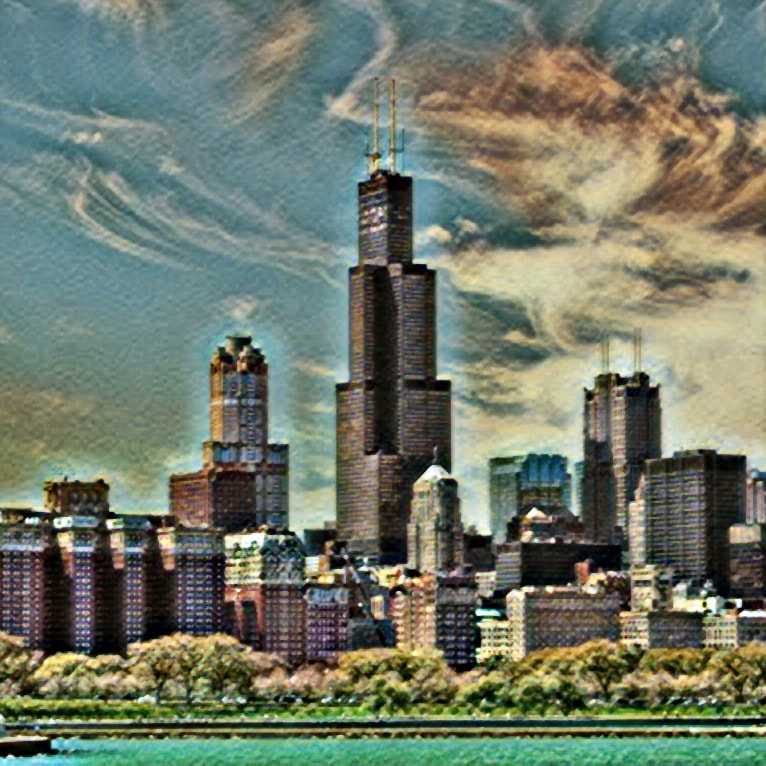} \\
        \includegraphics[width=0.13\linewidth]{fig/blank.jpg}
        \includegraphics[width=0.13\linewidth]{fig/golden.jpg}
        \includegraphics[width=0.13\linewidth]{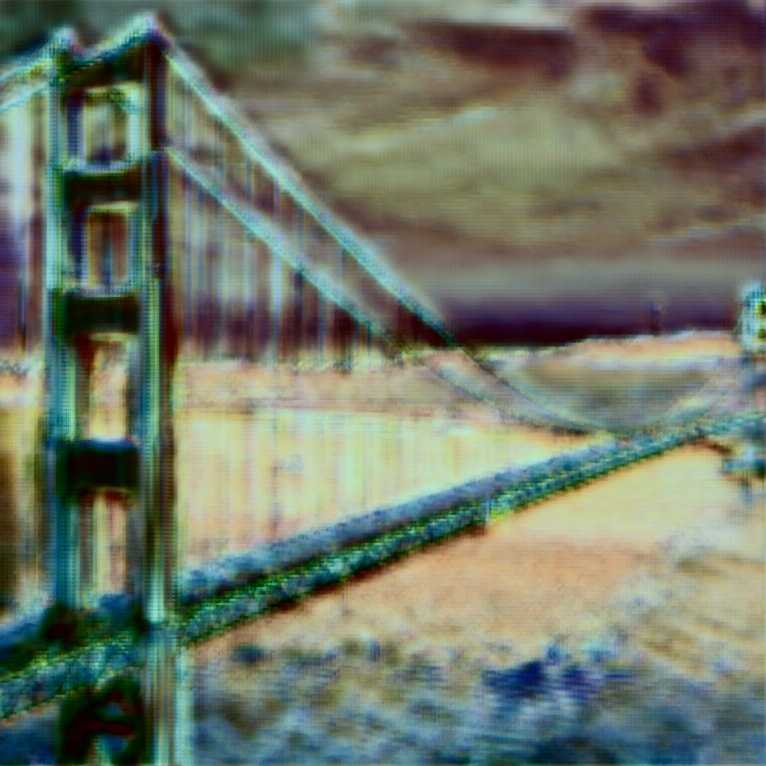}
        \includegraphics[width=0.13\linewidth]{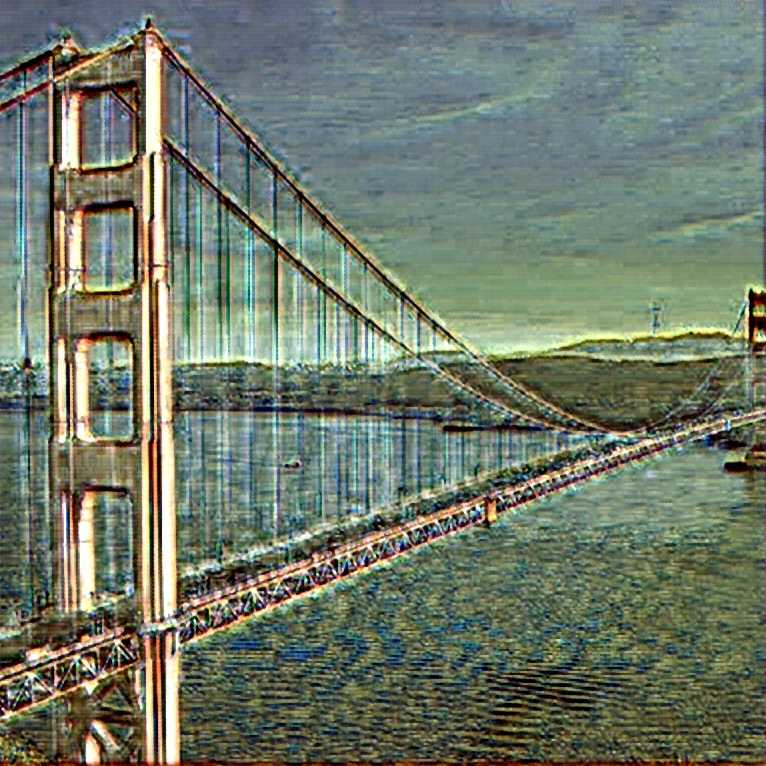}
        \includegraphics[width=0.13\linewidth]{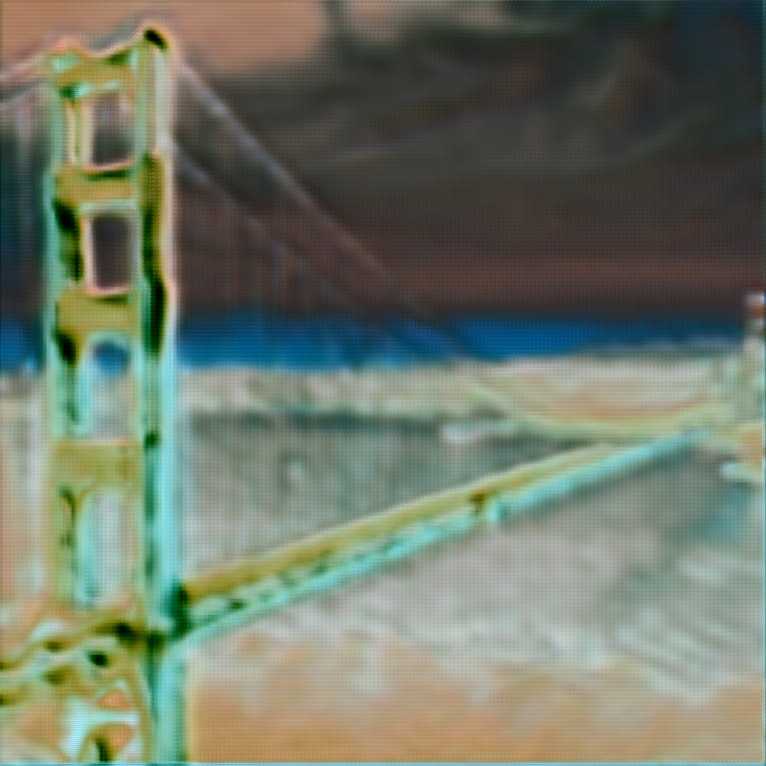}
        \includegraphics[width=0.13\linewidth]{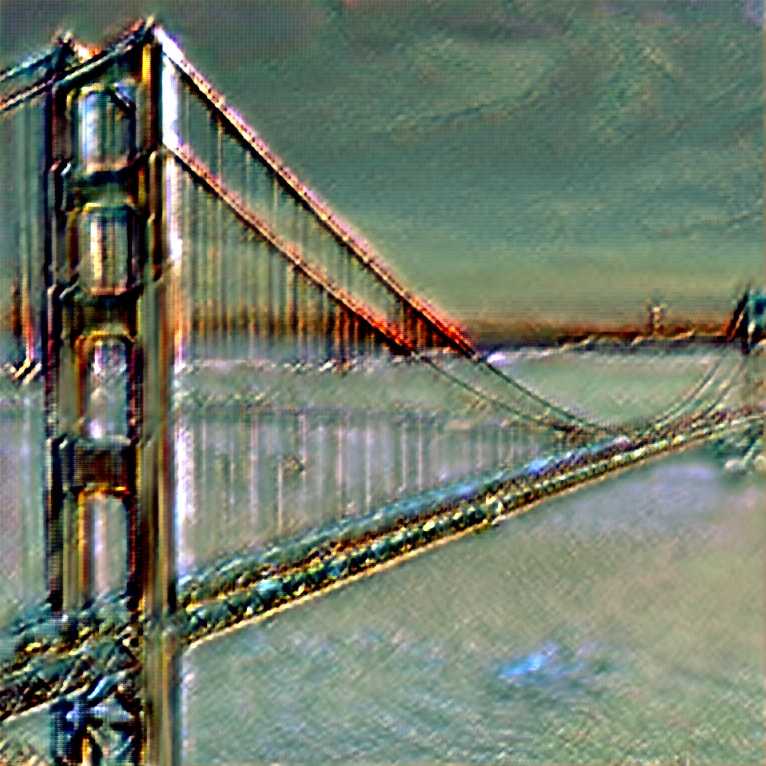}
        \includegraphics[width=0.13\linewidth]{fig/golden-m1-0s.jpg} \\
        \vskip -0.1in
        \caption{\label{fig:layers} \small Zero-shot style transfer using a $20{,}938$-style ZM-Net with DIN in some layers turned on. Column 1: Style image. Column 2: Content images. Column 3: DIN in all layers is off. Column 4 to 6: DIN in layer $1\sim 3$, $4\sim 6$, and $7\sim 9$ is on, respectively. Column 7: DIN in all layers is on. }
    \end{center}
    \vskip -0.39in
\end{figure*}

\begin{figure}[!tb]
\begin{center}
\vskip -0.05in
\includegraphics[width=0.33\linewidth]{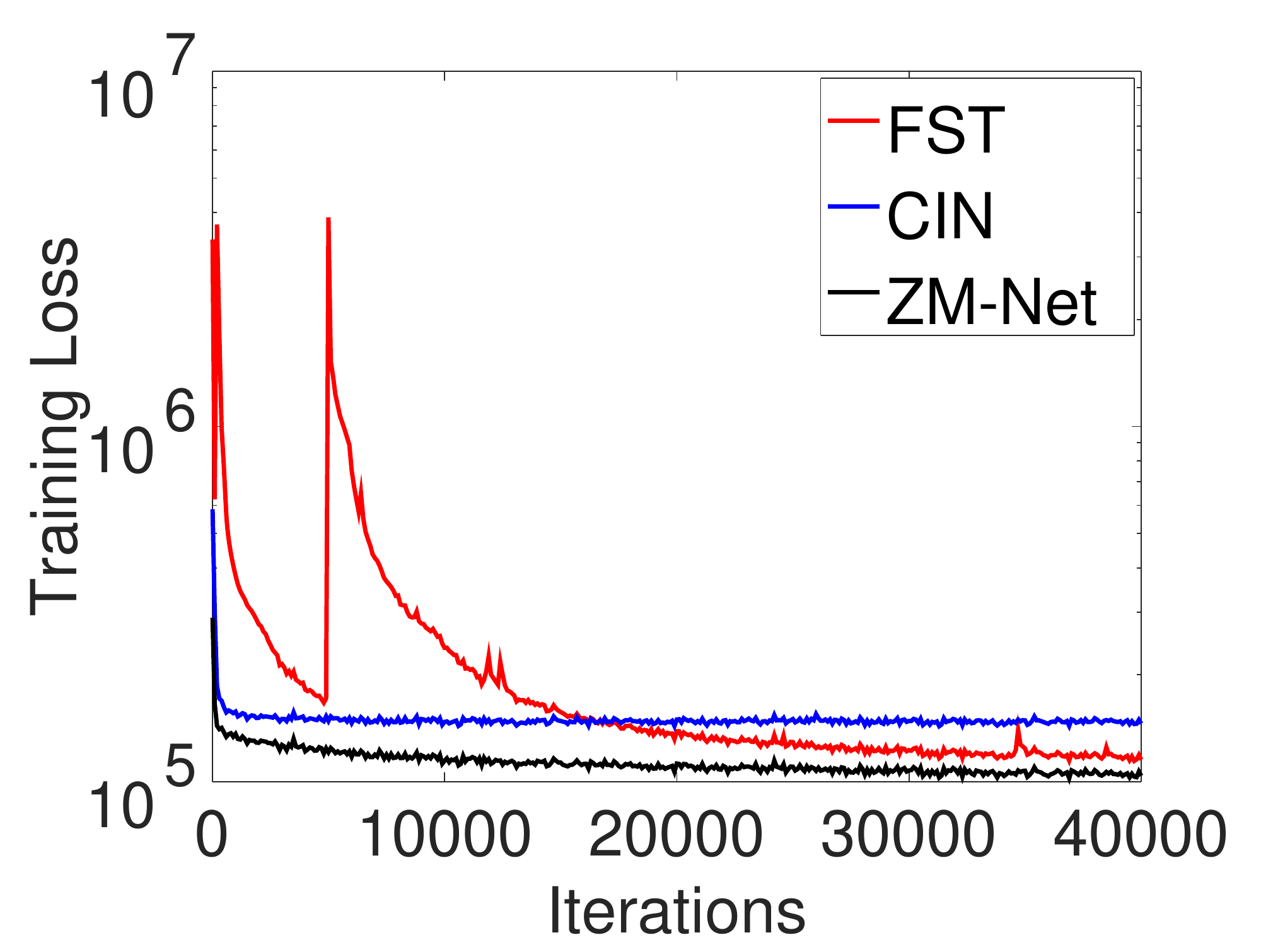}
\hspace{-0.15in}
\includegraphics[width=0.33\linewidth]{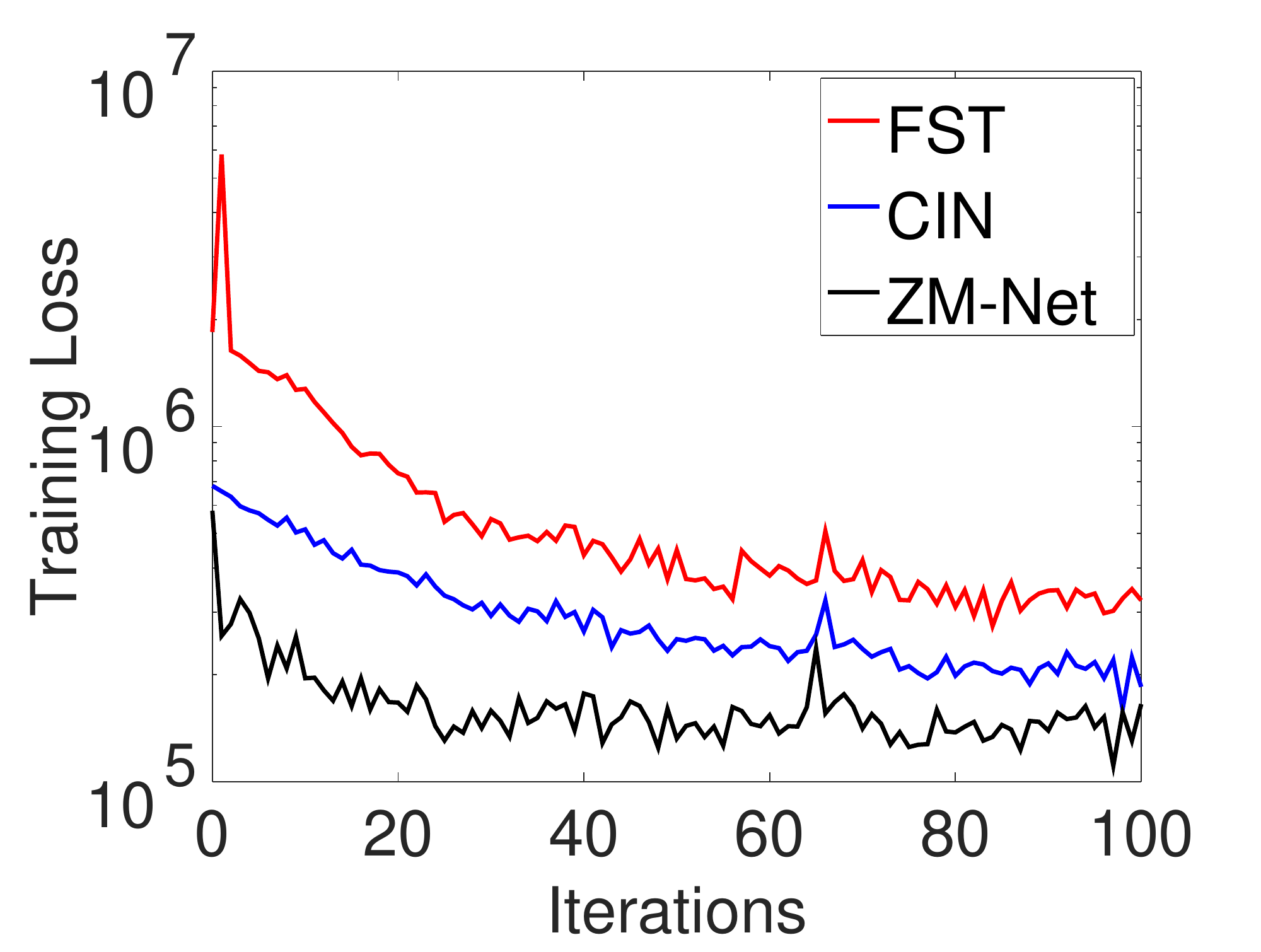}
\hspace{-0.15in}
\includegraphics[width=0.33\linewidth]{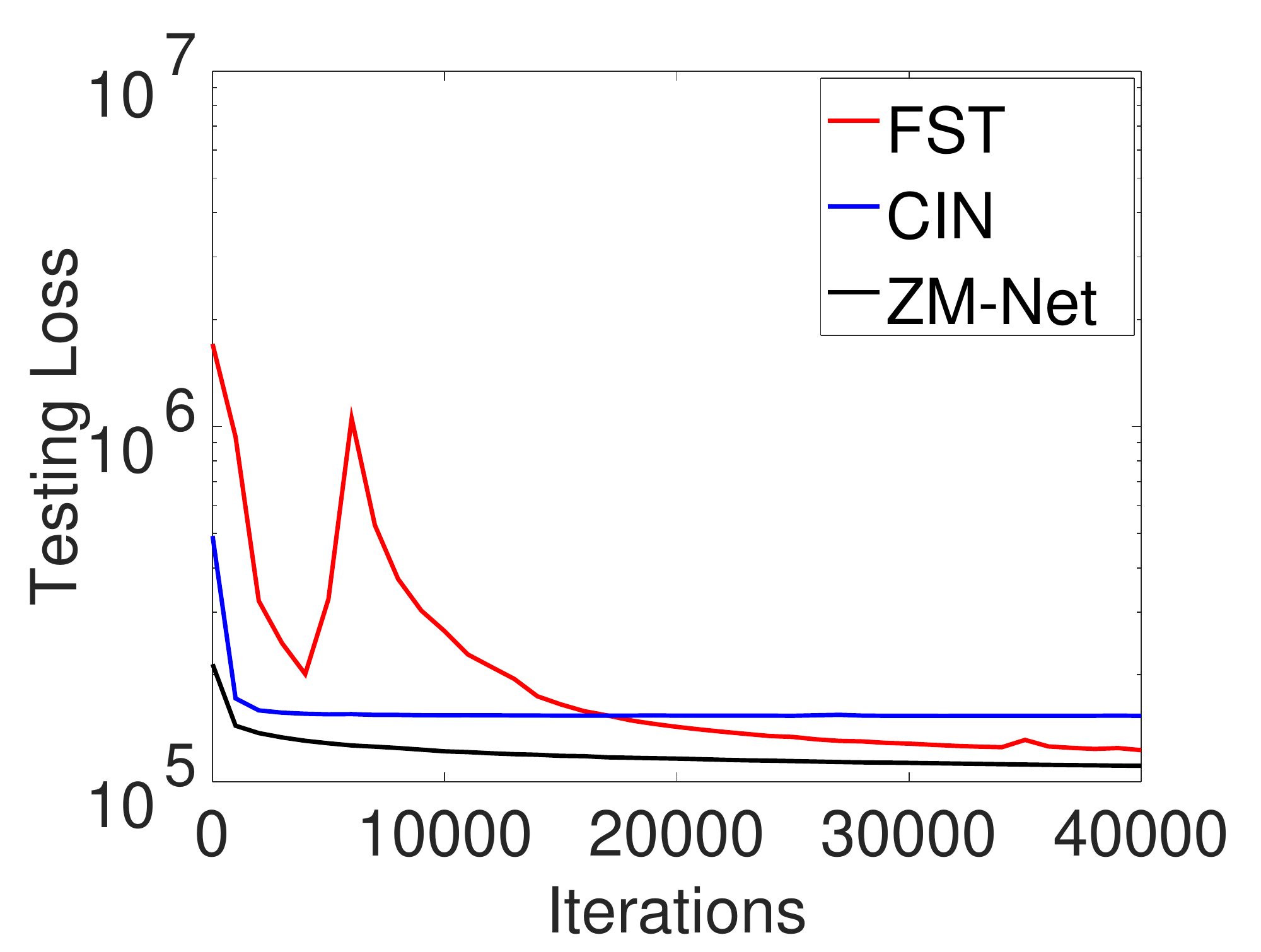}
\end{center}
\vskip -0.2in
\caption{\small Training loss for all iterations (left), training loss for the first $100$ iterations (middle), and testing loss for all iterations (right) of FST, CIN, and our ZM-Net. Testing loss is computed every $1{,}000$ iterations.}
\vskip -0.0in
\label{fig:loss}
\vskip -0.29in
\end{figure}

\begin{figure*}[t]
    \begin{center}
        \includegraphics[width=0.13\linewidth]{fig/golden.jpg}
        \includegraphics[width=0.13\linewidth]{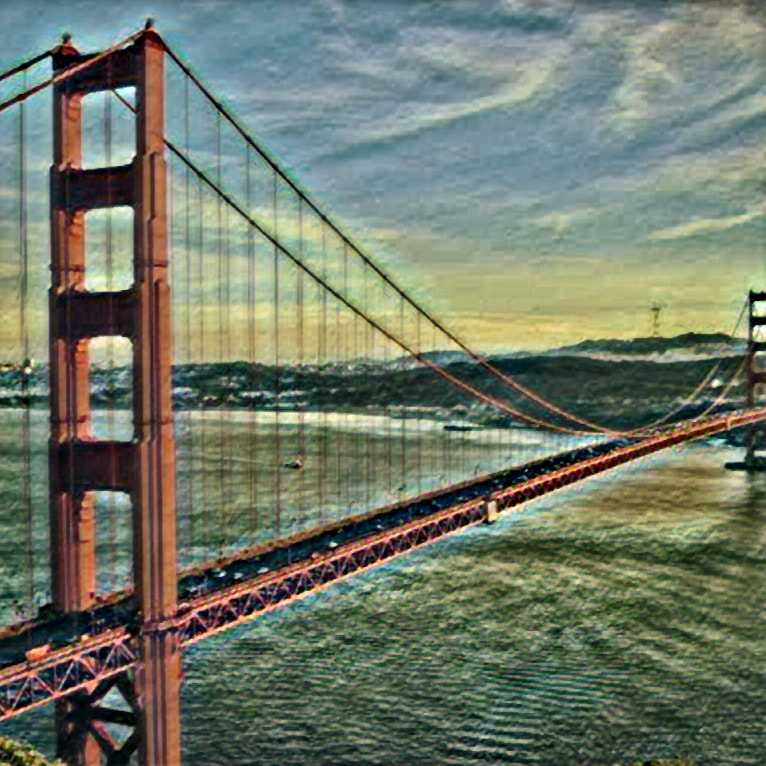}
        \includegraphics[width=0.13\linewidth]{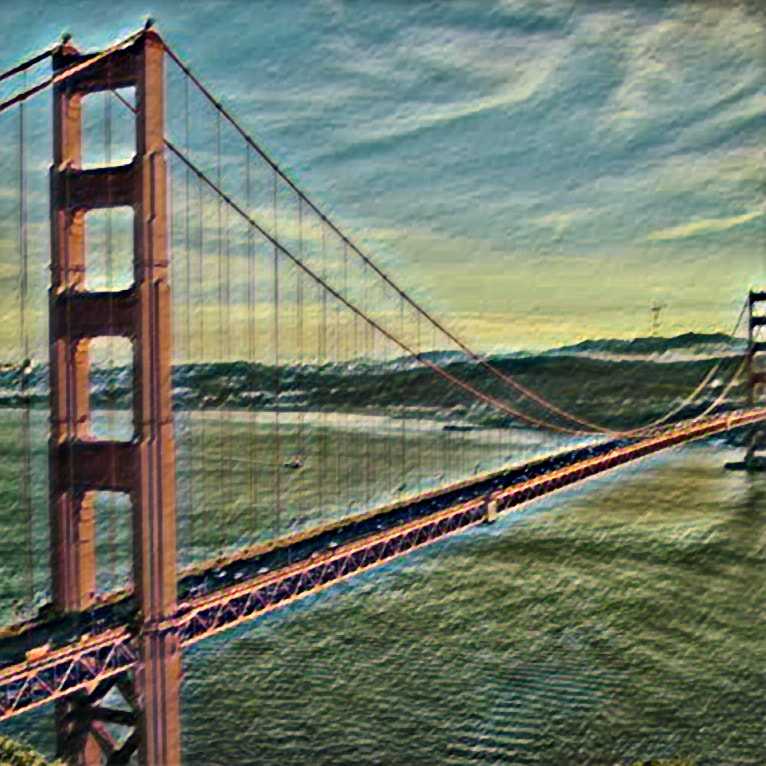}
        \includegraphics[width=0.13\linewidth]{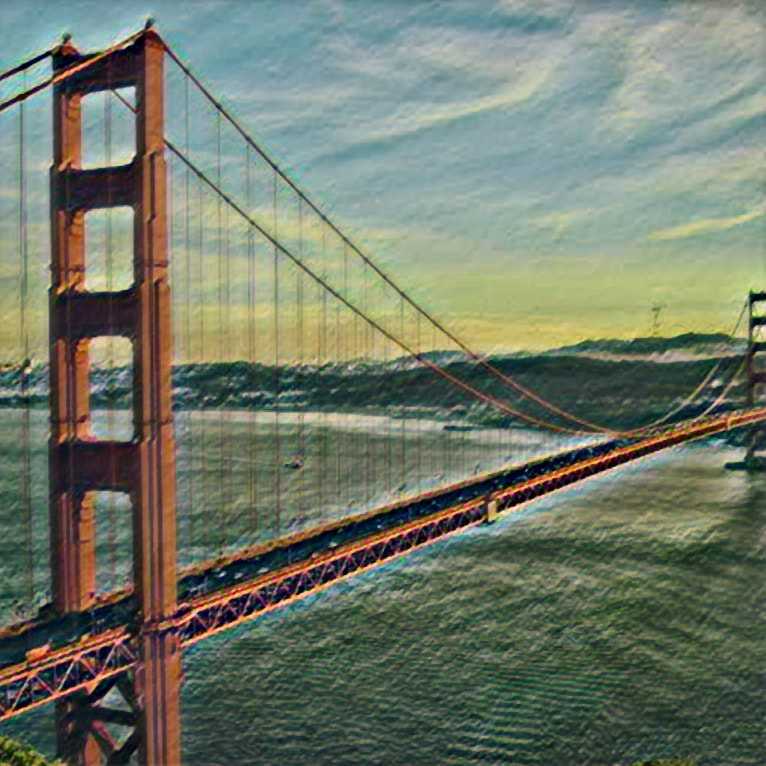}
        \includegraphics[width=0.13\linewidth]{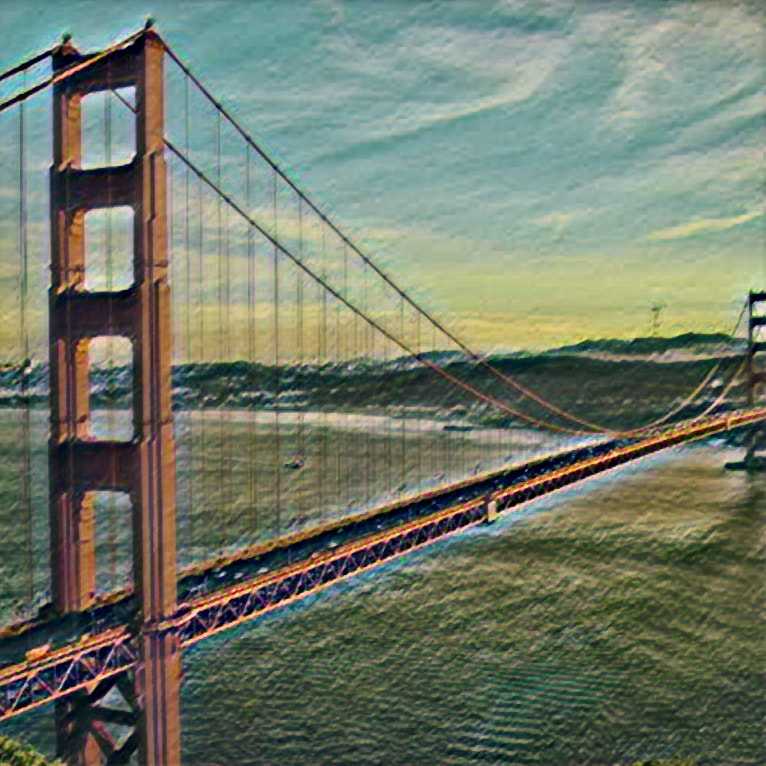}
        \includegraphics[width=0.13\linewidth]{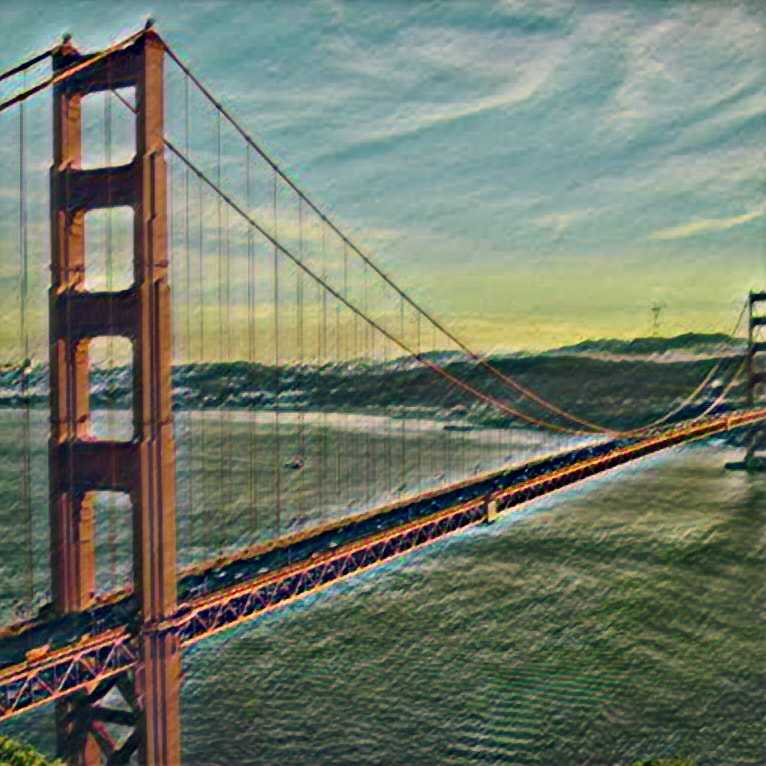}
        \includegraphics[width=0.13\linewidth]{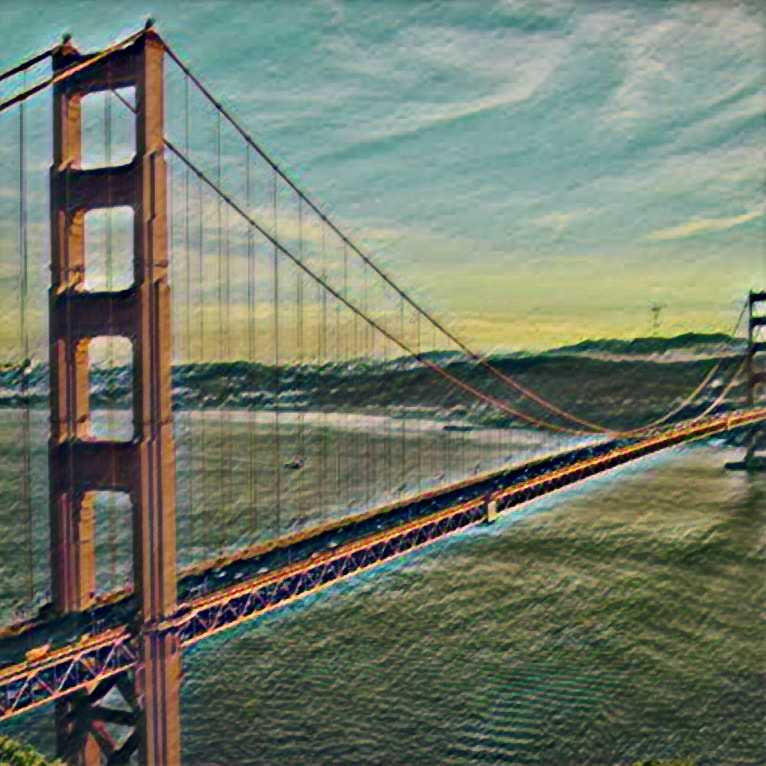} \\
        \includegraphics[width=0.13\linewidth]{fig/m1s.jpg}
        \includegraphics[width=0.13\linewidth]{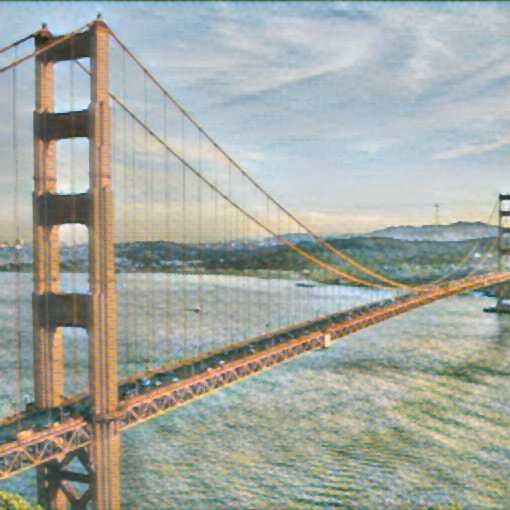}
        \includegraphics[width=0.13\linewidth]{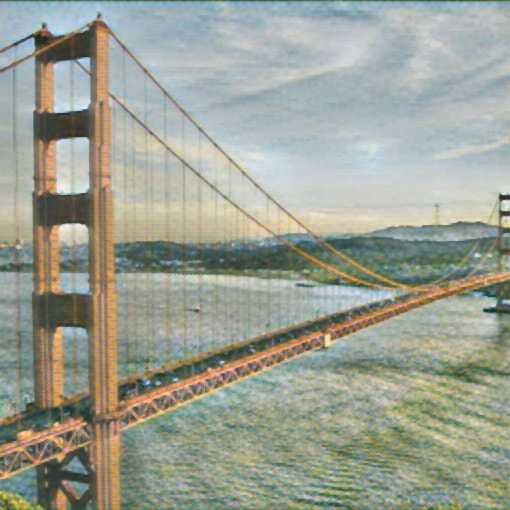}
        \includegraphics[width=0.13\linewidth]{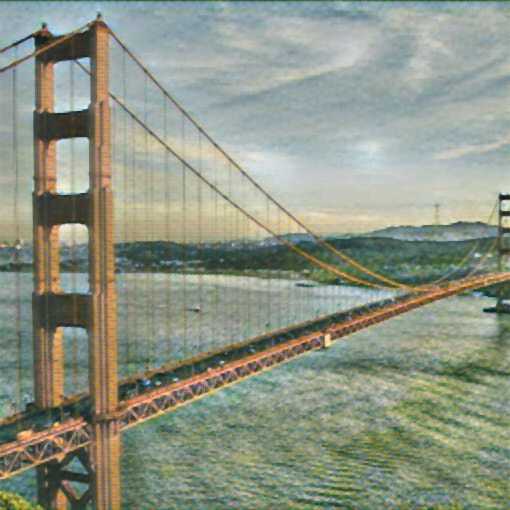}
        \includegraphics[width=0.13\linewidth]{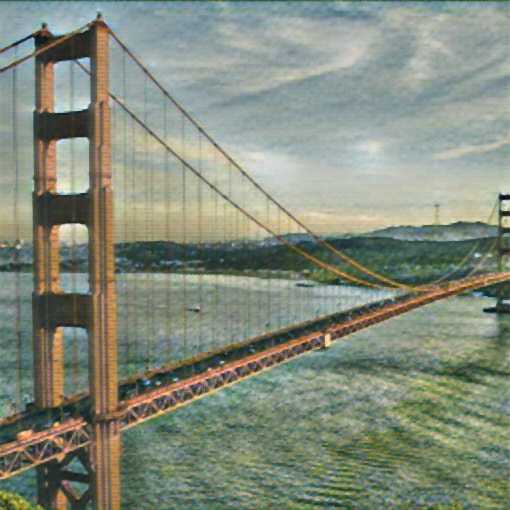}
        \includegraphics[width=0.13\linewidth]{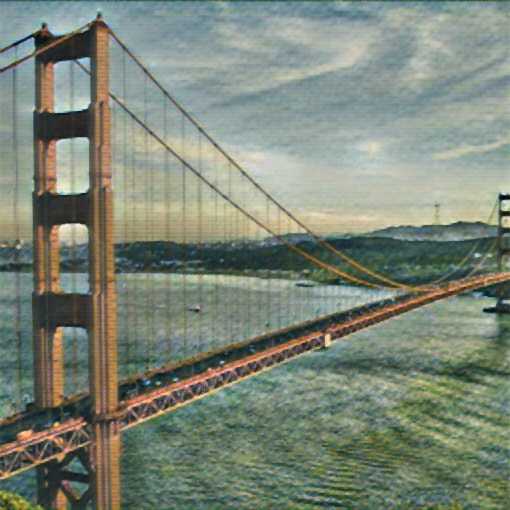}
        \includegraphics[width=0.13\linewidth]{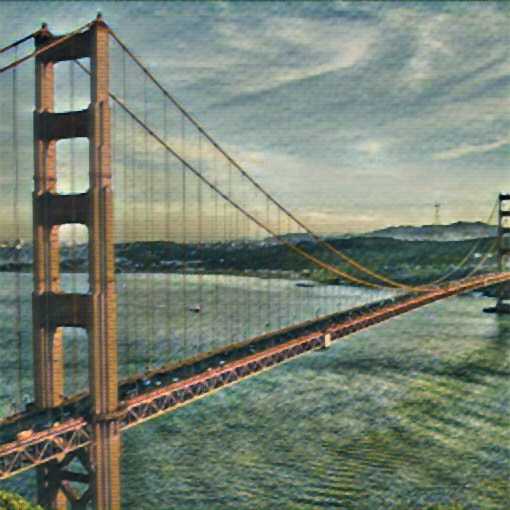}
        \vskip -0.1in
        \caption{\label{fig:finetune} \small Column 1: The content image and style image. Column 2 to 7: Style transfer for the unseen style image after finetuning ZM-Net (Row 1) and CIN \cite{RNS} (Row 2) for $1$, $10$, $20$, $30$, $40$, and $50$ iterations. The CIN model is first trained on another style image before finetuning.}
    \end{center}
    \vskip -0.25in
\end{figure*}

\textbf{Fast Zero-shot Style Transfer.}
Note that in style transfer, there are two levels of generalization involved: (1) generalization to new content images, which is achieved by \cite{RNS,TextureNet,PLoss}, and (2) generalization to not only new content images but also new style images. Since the second level involves style transfer with style images (guiding signals) unseen during training, we call this zero-shot style transfer. Figure \ref{fig:zero} shows the results of fast zero-shot style transfer using our $10$-style ZM-Net, $20{,}938$-style ZM-Net, and FST \cite{PLoss} (see the supplementary material for more results). As we can see, the $10$-style ZM-Net severely overfits the $10$ style images in the training set and generalizes poorly to unseen styles. The $20{,}938$-style ZM-Net, with the help of enough diversity in the training style images, can perform satisfactory style transfer even for unseen styles, while models like FST \cite{RNS,TextureNet,PLoss} are tied to specific styles and fail to generalize to unseen styles.

Note that both the TNet and the PNet in ZM-Net have $10$ layers ($5$ of them are residual blocks with $2$ convolutional layers each), and the PNet connects to the TNet through the first $9$ layers with the DIN operations in Equation (\ref{eq:din}). To investigate the function of DIN in different layers, we turn off the DIN operations in some layers (set $\ga_i=1$ and $\bet_i=0$) and perform zero-shot style transfer using ZM-Net. As shown in Figure \ref{fig:layers}, DIN in layer $1\sim 3$ focuses on generating content details (e.g., edges), DIN in layer $4\sim 6$ focuses on roughly adjusting colors, and DIN in layer $7\sim 9$ focuses transfer texture-related features.

\cite{RNS} proposes CIN to share convolutional layers across different styles and finetune only the scaling/shifting factors of the instance normalization, $\ga_i$ and $\bet_i$, for a new style. Figure \ref{fig:finetune} shows the style transfer for an unseen style image after finetuning CIN \cite{RNS} and ZM-Net for $1\sim 40$ iterations. As we can see, with the ability of zero-shot learning, ZM-Net can perform much better than CIN \emph{even without finetuning} for a new style. Figure \ref{fig:loss} shows the training and testing loss (sum of content and style loss) of training FST (train a transformation network from scratch), finetuning CIN, and finetuning our ZM-Net. We can conclude that, (1) finetuning CIN has much lower initial training/testing loss than FST, and finetuning ZM-Net can do even better; (2) ZM-Net converges faster and to lower training/testing loss.





\begin{figure*}[t]
    \begin{center}
        \includegraphics[width=0.13\linewidth]{fig/chicagos.jpg}
        \includegraphics[width=0.13\linewidth]{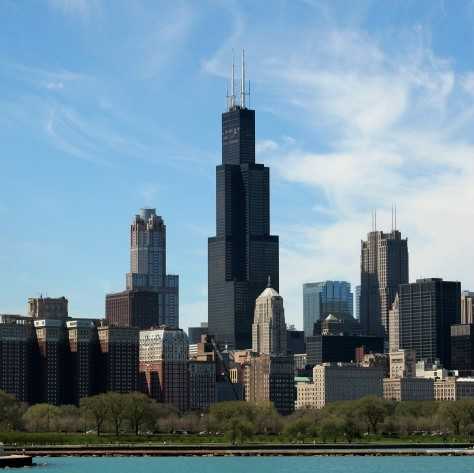}
        \includegraphics[width=0.13\linewidth]{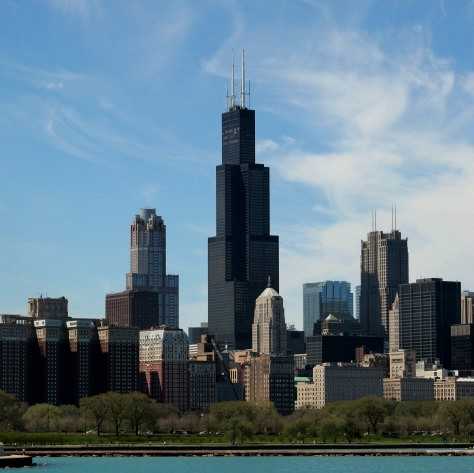}
        \includegraphics[width=0.13\linewidth]{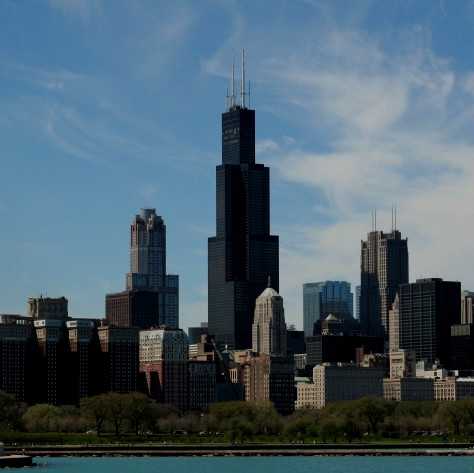}
        \includegraphics[width=0.13\linewidth]{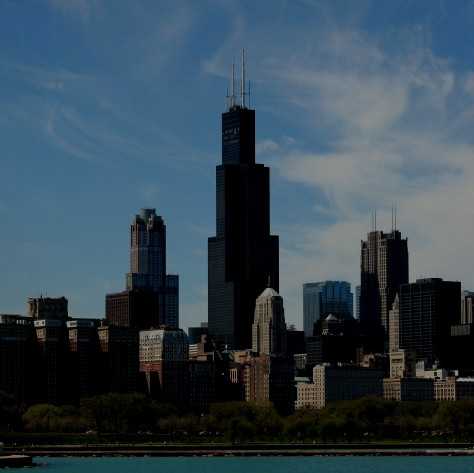}
        \includegraphics[width=0.13\linewidth]{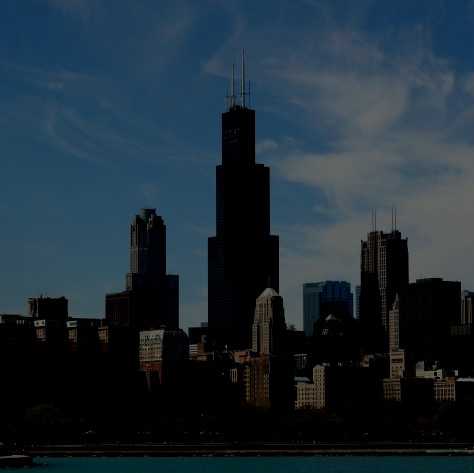}
        \includegraphics[width=0.13\linewidth]{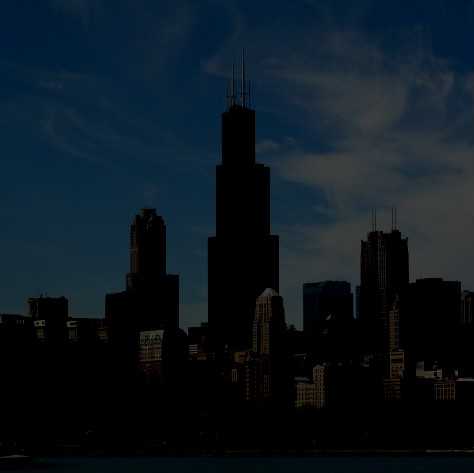} \\
        \includegraphics[width=0.13\linewidth]{fig/word.jpg}
        \includegraphics[width=0.13\linewidth]{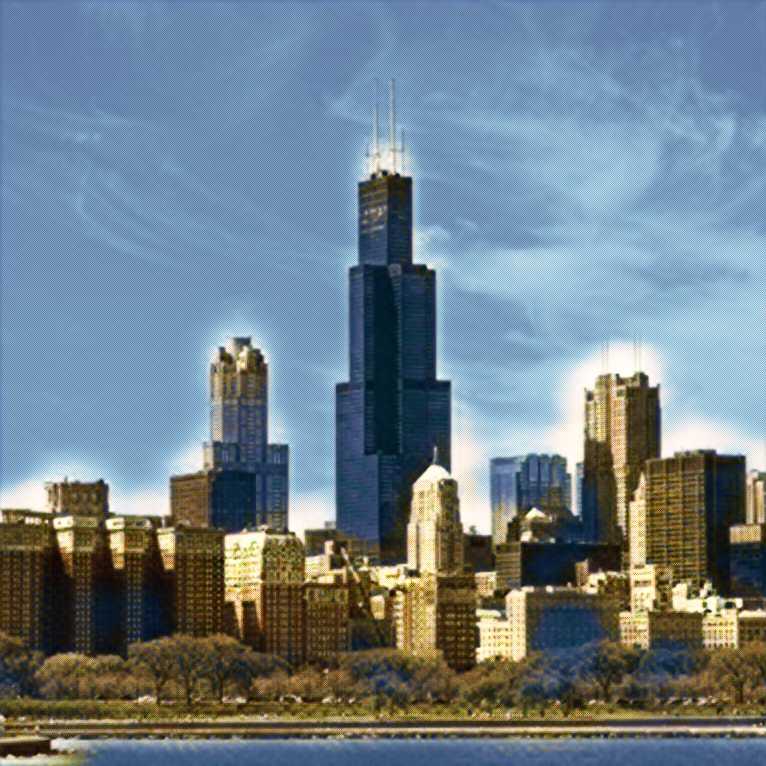}
        \includegraphics[width=0.13\linewidth]{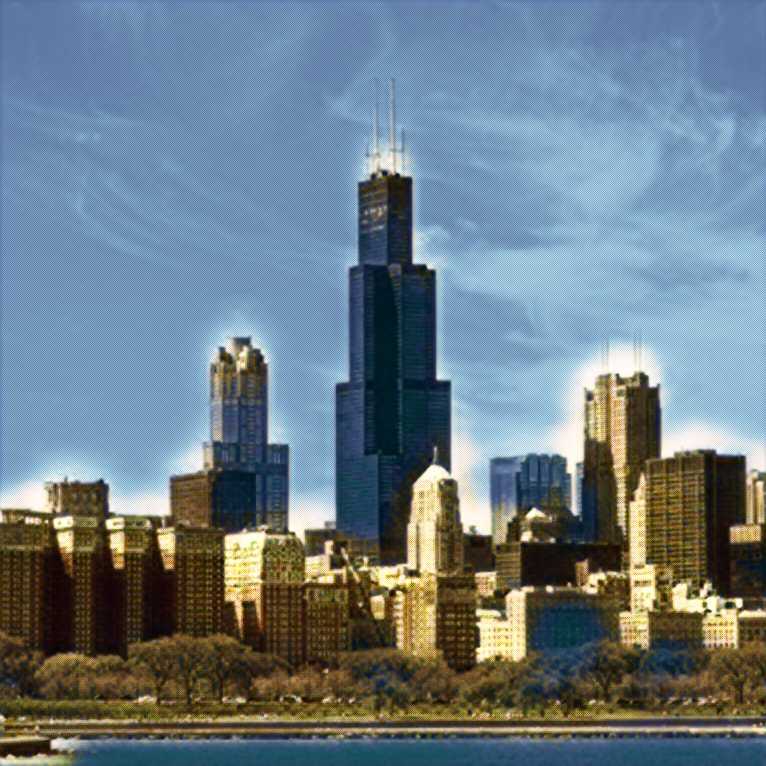}
        \includegraphics[width=0.13\linewidth]{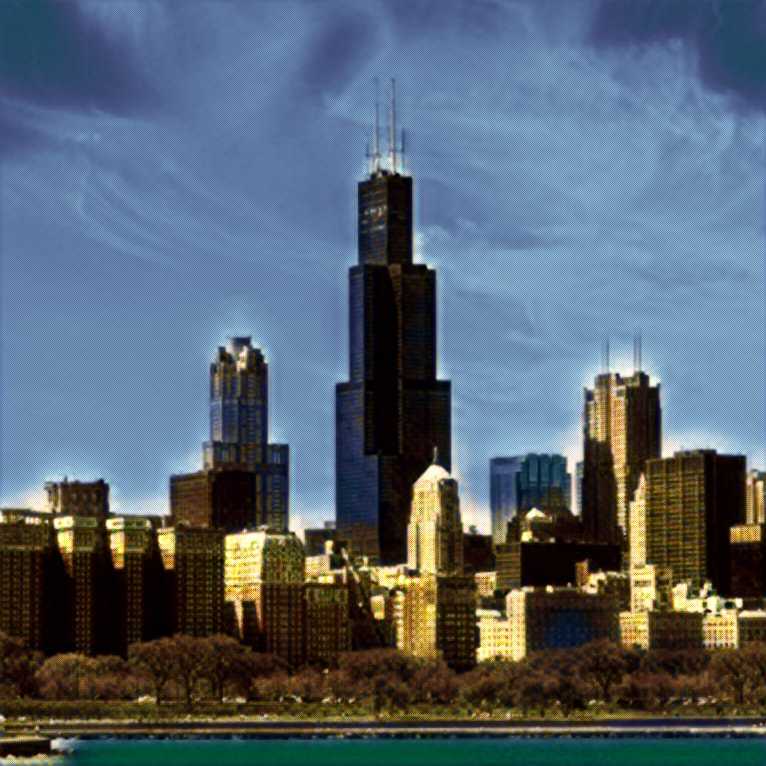}
        \includegraphics[width=0.13\linewidth]{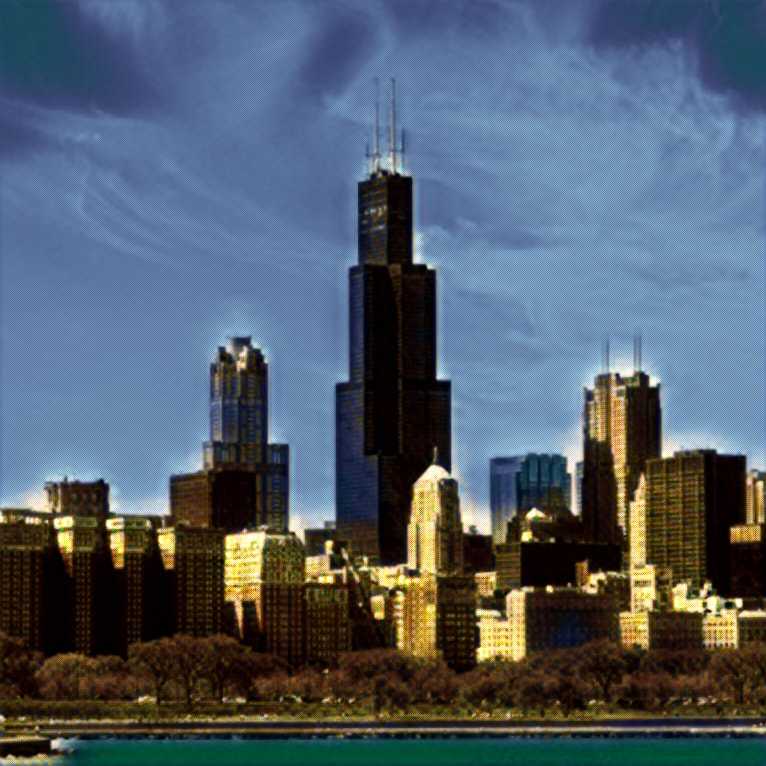}
        \includegraphics[width=0.13\linewidth]{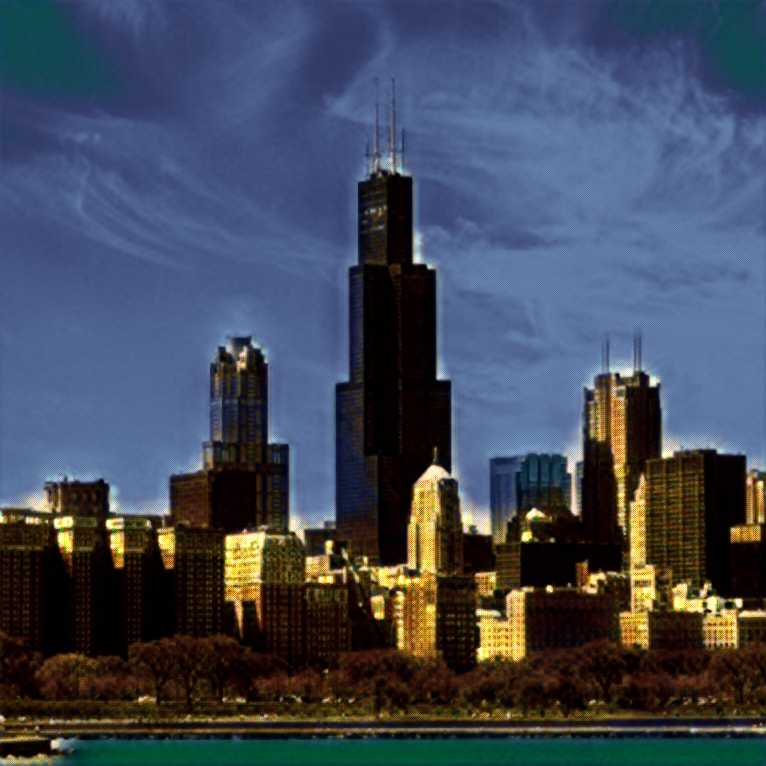}
        \includegraphics[width=0.13\linewidth]{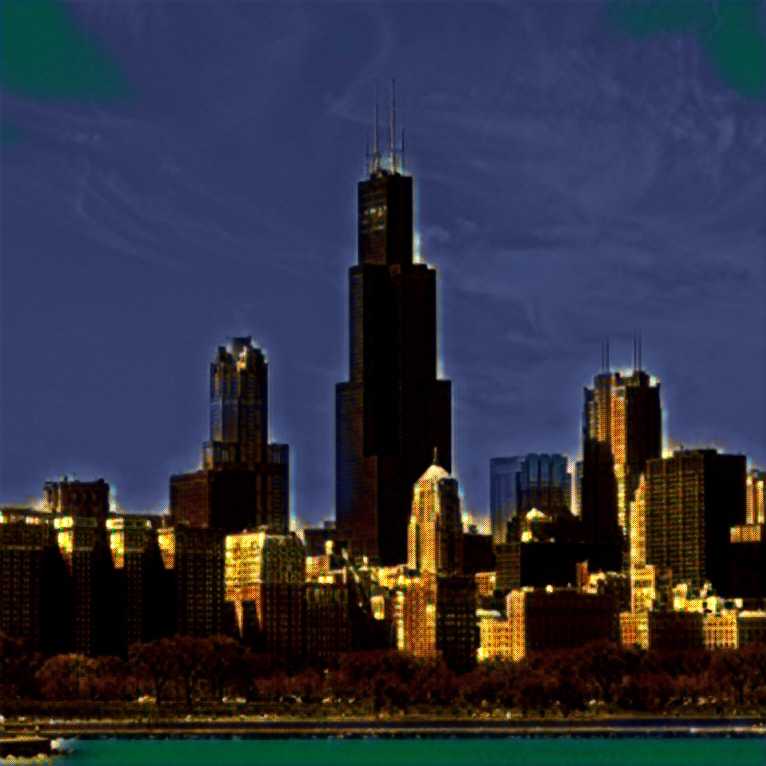} \\
        \includegraphics[width=0.13\linewidth]{fig/blank.jpg}
        \includegraphics[width=0.13\linewidth]{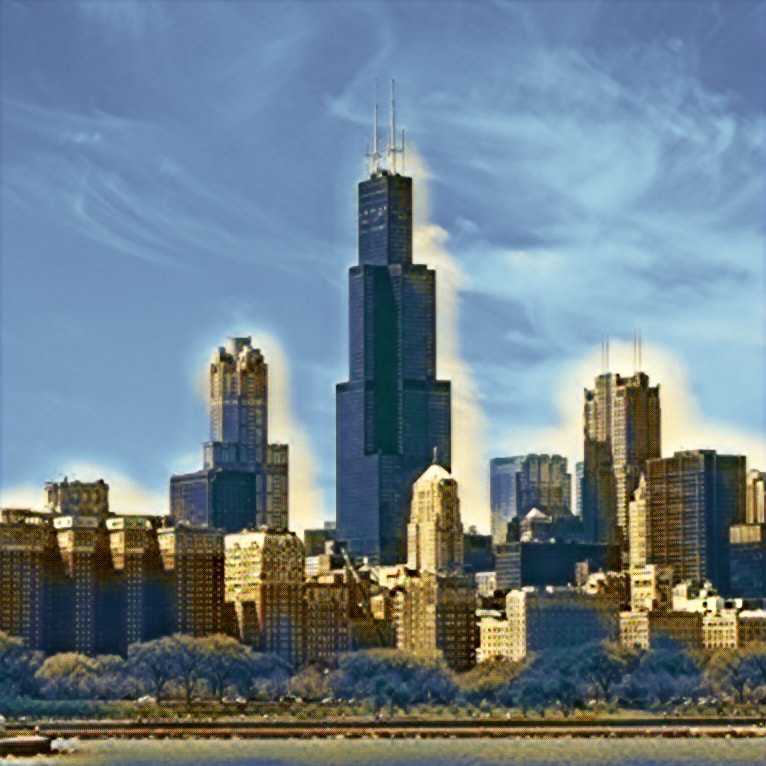}
        \includegraphics[width=0.13\linewidth]{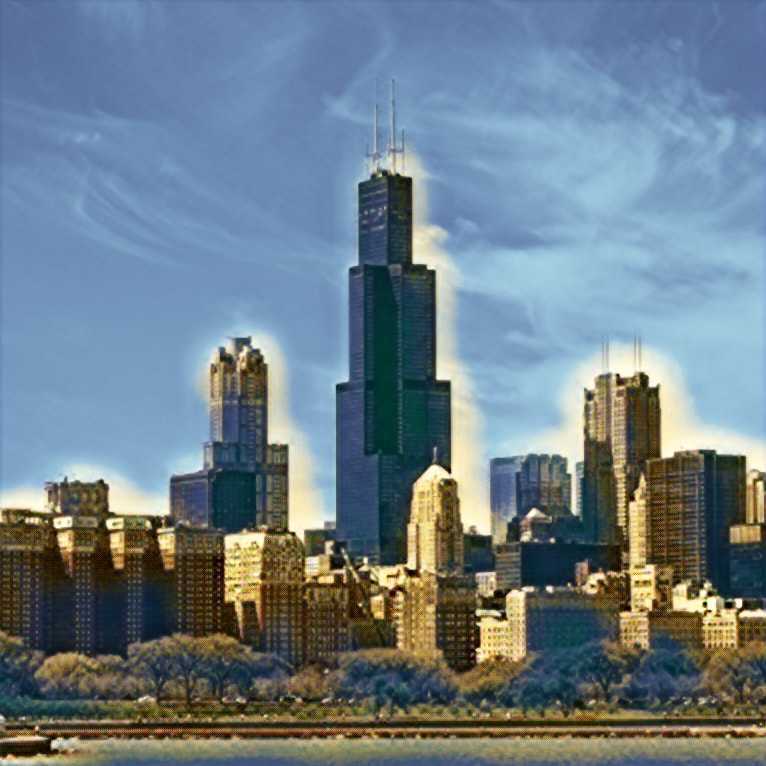}
        \includegraphics[width=0.13\linewidth]{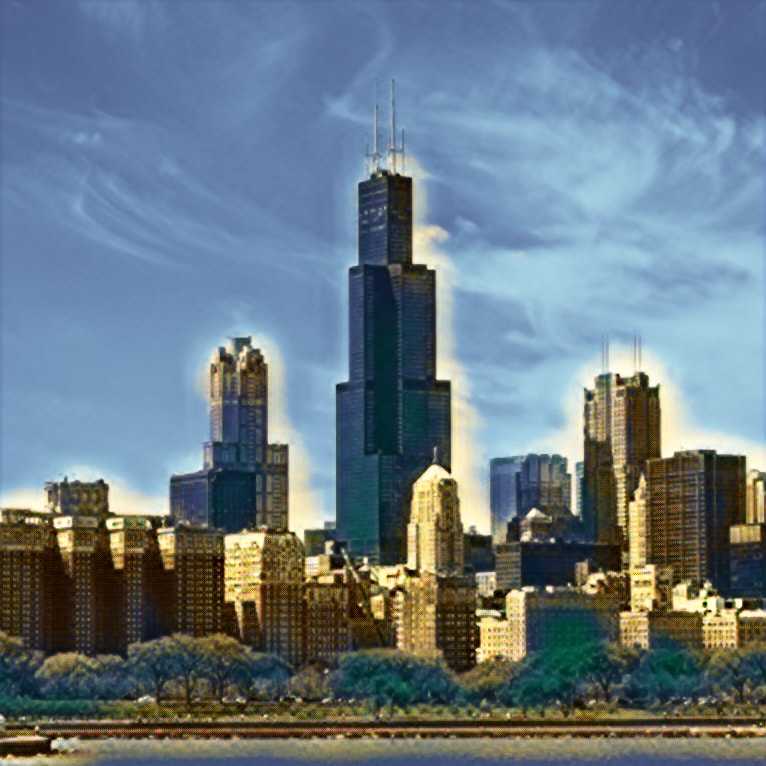}
        \includegraphics[width=0.13\linewidth]{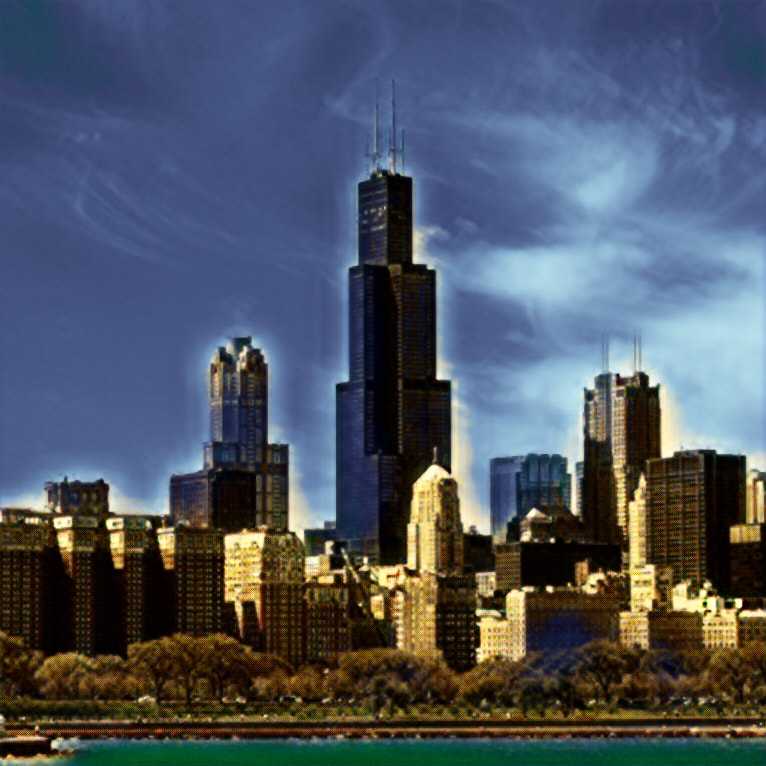}
        \includegraphics[width=0.13\linewidth]{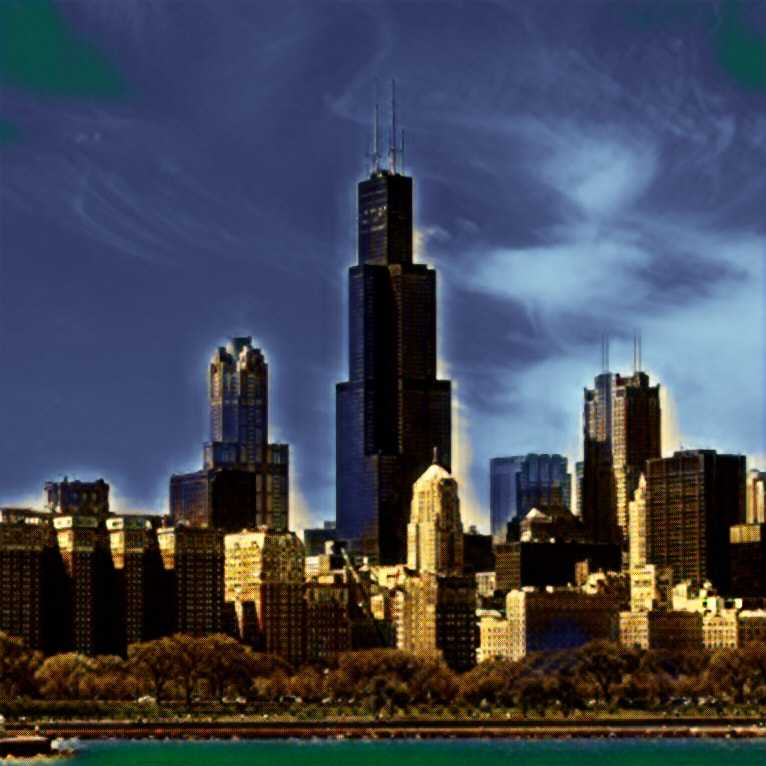}
        \includegraphics[width=0.13\linewidth]{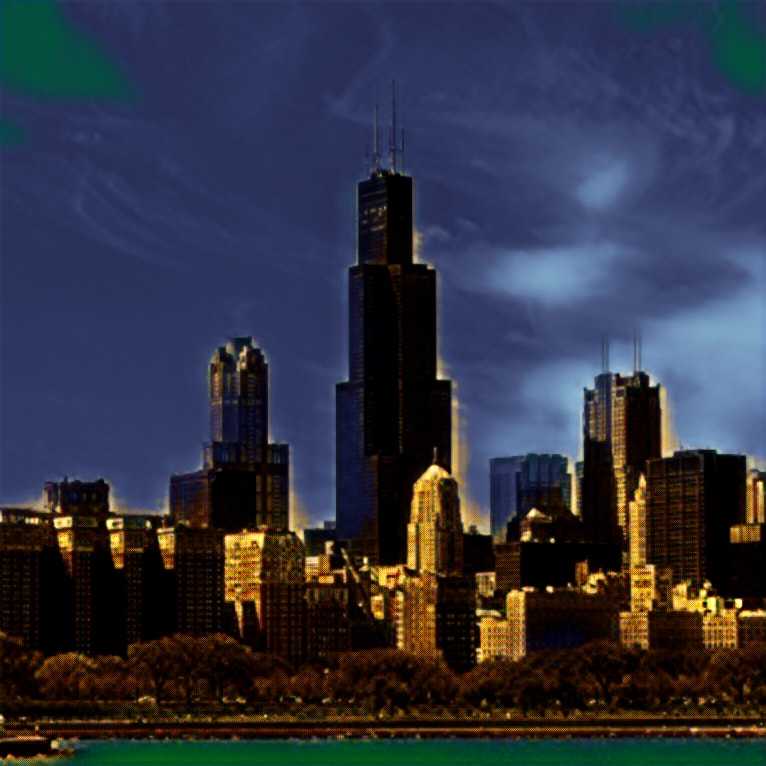}
        \vskip -0.1cm
        \caption{\label{fig:daynight} \small Zero-shot image manipulation with word embeddings as guiding signals compared to simply changing image illumination (Row 1). Row 2 shows the $6$ images corresponding to compressed word embeddings of `noon', $0.5\mbox{`noon'}+0.5\mbox{`afternoon'}$, `afternoon', `morning', $0.5\mbox{`morning'}+0.5\mbox{`night'}$, and `night' when a serial PNet is used. Row 3 shows the results when a parallel PNet is used. Column 1 shows the content image and the compressed word embeddings.}
    \end{center}
    \vskip -0.8cm
\end{figure*}

\subsection{Word Embeddings as Guiding Signals}\label{sec:word}
Besides style transfer, which uses style images as guiding signals, we also try ZM-Net with word embeddings as input to embed specific semantics into images. For example, taking the word embedding of the word `night' will transform a photo taken during daytime to a photo with a night view. In this setting, if we train ZM-Net with only the words `noon' and `night', a successful zero-shot manipulation would take the word embeding of `morning' or `afternoon' and transform the content image taken at noon to an image taken in the morning the in the afternoon (though `morning' and `afternoon' never appear in the training set).

To perform such tasks, we design a ZM-Net with a deep convolutional TNet identical to the one used for style transfer and a deep fully connected PNet with residual connections (see the supplementary material for details on the structure). To facilitate analysis and avoid overfitting, we compressed the pretrained $50$-dimensional word embeddings from \cite{glove} to $2$-dimensional vectors.


We crawl $30$ images with the tag `noon' and $30$ with the tag `night' as training images. Note that different from the ZM-Net for style transfer where the same style image is used both as input to the PNet and as input to the fixed loss network (as shown in Figure \ref{fig:pnet}), here we use word embeddings as input to the PNet and use the corresponding `noon/night' images as input to the loss network. In each iteration, we randomly select the word embeddings of `noon' or `night' as the input guiding signal and use a corresponding image to feed into the loss network. \emph{Different from style transfer}, even for the same input guiding signal, different `noon/night' images are fed into the loss network. In this case, ZM-Net is actually extracting the \emph{common patterns/semantics} from `noon' or `night' images instead of simply learning to perform style transfer.

Row 2 of Figure \ref{fig:daynight} shows the zero-shot image manipulation with a serial PNet in ZM-Net. We train the model with word embeddings of `noon' and `night' and use word embeddings of `morning' and `afternoon' (which never appear during training) as guiding signals during testing. As we can see, the transformed images gradually change from daytime (noon) views (with \emph{bright} sky and buildings) to nighttime views (with \emph{dark} sky and buildings with \emph{lights} on), with `morning/afternoon views' in between. Note that with ZM-Net's ability of fast zero-shot manipulation, it can generate \emph{animation} of a single image in real-time even though the model is \emph{image-based} (see the demonstration in the supplementary material). As a baseline, Row 1 of Figure \ref{fig:daynight} shows the results of simple illumination change. We can see that ZM-Net automatically transfer the \emph{lighting effect} (lights in the buildings) to the content image while \emph{simple illumination fails to do so}. Besides the serial PNet, we also perform the same task with a parallel PNet and report the results in Row 3 of Figure \ref{fig:daynight}. We can see that comparing to results using an serial PNet, the parallel PNet produces much more redundant yellow pixels surrounding the buildings, which is not reasonable for a daytime photo. The comparison shows that the serial PNet with its deep structure tends to perform higher-quality image manipulation than the parallel PNet.


\vspace{-5pt}
\section{Conclusion}
\vspace{-5pt}
In this paper we present ZM-Net, a general network architecture with dynamic instance normalization, to perform real-time zero-shot image manipulation. 
Experiments show that ZM-Net produces high-quality transformed images  with different modalities of guiding signals (e.g. style images and text attributes) and can generalize to \emph{unseen guiding signals}. ZM-Net can even produce real-time \emph{animation} for a single image even though the model is trained on \emph{images}. Besides, we construct the largest dataset of $23{,}307$ style images to provide much more content diversity and reduce the testing loss nearly by half. 

{\small
\bibliographystyle{ieee}
\bibliography{HTN}

\begin{thebibliography}{10}\itemsep=-1pt

\bibitem{DeepColor}
Z.~Cheng, Q.~Yang, and B.~Sheng.
\newblock Deep colorization.
\newblock In {\em ICCV}, pages 415--423, 2015.

\bibitem{RNS}
V.~Dumoulin, J.~Shlens, and M.~Kudlur.
\newblock A learned representation for artistic style.
\newblock {\em CoRR}, abs/1610.07629, 2016.

\bibitem{iccv/Eigen15}
D.~Eigen and R.~Fergus.
\newblock Predicting depth, surface normals and semantic labels with a common
  multi-scale convolutional architecture.
\newblock In {\em ICCV}, pages 2650--2658, 2015.

\bibitem{pami/Farabet13}
C.~Farabet, C.~Couprie, L.~Najman, and Y.~LeCun.
\newblock Learning hierarchical features for scene labeling.
\newblock {\em TPAMI}, 35(8):1915--1929, 2013.

\bibitem{Pandora}
C.~Florea, R.~Condorovici, C.~Vertan, R.~Butnaru, L.~Florea, and
  R.~Vr{\^a}nceanu.
\newblock Pandora: Description of a painting database for art movement
  recognition with baselines and perspectives.
\newblock In {\em EUSIPCO}, pages 918--922. IEEE, 2016.

\bibitem{SplitMatch}
O.~Frigo, N.~Sabater, J.~Delon, and P.~Hellier.
\newblock Split and match: Example-based adaptive patch sampling for
  unsupervised style transfer.
\newblock In {\em CVPR}, pages 553--561, 2016.

\bibitem{NS}
L.~A. Gatys, A.~S. Ecker, and M.~Bethge.
\newblock Image style transfer using convolutional neural networks.
\newblock In {\em CVPR}, pages 2414--2423, 2016.

\bibitem{ImageAn}
A.~Hertzmann, C.~E. Jacobs, N.~Oliver, B.~Curless, and D.~Salesin.
\newblock Image analogies.
\newblock In {\em SIGGRAPH}, pages 327--340, 2001.

\bibitem{PLoss}
J.~Johnson, A.~Alahi, and L.~Fei{-}Fei.
\newblock Perceptual losses for real-time style transfer and super-resolution.
\newblock In {\em ECCV}, pages 694--711, 2016.

\bibitem{Adam}
D.~Kingma and J.~Ba.
\newblock Adam: {A} method for stochastic optimization.
\newblock {\em arXiv preprint arXiv:1412.6980}, 2014.

\bibitem{MGan}
C.~Li and M.~Wand.
\newblock Precomputed real-time texture synthesis with markovian generative
  adversarial networks.
\newblock In {\em ECCV}, pages 702--716, 2016.

\bibitem{DNS}
Y.~Li, C.~Fang, J.~Yang, Z.~Wang, X.~Lu, and M.-H. Yang.
\newblock Diversified texture synthesis with feed-forward networks.
\newblock {\em arXiv preprint arXiv:1703.01664}, 2017.

\bibitem{DeNS}
Y.~Li, N.~Wang, J.~Liu, and X.~Hou.
\newblock Demystifying neural style transfer.
\newblock {\em CoRR}, abs/1701.01036, 2017.

\bibitem{COCO}
T.~Lin, M.~Maire, S.~J. Belongie, J.~Hays, P.~Perona, D.~Ramanan,
  P.~Doll{\'{a}}r, and C.~L. Zitnick.
\newblock Microsoft {COCO:} common objects in context.
\newblock In {\em ECCV}, pages 740--755, 2014.

\bibitem{cvpr/Liu15}
F.~Liu, C.~Shen, and G.~Lin.
\newblock Deep convolutional neural fields for depth estimation from a single
  image.
\newblock In {\em CVPR}, pages 5162--5170, 2015.

\bibitem{FCN}
J.~Long, E.~Shelhamer, and T.~Darrell.
\newblock Fully convolutional networks for semantic segmentation.
\newblock In {\em CVPR}, pages 3431--3440, 2015.

\bibitem{iccv/Noh15}
H.~Noh, S.~Hong, and B.~Han.
\newblock Learning deconvolution network for semantic segmentation.
\newblock In {\em ICCV}, pages 1520--1528, 2015.

\bibitem{glove}
J.~Pennington, R.~Socher, and C.~D. Manning.
\newblock Glove: Global vectors for word representation.
\newblock In {\em EMNLP}, 2014.

\bibitem{icml/Pinheiro14}
P.~H.~O. Pinheiro and R.~Collobert.
\newblock Recurrent convolutional neural networks for scene labeling.
\newblock In {\em ICML}, pages 82--90, 2014.

\bibitem{ImageNet}
O.~Russakovsky, J.~Deng, H.~Su, J.~Krause, S.~Satheesh, S.~Ma, Z.~Huang,
  A.~Karpathy, A.~Khosla, M.~S. Bernstein, A.~C. Berg, and F.~Li.
\newblock Image{N}et large scale visual recognition challenge.
\newblock {\em IJCV}, 115(3):211--252, 2015.

\bibitem{VGG}
K.~Simonyan and A.~Zisserman.
\newblock Very deep convolutional networks for large-scale image recognition.
\newblock {\em CoRR}, abs/1409.1556, 2014.

\bibitem{TextureNet}
D.~Ulyanov, V.~Lebedev, A.~Vedaldi, and V.~S. Lempitsky.
\newblock Texture networks: Feed-forward synthesis of textures and stylized
  images.
\newblock In {\em ICML}, pages 1349--1357, 2016.

\bibitem{InsNorm}
D.~Ulyanov, A.~Vedaldi, and V.~S. Lempitsky.
\newblock Instance normalization: The missing ingredient for fast stylization.
\newblock {\em CoRR}, abs/1607.08022, 2016.

\bibitem{eccv/Zhang16}
R.~Zhang, P.~Isola, and A.~A. Efros.
\newblock Colorful image colorization.
\newblock In {\em ECCV}, pages 649--666, 2016.

\bibitem{CRFRNN}
S.~Zheng, S.~Jayasumana, B.~Romera{-}Paredes, V.~Vineet, Z.~Su, D.~Du,
  C.~Huang, and P.~H.~S. Torr.
\newblock Conditional random fields as recurrent neural networks.
\newblock In {\em ICCV}, pages 1529--1537, 2015.

\end{thebibliography}
}

\end{document}